\documentclass[twoside,11pt]{article}

\usepackage{jmlr2e}
\usepackage{macros}
\usepackage{epstopdf}
\usepackage{amsmath}

\newcommand{\BEAS}{\begin{eqnarray*}}
\newcommand{\EEAS}{\end{eqnarray*}}
\newcommand{\BEQ}{\begin{equation}}
\newcommand{\EEQ}{\end{equation}}
\newcommand{\BIT}{\begin{itemize}}
\newcommand{\EIT}{\end{itemize}}

\newcommand{\expec}{\mathbb{E}}

\usepackage{color}
\usepackage{amsmath}
\usepackage{amssymb}
\usepackage{graphicx}
\usepackage{comment,xspace}
\usepackage{fancybox}
\newcommand{\real}{{\mathbb{R}}}
\newcommand{\reals}{\real}

\newtheorem{assumption}[theorem]{Assumption}

\newcommand{\argmin}{\operatornamewithlimits{argmin}}

\def\A{\mathcal{A}}
\def\R{R}

\def\P{P}

\def\Q{Q}
\def\X{\mathcal{X}}

\jmlrheading{1}{xxxx}{x-xx}{x/xx}{xx/xx}{Y. Chow, M. Ghavamzadeh,  L. Janson, and M. Pavone}

\ShortHeadings{Risk-Constrained Reinforcement Learning with Percentile Risk Criteria}{Chow, Ghavamzadeh, Janson and Pavone}
\firstpageno{1}

\begin{document}

\title{Risk-Constrained Reinforcement Learning \\with Percentile Risk Criteria}

\author{\name Yinlam Chow \email ychow@stanford.edu \\
       \addr Institute for Computational \& Mathematical Engineering\\
       Stanford University\\
       Stanford, CA 94305, USA
        \AND
       \name Mohammad Ghavamzadeh \email ghavamza@adobe.com  \\
       \addr Adobe Research \& INRIA Lille \\
       San Jose, CA 95110, USA
       \AND
       \name Lucas Janson \email ljanson@stanford.edu \\
       \addr Department of Statistics\\
       Stanford University\\
       Stanford, CA 94305, USA
        \AND
     \name Marco Pavone  \email pavone@stanford.edu \\
       \addr Aeronautics and Astronautics\\
       Stanford University\\
       Stanford, CA 94305, USA
}

\editor{Jan Peters}

\maketitle

\begin{abstract}%   <- trailing '%' for backward compatibility of .sty file
%{\bf TODO: The title is too general, perhaps something to show that it is not about any risk constrained optimization, but percentile based risk constrained optimization.}
In many sequential decision-making problems one is interested in
minimizing an expected cumulative cost while taking into account
\emph{risk}, i.e., increased awareness of events of small probability
and high consequences. Accordingly, the objective of this paper is to
present efficient reinforcement learning algorithms for
risk-constrained Markov decision processes (MDPs), where risk is
represented via a {\color{black} chance constraint} or a constraint on the conditional value-at-risk (CVaR) of the cumulative cost. We collectively refer to such problems as  percentile risk-constrained MDPs.
%we may want to manage risk by minimizing some measure of variability in costs in addition to minimizing the standard expected cumulative cost. In many engineering applications, the notion of risk refers to a metric that guarantees safety at each stage of an operation, where controlling risk is equivalent to bounding the mission failure probability. On the other hand in many financial and marketing applications, the notion of risk refers to a measure that controls the costs in the tail distribution (the worst-case distribution). Often times, the tail risk in such applications are captured using Conditional value-at-risk (CVaR). CVaR is a relatively new risk measure that addresses some of the shortcomings of the well-known variance-related risk measures, and because of its computational efficiencies has gained popularity in finance and operations research. In this paper, we consider both the chance constrained and CVaR constrained optimization in Markov decision processes.
Specifically, we first derive a formula for computing the gradient of the Lagrangian function for percentile risk-constrained MDPs. Then, we devise policy gradient and actor-critic algorithms that (1) estimate such gradient, (2) update the policy in the descent direction, and (3) update the Lagrange multiplier in the ascent direction. For these algorithms we prove convergence to {\color{black} locally optimal} policies. Finally, we demonstrate the effectiveness of our algorithms in an optimal stopping problem and an online marketing application.
\end{abstract}


\section{Introduction}
\label{sec:introduction}
The most widely-adopted optimization criterion for  Markov decision
processes (MDPs) is represented by the \emph{risk-neutral} expectation
of a cumulative cost. However, in many applications one is interested
in taking into account risk, i.e.,  increased awareness of events of small probability and high consequences. Accordingly, 
%  in addition to this standard optimization criterion. In such cases, we would like to use a criterion that incorporates a penalty for the {\em variability} induced by a given policy. 
%The variability is due to uncertainties in model parameters (\emph{robust} MDPs) and the stochastic nature of the system (\emph{risk sensitive} MDPs). 
in {\em risk-sensitive} MDPs the objective is to minimize a
risk-sensitive criterion such as the expected exponential utility, a
variance-related measure, or percentile performance. There are several
risk metrics available in the literature, and constructing a ``good"
risk criterion in a manner that is both conceptually meaningful and
computationally tractable {\color{black} remains a topic of current research}.

\emph{Risk-Sensitive MDPs}: 
One of the earliest risk metrics used for risk-sensitive MDPs is the exponential risk metric $(1/\gamma)\mathbb E\big[\exp(\gamma Z)\big]$, where $Z$ represents the cumulative cost for a sequence of decisions~\citep{Howard72RS}. 
%In~\citet{kadota1995discounted,chung1987discounted}, moment optimality of the exponential utility in infinite horizon discounted was investigated.
In this setting, the degree of risk-aversion is controlled by the parameter $\gamma$, whose  selection, however,  is often challenging. This motivated the study of several different approaches.  In~\citet{collins1997using}, the authors considered the maximization of a strictly concave functional of the distribution of the terminal state. In~\citet{wu1999minimizing,boda2004stochastic,Filar95PP}, risk-sensitive MDPs are cast as the problem of maximizing  percentile performance. Variance-related risk metrics are considered, e.g., in~\citet{Sobel82VD,filar1989variance}. Other mean, variance, and probabilistic criteria for risk-sensitive MDPs are discussed in the  survey \citep{white1988mean}. 

Numerous alternative risk metrics have recently  been  proposed in the
literature, usually with the goal of providing an ``intuitive'' notion
of risk and/or to ensure computational tractability. {\em
  Value-at-risk} (VaR) and {\em conditional value-at-risk} (CVaR)
represent two promising such alternatives. They both aim at
quantifying costs that might be encountered in the tail of a cost
distribution, but in different ways. Specifically, for continuous cost
distributions, VaR$_\alpha$ measures risk as the maximum cost that
might be incurred with respect to a given confidence level
$\alpha$. {\color{black} This risk metric is particularly useful when there is a well-defined
  failure state, e.g., a state that leads a robot to collide with an
  obstacle. A VaR$_\alpha$ constraint is often referred to as a chance
  (probability)  constraint, especially in the engineering literature, and we will use this terminology in the remainder of the paper.} In contrast, CVaR$_\alpha$ measures risk as the
expected cost given that such cost is greater than or equal to
VaR$_\alpha$, and provides a number of  theoretical and computational
advantages. CVaR optimization was first developed by Rockafellar and
Uryasev~{\color{black} \citep{Rockafellar02CV,Rockafellar00OC}} and its numerical effectiveness has
been demonstrated in several portfolio optimization and option hedging problems. Risk-sensitive MDPs with a conditional value at risk metric were considered in~\citet{Boda06TC,Ott10MD,Bauerle11MD}, and a mean-average-value-at-risk problem has been solved in~\citet{bauerle2009dynamic} for minimizing risk in financial markets.  

%More applications in risk-sensitive sequential decision-making problems include revenue maximization~\citep{barz2007risk} and portfolio management~\citep{bielecki2002credit}. 

The aforementioned works focus on the derivation of exact solutions, and the ensuing algorithms are only applicable to relatively small problems. This has recently motivated the application of reinforcement learning (RL) methods to risk-sensitive MDPs. We will refer to such problems as risk-sensitive RL.

% add sub-section to SPSA
\emph{Risk-Sensitive RL}: To address large-scale problems, it is natural to apply reinforcement learning (RL) techniques to risk-sensitive MDPs. Reinforcement learning~\citep{BertsekasT96,sutton1998introduction} can be viewed as a class of sampling-based methods for solving MDPs.  
%Model-based RL~\citep{tesauro1995temporal,crites1998elevator} uses sample trajectories obtained from simulations to approximately compute optimal policies. If no simulator is available, one can use model-free RL methods~\citep{kohl2004policy,ng2006autonomous}, whereby sample trajectories are obtained online. Popular 
Popular reinforcement learning techniques include policy
gradient~\citep{williams1992simple,Marbach98SM,Baxter01IP} and
actor-critic
methods~\citep{Sutton00PG,Konda00AA,Peters05NA,borkar2005actor,bhatnagar2009natural,bhatnagar2012online},
whereby policies are parameterized in terms of a parameter vector and
policy search is performed via gradient flow approaches. One effective
way to estimate gradients in RL problems is by simultaneous
perturbation stochastic approximation
(SPSA)~\citep{Spall92MS}. Risk-sensitive RL with expected exponential
utility has been  considered in ~\citet{Borkar01SF,Borkar02QR}. More
recently, the works in ~\citet{tamar2012policy,Prashanth13AC} present
RL algorithms for several variance-related risk measures, the
works in \citet{Morimura10NR,Tamar15PG,Petrik12AS} consider CVaR-based
formulations, {\color{black} and the works in \citet{le2007robust,
    shapiro2013risk} consider nested CVaR-based formulations.}

\emph{Risk-Constrained RL and Paper Contributions}:
Despite the rather large literature on risk-sensitive MDPs and RL, \emph{risk-constrained} formulations have largely gone unaddressed, with only a few exceptions, e.g., \citet{YC-MP:13,Borkar14RC}. Yet constrained formulations naturally
arise in several domains, including engineering, finance, and
logistics, and provide a principled approach to address multi-objective problems. The objective of this paper is to fill this gap by devising policy gradient and actor-critic algorithms for risk-constrained MDPs, where risk is represented via a constraint on the conditional value-at-risk (CVaR) of the cumulative cost or as a {\color{black} chance constraint}.   Specifically, the contribution of this paper is fourfold. 
\begin{enumerate}
\item We formulate two risk-constrained MDP problems. The first one
  involves a CVaR constraint and the second one involves a chance
 constraint. For the CVaR-constrained optimization problem, we
 consider both discrete and continuous cost distributions. By
 re-writing the problems using a Lagrangian formulation, we derive for
 both problems a Bellman optimality condition with respect to an
 augmented MDP {\color{black} whose state consists of two parts, with
   the first part capturing the state of the original MDP and the
   second part keeping track of the cumulative constraint cost}.
\item We devise a trajectory-based policy gradient algorithm for both CVaR-constrained and {\color{black} chance-constrained} MDPs. The key novelty of this algorithm lies in an unbiased gradient estimation procedure  under Monte Carlo sampling. Using an ordinary differential equation (ODE) approach, we establish convergence of the algorithm to {\color{black} locally optimal}  policies.
\item Using the aforementioned Bellman optimality condition, we derive several actor-critic algorithms to optimize policy and value function approximation parameters in an online fashion. As for the trajectory-based policy gradient algorithm, we show that the proposed actor-critic algorithms converge to {\color{black} locally optimal}  solutions. 
\item We demonstrate the effectiveness of our algorithms in an optimal stopping problem as well as in a realistic personalized {\color{black} advertisement recommendation (ad recommendation)} problem (see~\citet{derfer2007online} for more details). For the latter problem, we empirically show that our CVaR-constrained RL algorithms successfully guarantee that the worst-case revenue is lower-bounded by the pre-specified company yearly target. 
\end{enumerate}

The rest of the paper is structured as follows. In Section~\ref{sec:preliminaries} we introduce our notation and rigorously state the problem we wish to address, namely risk-constrained RL. The next two sections provide various RL methods to
approximately compute (locally) optimal policies for CVaR constrained
MDPs. A trajectory-based policy gradient algorithm is presented in
Section~\ref{sec:PG-alg} and its convergence analysis is provided in
Appendix~\ref{sec:appendix_PG} {\color{black} (Appendix~\ref{subsec:grad-comp} provides the gradient estimates of the CVaR
  parameter, the policy parameter, and the Lagrange multiplier, and
  Appendix~\ref{subsec:conv-proof} gives their convergence proofs)}. Actor-critic algorithms are presented in
Section~\ref{sec:AC-alg} and their convergence analysis is provided in Appendix~\ref{sec:appendix_AC}
{\color{black} (Appendix~\ref{subsec:grad-lambda-comp} derives the gradient of the Lagrange
  multiplier as a function of the state-action value function,
  Appendix~\ref{subsec:v_update} analyzes the convergence of the critic, and
  Appendix~\ref{subsec:actor_update} provides the multi-timescale convergence results of the
  CVaR parameter, the policy parameter, and the Lagrange
  multiplier)}. Section~\ref{sec:gen_chance} {\color{black} extends} the above policy gradient and
actor-critic methods to the {\color{black} chance-constrained}
case. Empirical evaluation of our algorithms  is the subject of
Section~\ref{sec:example}. Finally, we conclude the paper in Section~\ref{sec:conclusions}, where we also
provide directions for  future work.

This paper generalizes earlier results by the authors presented in \citet{chow2014algorithms}.

\section{Preliminaries} 
\label{sec:preliminaries}
We begin by defining some notation that is used throughout the paper, as well as defining the problem addressed herein and stating some basic assumptions.
\subsection{Notation}
\label{sec:notation}
We consider {\color{black} decision-making} problems modeled as a finite MDP ({\color{black} an}{ MDP with finite state and
action spaces). A finite MDP is a tuple $(\X,\A,C,D,P,P_0)$ where
$\X=\{1,\ldots,n,{\color{black} x_{\text{Tar}}}\}$ and $\A=\{1,\ldots,m\}$ are the state and
action spaces, ${\color{black} x_{\text{Tar}}}$ is a recurrent target
state, {\color{black} and for a state $x$ and an action $a$,} $C(x,a)$ is a cost function
with $|C(x,a)|\le C_{\max}$, $D(x,a)$ is a constraint cost function
with $|D(x,a)|\le D_{\max}$ \footnote{{\color{black} Without loss of generality, we set the cost function $C(x,a)$ and constraint cost function $D(x,a)$ to zero when $x=x_{\text{Tar}}$}.}, $P(\cdot|x,a)$ is the transition probability distribution, and $P_0(\cdot)$ is the initial state distribution. For simplicity, in this paper we assume $P_0=\mathbf 1\{x=x^0\}$ for some given initial state $x^0\in\{1,\ldots,n\}$.  Generalizations to non-atomic initial state distributions are straightforward, for which the details are omitted for the sake of  brevity. A {\em stationary policy} $\mu(\cdot|x)$ for an MDP is a probability distribution over actions, conditioned on the current state. In policy gradient methods, such policies are parameterized by a $\kappa$-dimensional vector $\theta$, so the space of policies can be written as $\big\{\mu(\cdot|x;\theta),x\in\X,\theta\in\Theta\subseteq\R^{\kappa}\big\}$. Since in this setting a policy $\mu$ is uniquely defined by its parameter vector $\theta$, policy-dependent functions can be written as a function of {\color{black} $\mu$ or $\theta$}, and we use {\color{black} $\mu(\cdot|x;\theta)$ to denote the policy and $\theta$ to denote the dependency on the policy (parameter)}.

{\color{black} Given} a fixed $\gamma\in(0,1)$, we denote by
$d_\gamma^\mu(x|x^0)=(1-\gamma)\sum_{k=0}^\infty\gamma^k \mathbb
P(x_k=x|x_0=x^0;\mu)$ and
$\pi_\gamma^\mu(x,a|x^0)=d_\gamma^\mu(x|x^0)\mu(a|x)$, the
$\gamma$-discounted {\color{black} occupation measure}  of state $x$
and state-action pair $(x,a)$ under policy $\mu$, respectively. This {\color{black} occupation measure}  is a
$\gamma$-discounted probability distribution for visiting
each state and action pair, and it plays an important role in sampling states and actions from the real
system in policy gradient and actor-critic algorithms, and in
guaranteeing their convergence. {\color{black} Because the state and action spaces are finite, Theorem 3.1 in \cite{altman1999constrained} shows that the occupation measure $d_\gamma^\mu(x|x^0)$ is a well-defined probability distribution. 
On the other hand, when $\gamma=1$ the occupation measure of state $x$ and state-action pair $(x,a)$ under policy $\mu$ are respectively defined by 
$d^\mu(x|x^0)=\sum_{t=0}^\infty \mathbb P(x_t=x|x^0;\mu)$ and
$\pi^\mu(x,a|x^0)=d^\mu(x|x^0)\mu(a|x)$. In this case the occupation
measures characterize the total sums of visiting probabilities
(although they are not in general probability distributions themselves) of state $x$ and state-action pair $(x,a)$.
{To study the well-posedness of the occupation measure, we define the following notion of a transient MDP.}
\begin{definition}\label{def:transient_mdp}
Define $\X^\prime=\X\setminus\{{\color{black} x_{\text{Tar}}}\}=\{1,\ldots,n\}$ as a state space of transient states. An MDP is said to be transient if,
\begin{enumerate}
\item $\sum_{k=0}^\infty \mathbb P(x_k=x|x^0,\mu)< \infty$ for every $x\in \X^\prime$ and every stationary policy $\mu$,
\item $P({\color{black} x_{\text{Tar}}}|{\color{black} x_{\text{Tar}}},a)=1$ for every admissible control action $a\in\A$.
\end{enumerate}
\end{definition}
} 

%\noindent
%{\bf TODO: The constraint cost function $D$ is not something standard in the ML/RL community. If you want to keep it in the paper, you should explain it more and mention why you may need two different cost functions. Motivate it.} \\
{\color{black} Furthermore let $T_{\mu,x}$ be the first-hitting time of the target state $x_{\text{Tar}}$ from an arbitrary initial state $x\in\mathcal X$ in the Markov chain induced by transition probability $P(\cdot|x,a)$ and policy $\mu$. 
Although transience implies the first-hitting time is square
integrable and finite almost surely, we will make the stronger
assumption (which implies transience) on the uniform boundedness of the first-hitting time. 
\begin{assumption}\label{ass:finite_time}
The first-hitting time $T_{\mu,x}$ is bounded almost surely over all
stationary policies $\mu$ and all initial states $x\in\mathcal X$. We
will refer to this upper bound as $T$, i.e., $T_{\mu,x}\le T$ almost surely.
\end{assumption}
The above assumption can be justified by the fact that sample
trajectories collected in most reinforcement learning algorithms
(including policy gradient and actor-critic methods) consist of {\color{black} bounded}
finite stopping time (also known as a time-out). Note that
although a bounded stopping time would seem to conflict with the
time-stationarity of the transition probabilities, this can be resolved
by augmenting the state space with a time-counter state, analogous to the arguments given in Section 4.7 in \cite{BertsekasDP01}.}

{\color{black} Finally, we define the constraint and cost functions.}
Let $Z$ be a finite-mean ($\E[|Z|]<\infty$) random variable
{\color{black} representing cost,} with the cumulative distribution function ${\color{black} F_Z(z)}=\mathbb{P}(Z\leq z)$ (e.g.,~one may think of $Z$ as the total cost of an investment strategy $\mu$). We define the {\em value-at-risk} at confidence level $\alpha\in (0,1)$ as 
\[ \text{VaR}_\alpha(Z) = \min\big\{z\mid {\color{black} F_Z(z)}\geq\alpha\big\}.\] 
Here the minimum is attained because $F_Z$ is non-decreasing and
right-continuous in $z$. When $F_Z$ is continuous and strictly
increasing, VaR$_\alpha(Z)$ is the unique $z$ satisfying
${\color{black} F_Z(z)}=\alpha$. {\color{black} As mentioned,} we refer to a constraint on the VaR as a {\color{black} chance constraint}.

%The VaR risk measure has many fundamental \emph{engineering applications} such as motion planning, where a safety constraint is imposed to upper bound the probability of maneuvering into dangerous regimes.

Although VaR is a popular risk measure, it is not a {\em coherent}
risk measure~\citep{Artzner99CM} and does not quantify the costs that
might be suffered beyond its value in the $\alpha$-tail of the
distribution~{\color{black}\citep{Rockafellar00OC}, \cite{Rockafellar02CV}}. In many \emph{financial
  applications} such as portfolio optimization where the probability
of undesirable events could be small but the cost incurred could still
be significant, besides describing risk as the probability of
incurring costs, it will be more interesting to study the cost in the
tail of the risk distribution. In this case, an alternative measure
that addresses most of the VaR's shortcomings is the {\em conditional
  value-at-risk}, {\color{black} defined as \citep{Rockafellar00OC}}
%
%To make the constraint more {analytically tractable}, we note that {Theorem~10 of~\cite{Rockafellar00OC}} shows that:
%
\begin{equation}
\label{eq:CVaR2}
\text{CVaR}_\alpha(Z):= \min_{\nu\in\reals}\Big\{\nu + \frac{1}{1-\alpha}\E\big[(Z-\nu)^+\big]\Big\},
\end{equation}
where $(x)^+=\max(x,0)$ represents the positive part of
$x$. {\color{black} While it might not be an immediate observation, it has been shown in Theorem 1 of \citet{Rockafellar00OC} that the CVaR of the loss random variable $Z$ is equal to the average of the worst-case $\alpha$-fraction
  of losses. }
%If there is no probability atom at VaR$_\alpha(Z)$, CVaR$_\alpha(Z)$ has a unique value that is defined as \[ \text{CVaR}_\alpha(Z) = \E\big[Z\mid Z\geq \text{VaR}_\alpha(Z)\big].\]

We define the parameter $\gamma\in(0,1]$ as the \emph{discounting
  factor} for the cost and constraint cost functions. {\color{black}
When $\gamma<1$, we are aiming} to
solve the MDP problem with more focus on optimizing current costs over
future costs. For a policy $\mu$, we define the cost of a state $x$ (state-action pair $(x,a)$) as the sum of (discounted) costs encountered by the {\color{black} decision-maker} when it starts at state $x$ (state-action pair $(x,a)$) and then follows policy $\mu$, i.e.,
\[
{\color{black}\mathcal G}^\theta(x)=\sum_{k=0}^{T-1}\gamma^k C(x_k,a_k)\mid x_0=x,\;{\color{black}\mu(\cdot|\cdot,\theta)}, \quad\quad {\color{black}\mathcal J}^\theta(x)=\sum_{k=0}^{T-1}\gamma^k D(x_k,a_k)\mid x_0=x,\;{\color{black}\mu(\cdot|\cdot,\theta)},
\]
and 
\[
\begin{split}
{\color{black}\mathcal G}^\theta(x,a)=\sum_{k=0}^{T-1}\gamma^k C(x_k,a_k)\mid x_0=x,\;a_0=a,\;{\color{black}\mu(\cdot|\cdot,\theta)},\\
{\color{black}\mathcal J}^\theta(x,a)=\sum_{k=0}^{T-1}\gamma^k D(x_k,a_k)\mid x_0=x,\;a_0=a,\;{\color{black}\mu(\cdot|\cdot,\theta)}.
\end{split}
\] 
{\color{black}The expected values of the random variables ${\color{black}\mathcal G}^\theta(x)$ and ${\color{black}\mathcal G}^\theta(x,a)$ are known as the value and action-value functions of policy $\mu$, and  are denoted by}
\[V^\theta(x)=\E\big[{\color{black}\mathcal G}^\theta(x)\big],\qquad \qquad Q^\theta(x,a)=\E\big[{\color{black}\mathcal G}^\theta(x,a)\big].\] 

\subsection{Problem Statement}
\label{sec:Risk-Opt}
The goal {\color{black} for}  standard discounted MDPs is to find an optimal policy {\color{black} that solves} 
\[
\theta^*=\argmin_\theta V^\theta(x^0).
\]

For {\em CVaR-constrained} optimization in MDPs, we consider the discounted cost optimization problem with $\gamma\in(0,1)$, i.e.,~for a given confidence level $\alpha\in (0,1)$ and cost tolerance $\beta\in\reals$,
\begin{equation}
\label{eq:norm_reward_eqn1}
\min_\theta V^\theta(x^0)\quad\quad \text{{\color{black} subject to}} \quad\quad \text{CVaR}_\alpha\big({\color{black}\mathcal J}^\theta(x^0)\big)\leq\beta.
\end{equation}
 {\color{black} Using the definition of $H_\alpha(Z,\nu)$,} one can reformulate \eqref{eq:norm_reward_eqn1} as:
\begin{equation}
\label{eq:norm_reward_eqn2}
\min_{\theta,\nu} V^\theta(x^0)\quad\quad \text{{\color{black} subject to}} \quad\quad H_\alpha\big({\color{black}\mathcal J}^\theta(x^0),\nu\big)\leq\beta,
\end{equation}
{\color{black} where 
\[
H_\alpha(Z,\nu) := \nu + \frac{1}{1-\alpha}\E\big[(Z-\nu)^+\big].
\]
The equivalence between problem \eqref{eq:norm_reward_eqn1} and problem \eqref{eq:norm_reward_eqn2} can be shown as follows. Let $\theta_2\in\Theta$ be any arbitrary feasible policy parameter of problem \eqref{eq:norm_reward_eqn1}. With $\theta_2$, one can always construct $\nu_2=\text{VaR}_\alpha(\mathcal J^{\theta_2}(x^0))$, such that $(\theta_2,\nu_2)$ is feasible to problem  \eqref{eq:norm_reward_eqn2}. This in turn implies that the solution of \eqref{eq:norm_reward_eqn2} is less than the solution of  \eqref{eq:norm_reward_eqn1}. On the other hand, the following chain of inequalities holds for any $\nu\in\real$: $\text{CVaR}_\alpha\big(\mathcal J^\theta(x^0)\big)\leq H_\alpha\big(\mathcal J^\theta(x^0),\nu\big)\leq\beta$. This implies that the feasible set of $\theta$ in problem \eqref{eq:norm_reward_eqn2} is a subset of the feasible set of $\theta$ in problem \eqref{eq:norm_reward_eqn1}, which further indicates that the solution of problem \eqref{eq:norm_reward_eqn1} is less than the solution of problem \eqref{eq:norm_reward_eqn2}. By combining both arguments, one concludes the equivalence relation of these two problems. }

It is shown in {\color{black} \cite{Rockafellar00OC} and \cite{Rockafellar02CV}} that the optimal $\nu$ actually
equals VaR$_{\alpha}$, so we refer to this parameter as the VaR
parameter. Here we choose to analyze the discounted-cost CVaR-constrained optimization {\color{black} problem}, i.e., {\color{black} with} ~$\gamma\in (0,1)$, {\color{black} as} in many financial and marketing applications where CVaR constraints are  used, it is more intuitive to put more emphasis on current costs rather than {\color{black} on} future costs. The analysis can be {easily} generalized for the case where $\gamma=1$.  

For \emph{{\color{black} chance-constrained}} optimization in MDPs, we consider the stopping cost optimization problem with $\gamma=1$, i.e.,~{\color{black} for a given confidence level $\beta\in (0,1)$ and cost tolerance $\alpha\in\reals$},
\begin{equation}
\label{eq:norm_reward_eqn3}
\min_\theta V^\theta(x^0)\quad\quad \text{{subject to}} \quad\quad \mathbb P\big({\color{black}\mathcal J}^\theta(x^0)\geq \alpha\big)\leq \beta.
\end{equation}
Here we choose $\gamma=1$ because in many engineering applications, where {\color{black} chance constraints} are used to ensure overall safety, there is no notion of discounting since future threats are often as important as the current one. Similarly, the analysis can be {\color{black} easily extended} to the case where $\gamma\in(0,1)$.

There are a number of mild technical and notational assumptions which we will make throughout the paper, so we state them here:
\begin{assumption}[Differentiability]\label{ass:differentiability}
For any state-action pair $(x,a)$, $\mu(a|x;\theta)$ is continuously
differentiable in $\theta$ and $\nabla_\theta\mu(a|x;\theta)$ is a
Lipschitz function in $\theta$ for every $a\in\A$ and $x\in\X$.\footnote{In actor-critic algorithms, the assumption on continuous differentiability holds for {\color{black}the} augmented state Markovian policies $\mu(a|x,s;\theta)$.}
\end{assumption}
\begin{assumption}[Strict Feasibility]\label{ass:feasibility}
There exists a transient policy $\mu(\cdot|x;\theta)$ such that 
\[
H_\alpha\big({\color{black}\mathcal J}^\theta(x^0),\nu\big) < \beta
\]
in the CVaR-constrained optimization problem, {\color{black} and} $P\big({\color{black}\mathcal J}^\theta(x^0)\geq \alpha\big) < \beta$ in the {\color{black} chance-constrained} problem.
\end{assumption}
%
% Note that Assumption~\ref{ass:differentiability} imposes smoothness on the optimal policy. Assumption~\ref{ass:feasibility} guarantees the existence of a {locally optimal policy for the CVaR-constrained optimization problem via the Lagrangian analysis} introduced in the next subsection. 

%Assumption~\ref{ass:steps_ac} refers to step sizes corresponding to policy updates that will be introduced for the algorithms in this paper, and indicates that the update
% corresponding to $\{\zeta_4(k)\}$ is on the fastest time-scale, the
% updates corresponding to $\{\zeta_3(k)\}$, $\{\zeta_2(k)\}$ are on the intermediate time-scale, where $\zeta_3(k)$ converges faster than
% $\zeta_2(k)$, and the update corresponding to $\{\zeta_1(k)\}$ is on the slowest time-scale. Assumption~\ref{ass:basis} refers to basis
% functions for approximating the value function in actor-critic algorithms. As these last two assumptions refer to user-defined parameters, they can always be chosen to be satisfied.

%We have presented two different problem formulations which differ in the {structure of the}  constraints. While both problems are interesting and applicable to many domains, we will later see that the procedure of finding a {\color{black} locally optimal}  policy for {\color{black} chance-constrained} optimization is analogous to the procedure in CVaR-constrained optimization. To simplify the analysis,

In the {\color{black} remainder of the paper} we first focus on studying stochastic approximation algorithms for the CVaR-constrained  optimization problem (Sections \ref{sec:PG-alg} and \ref{sec:AC-alg}) and then {\color{black} adapt} the results to the {\color{black} chance-constrained} optimization problem in Section~\ref{sec:gen_chance}. {\color{black} Our solution approach relies on a Lagrangian relaxation procedure, which is discussed next.}

\subsection{Lagrangian Approach and Reformulation}
To solve~\eqref{eq:norm_reward_eqn2}, we employ a Lagrangian relaxation procedure~{\color{black}(Chapter 3 of \citet{bertsekas1999nonlinear})}, {\color{black} which leads} to the unconstrained problem:  
\begin{equation}
\label{eq:unconstrained-discounted-risk-measure}
\max_{\lambda\geq 0}\min_{\theta,\nu}\bigg(L(\nu,\theta,\lambda):= V^\theta(x^0)+\lambda\Big(H_\alpha\big({\color{black}\mathcal J}^\theta(x^0),\nu\big)-\beta\Big)\bigg),
\end{equation}
where $\lambda$ is the Lagrange multiplier.
%Now similar to the assumption in \citep{Prashanth13AC}, we have the following condition that guarantees the existence
%of a strictly feasible solution.
%
%\begin{assumption}
%There exists a policy $\tilde\theta\in\Theta$ such that $\min_{\nu\in\reals}H_\alpha\big({\color{black}\mathcal J}^{\tilde\theta}(x^0),\nu\big)<\beta$.
%\end{assumption}
%
%The following proposition helps to guarantee the existence of a saddle point solution.
%
%\begin{proposition}\label{prop_lag_lip}
%For any given $\nu$ and $\lambda$, $L(\nu,\theta,\lambda)$ is a Lipschitz function in $\theta$.
%\end{proposition}
%
%\begin{prooff}
%Details of the proof can be found in Appendix \ref{sec:lag_lip_pf}.
%\end{prooff}
Notice that $L(\nu,\theta,\lambda)$ is a linear function in $\lambda$ and $H_\alpha\big({\color{black}\mathcal J}^\theta(x^0),\nu\big)$ is a continuous function in $\nu$. 
{\color{black} The saddle point theorem~{\color{black}from Chapter 3 of \citet{bertsekas1999nonlinear}} states that a local saddle point $(\nu^*,\theta^*,\lambda^*)$  for the maximin optimization problem $\max_{\lambda\geq 0}\min_{\theta,\nu}L(\nu,\theta,\lambda)$ is indeed a locally optimal policy $\theta^*$ for the CVaR-constrained optimization problem. To further explore this connection, we first have the following definition of a saddle point:}
\begin{definition}
A local saddle point of { $L(\nu,\theta,\lambda)$} is a point {\color{black}
  $(\nu^*,\theta^*,\lambda^*)$} such that for some $r>0$,
$\forall(\theta,\nu)\in\Theta\times\left[-\frac{D_{\max}}{1-\gamma},\frac{D_{\max}}{1-\gamma}\right]\cap
\mathcal B_{(\theta^*,\nu^*)}(r)$ and $\forall \lambda\geq 0$, we have
\begin{equation}\label{eq:local_saddle_point}
L(\nu, \theta,\lambda^*) \ge L(\nu^*, \theta^*,\lambda^*) \ge L(\nu^*, \theta^*,\lambda),
\end{equation}
where $\mathcal B_{(\theta^*,\nu^*)}(r)$ is a hyper-dimensional ball centered at $(\theta^*,\nu^*)$ with radius $r>0$.
\end{definition}

In {\color{black}Chapter 7 of \citet{Ott10MD}} and in \cite{Bauerle11MD} it is shown that there exists a {\em deterministic history-dependent} optimal policy for CVaR-constrained optimization. The important point is that this policy does not depend on the complete history, but only on the current time step $k$, current state of the system $x_k$, and accumulated discounted constraint cost $\sum_{i=0}^k\gamma^i  D(x_k,a_k)$.  

In the following {\color{black} two sections}, we present a policy gradient (PG) algorithm (Section~\ref{sec:PG-alg}) and several actor-critic (AC) algorithms (Section~\ref{sec:AC-alg}) to optimize~\eqref{eq:unconstrained-discounted-risk-measure} {\color{black} (and hence find a locally optimal solution to problem \eqref{eq:norm_reward_eqn2})}. While the PG algorithm updates its parameters after observing several trajectories, the AC algorithms are incremental and update their parameters at each time-step.

\section{A Trajectory-based Policy Gradient Algorithm}
\label{sec:PG-alg}

In this section, we present a policy gradient algorithm to solve the optimization problem~\eqref{eq:unconstrained-discounted-risk-measure}. The idea of the algorithm is to descend in $(\theta,\nu)$ and ascend in $\lambda$ using the gradients of $L(\nu,\theta, \lambda)$ w.r.t.~$\theta$, $\nu$, and $\lambda$, i.e.,\footnote{The notation $\ni$ in \eqref{eq:grad-nu} means that the right-most term is a member of the sub-gradient set $\partial_\nu L(\nu,\theta,\lambda)$.}
\begin{align}
\label{eq:grad-theta}
\nabla_\theta L(\nu,\theta,\lambda) &= \nabla_\theta V^\theta(x^0) + \frac{\lambda}{(1-\alpha)} \nabla_\theta\E\Big[\big({\color{black}\mathcal J}^\theta(x^0)- \nu\big)^+\Big], \\
\label{eq:grad-nu}
\partial_\nu L(\nu,\theta,\lambda) &= \lambda\bigg(1 + \frac{1}{(1-\alpha)}\partial_\nu\E\Big[\big({\color{black}\mathcal J}^\theta(x^0)- \nu\big)^+\Big]\bigg) \ni \lambda\bigg(1 - \frac{1}{(1-\alpha)}\mathbb{P}\big({\color{black}\mathcal J}^\theta(x^0) \geq \nu\big)\bigg), \\
\label{eq:grad-lambda}
\nabla_\lambda L(\nu,\theta, \lambda) &= \nu + \frac{1}{(1-\alpha)}\E\Big[\big({\color{black}\mathcal J}^\theta(x^0) - \nu\big)^+\Big] - \beta.
\end{align}

The unit of observation in this algorithm is a {\color{black}trajectory} generated by following the current policy. At each iteration, the algorithm generates $N$ trajectories by following the current policy, uses them to estimate the gradients in \eqref{eq:grad-theta}--\eqref{eq:grad-lambda}, and then uses these estimates to update the parameters $\nu,\theta,\lambda$.

Let $\xi=\{x_0,a_0,c_0,x_1,a_1,c_1,\ldots,x_{T-1},a_{T-1},c_{T-1},x_T\}$ be a trajectory generated by following the policy $\theta$, where $x_T={\color{black} x_{\text{Tar}}}$ is the target state of the system. The cost, constraint cost, and probability of $\xi$ are defined as ${\color{black}\mathcal G}(\xi)=\sum_{k=0}^{T-1}\gamma^k C(x_k,a_k)$, ${\color{black}\mathcal J}(\xi)=\sum_{k=0}^{T-1}\gamma^k D(x_k,a_k)$, and $\mathbb{P}_{\theta}(\xi)=P_0(x_0)\prod_{k=0}^{T-1}\mu(a_k|x_k;\theta)P(x_{k+1}|x_k,a_k)$, respectively. Based on the definition of $\mathbb{P}_{\theta}(\xi)$, one obtains $\nabla_\theta\log\mathbb{P}_{\theta}(\xi)=\sum_{k=0}^{T-1}\nabla_\theta\log\mu(a_k|x_k;\theta)$.

Algorithm~\ref{alg_traj} contains the pseudo-code of our proposed
policy gradient algorithm. What appears inside the parentheses on the
right-hand-side of the update equations are the estimates of the
gradients of $L(\nu,\theta,\lambda)$ w.r.t.~$\theta,\nu,\lambda$
(estimates of \eqref{eq:grad-theta}--\eqref{eq:grad-lambda}). Gradient
estimates of the Lagrangian function can be found in
Appendix~\ref{subsec:grad-comp}. In the algorithm, $\Gamma_\Theta$ is
an operator that projects a vector $\theta\in\reals^{\kappa}$ to the
closest point in a compact and convex set
$\Theta\subset\reals^{\kappa}$, i.e.,
$\Gamma_\Theta(\theta)=\arg\min_{\hat\theta\in\Theta}\|\theta-\hat\theta\|^2_2$,
$\Gamma_{\mathcal{N}}$ is a projection operator to
$[-\frac{D_{\max}}{1-\gamma},\frac{D_{\max}}{1-\gamma}]$, i.e.,
$\Gamma_{\mathcal{N}}(\nu)=\arg\min_{\hat\nu\in[-\frac{D_{\max}}{1-\gamma},\frac{D_{\max}}{1-\gamma}]}\|\nu-\hat\nu\|^2_2$,
and $\Gamma_\Lambda$ is a projection operator to
$[0,\lambda_{\max}]$, i.e.,
$\Gamma_\Lambda(\lambda)=\arg\min_{\hat\lambda\in[0,\lambda_{\max}]}\|\lambda-\hat\lambda\|^2_2$. These
projection operators are necessary to ensure the convergence of the
algorithm{\color{black}; see the end of Appendix~\ref{subsec:conv-proof} for details}. {\color{black} Next we introduce the following assumptions for the step-sizes of the policy gradient method in Algorithm~\ref{alg_traj}.
\begin{assumption}[Step Sizes for Policy Gradient]\label{ass:steps_pg}
The step size schedules $\{\zeta_1(k)\}$, $\{\zeta_2(k)\}$, and $\{\zeta_3(k)\}$ satisfy
\begin{align}
&\sum_k \zeta_1(k) = \sum_k \zeta_2(k) = \sum_k \zeta_3(k) =\infty, \\
&\sum_k \zeta_1(k)^2,\;\;\;\sum_k \zeta_2(k)^2,\;\;\;\sum_k
  \zeta_3(k)^2 < \infty, \\
&\zeta_1(k) = o\big(\zeta_2(k)\big), \;\;\; \zeta_2(k) = o\big(\zeta_3(k)\big).
\end{align}
\end{assumption}}
These step-size schedules satisfy the standard conditions for stochastic approximation algorithms, and ensure that the $\nu$ update is on the fastest time-scale $\big\{\zeta_3(k)\big\}$, the policy $\theta$ update is on the intermediate time-scale $\big\{\zeta_2(k)\big\}$, and the Lagrange multiplier $\lambda$ update is on the slowest time-scale $\big\{\zeta_1(k)\big\}$. This results in a three time-scale stochastic approximation algorithm. 

In the following theorem, we prove that our policy gradient algorithm converges to a {\color{black} locally optimal policy for the CVaR-constrained optimization problem}. 
\begin{theorem}\label{thm:converge_h}
Under Assumptions~\ref{ass:finite_time}--\ref{ass:steps_pg}, the sequence of policy updates in Algorithm \ref{alg_traj} converges almost surely to a {\color{black} locally optimal policy $\theta^*$ for the CVaR-constrained optimization problem} {\color{black} as $k$ goes to infinity}. 
\end{theorem}
While we refer the reader to Appendix~\ref{subsec:conv-proof} for the technical details of this proof, a high level overview of the proof technique is given as follows.  
\begin{enumerate}
\item First we show that each update of the multi-time scale discrete stochastic approximation algorithm $(\nu_k,\theta_k,\lambda_k)$ converges almost surely, but at different speeds, to the stationary point $(\nu^\ast,\theta^*,\lambda^*)$ of the corresponding continuous time system. 
\item Then by using Lyapunov analysis, we show that the continuous time system is locally asymptotically stable at the stationary point $(\nu^\ast,\theta^*,\lambda^*)$. 
\item Since the Lyapunov function used in the above analysis is the Lagrangian function $L(\nu,\theta,\lambda)$, we finally conclude that the stationary point $(\nu^\ast,\theta^*,\lambda^*)$ is also a local saddle point, {\color{black} which by the saddle point theorem (see e.g., Chapter 3 of \citet{bertsekas1999nonlinear}), implies that $\theta^*$ is a locally optimal solution of the CVaR-constrained MDP problem (the primal problem)}.
\end{enumerate}
This convergence proof procedure is standard {\color{black} for} stochastic approximation algorithms, see~\citep{bhatnagar2009natural,bhatnagar2012online,Prashanth13AC} for more details, and {\color{black} represents} the structural backbone {\color{black} for} the convergence analysis of the other policy gradient and actor-critic methods {\color{black} provided} in this paper.

 \begin{algorithm}[t]
\begin{algorithmic}
\STATE {\bf Input:} parameterized policy $\mu(\cdot|\cdot;\theta)$, confidence level $\alpha$, and cost tolerance $\beta$ %and Lagrangian threshold $\lambda_{\max}$
%Let $x^0\in\X$ be an initial state, $\alpha\in(0,1)$ be the given CVaR parameter and $K$ be the given threshold parameter. 
\STATE {\bf Initialization:} policy $\theta=\theta_0$, VaR parameter $\nu=\nu_0$, and the Lagrangian parameter $\lambda=\lambda_0$
%\STATE {\bf Step Length Conditions:} Step lengths $\zeta_{1,k}$, $\zeta_{2,i}$, $\zeta_{3,i}$ are square summable but not summable. Furthermore, $\{\zeta_{1,k}\}$ is on the fastest time-scale, the update corresponds to $\{\zeta_{2,i}\}$ is on the intermediate time-scale, and the update corresponds to $\{\zeta_{3,i}\}$ is on the slowest time-scale. More details can be found in the Appendix.
\WHILE{TRUE}
\FOR{$k=0,1,2,\ldots$}
\STATE Generate $N$ trajectories $\{\xi_{j,k}\}_{j=1}^N$ by starting at $x_0=x^0$ and following the current policy $\theta_k$.
\begin{align*}
\textrm{{\bf $\nu$ Update:}}\quad& \nu_{k+1} = \Gamma_{\mathcal{N}}\bigg[\nu_k - \zeta_3(k)\bigg(\lambda_k - \frac{\lambda_k}{(1-\alpha)N}\sum_{j=1}^N\mathbf{1}\big\{{\color{black}\mathcal J}(\xi_{j,k})\geq\nu_k\big\}\bigg)\bigg] \\
\textrm{\bf $\theta$ Update:} \quad &  \theta_{k+1} = \Gamma_\Theta\bigg[\theta_k -\zeta_2(k)\bigg(\frac{1}{N}\sum_{j=1}^N\nabla_\theta\log\mathbb{P}_\theta(\xi_{j,k})\vert_{\theta=\theta_k}{\color{black}{\color{black}\mathcal G}(\xi_{j,k})} \nonumber \\ 
&\hspace{0.3in}+ \frac{\lambda_k}{(1-\alpha)N}\sum_{j=1}^N\nabla_\theta\log\mathbb{P}_\theta(\xi_{j,k})\vert_{\theta=\theta_k}\big({\color{black}\mathcal J}(\xi_{j,k})-\nu_k\big)\mathbf{1}\big\{{\color{black}\mathcal J}(\xi_{j,k})\geq\nu_k\big\}\bigg)\bigg] \\
\textrm{{\bf $\lambda$ Update:}}\quad&\lambda_{k+1} = \Gamma_\Lambda\bigg[\lambda_k + \zeta_1(k)\bigg(\nu_k - \beta + \frac{1}{(1-\alpha)N}\sum_{j=1}^N\big({\color{black}\mathcal J}(\xi_{j,k})-\nu_k\big)\mathbf{1}\big\{{\color{black}\mathcal J}(\xi_{j,k})\geq\nu_k\big\}\bigg)\bigg]
\end{align*}
\ENDFOR  
\IF{$\{\lambda_k\}$ converges to $\lambda_{\max}$, {\color{black} i.e., $|\lambda_{i^*}-\lambda_{\max}|\leq \epsilon$ for some tolerance parameter $\epsilon>0$}}
\STATE{Set $\lambda_{\max}\leftarrow 2\lambda_{\max}$.}
\ELSE
\STATE {\bf return} parameters $\nu,\theta,\lambda$ and {\bf break}
\ENDIF
\ENDWHILE
\end{algorithmic}
%\vspace{-0.1in}
\caption{Trajectory-based Policy Gradient Algorithm for CVaR MDP}
\label{alg_traj}
%\vspace{-0.1in}
\end{algorithm}

Notice that the difference in convergence speeds between $\theta_k$, $\nu_k$, and $\lambda_k$ is due to the step-size schedules. Here  $\nu$ converges faster than $\theta$ and $\theta$ converges faster than $\lambda$. This multi-time scale convergence property allows us to simplify the convergence analysis by assuming that $\theta$ and $\lambda$ are fixed in $\nu$'s convergence analysis, assuming that $\nu$ converges to $\nu^*(\theta)$ and $\lambda$ is fixed in $\theta$'s convergence analysis, and finally assuming that $\nu$ and $\theta$ have already converged to $\nu^*(\lambda)$ and $\theta^*(\lambda)$ in $\lambda$'s convergence analysis. To illustrate this idea, consider the following two-time scale stochastic approximation algorithm for updating $(x_k,y_k)\in \mathbf X\times \mathbf Y$:
{\color{black} \begin{align}
\label{eq:time-scale1}
x_{k+1}&=x_k+\zeta_1(k)\big(f(x_k,y_k)+M_{k+1}\big), \\
y_{k+1}&=y_k+\zeta_2(k)\big(g(x_k,y_k)+N_{k+1}\big),
\label{eq:time-scale2}
\end{align}
where $f(x_k,y_k)$ and $g(x_k,y_k)$ are Lipschitz continuous functions, $M_{k+1}$, $N_{k+1}$ are square integrable Martingale differences w.r.t.~the $\sigma$-fields $\sigma(x_i,y_i,M_i,i\leq k)$ and $\sigma(x_i,y_i,N_i,i\leq k)$, and $\zeta_1(k)$ and $\zeta_2(k)$ are non-summable, square summable step sizes. If $\zeta_2(k)$ converges to zero faster than $\zeta_1(k)$, then~\eqref{eq:time-scale1} is a faster recursion than~\eqref{eq:time-scale2} after some iteration $k_0$ (i.e.,~for $k\geq k_0$), which means~\eqref{eq:time-scale1} has uniformly larger increments than~\eqref{eq:time-scale2}. Since~\eqref{eq:time-scale2} can be written as 
\begin{equation*}
y_{k+1}=y_k+\zeta_1(k)\Big(\frac{\zeta_2(k)}{\zeta_1(k)}\big(g(x_k,y_k)+N_{k+1}\big)\Big), 
\end{equation*}
and by the fact that $\zeta_2(k)$ converges to zero faster than
$\zeta_1(k)$,~\eqref{eq:time-scale1} and~\eqref{eq:time-scale2} can be
viewed as noisy Euler discretizations of the ODEs $\dot{x}=f(x,y)$ and
$\dot{y}=0$. Note that one can consider the ODE $\dot{x}=f(x,y_0)$ in
place of $\dot{x}=f(x,y)$, where  $y_0$ is constant, because
$\dot{y}=0$. One can then show (see e.g.,~{\color{black}Theorem 2 in Chapter 6 of \citet{borkar2008stochastic}})
the main two-timescale convergence result, i.e., under the above
assumptions associated with \eqref{eq:time-scale2}, the sequence
$(x_k,y_k)$ converges to $\big(\mu(y^\star),y^\star\big)$ as
$i\rightarrow\infty$, with probability one, where $\mu(y_0)$ is a
{\color{black} locally} asymptotically stable equilibrium of the ODE
$\dot{x}=f(x,y_0)$, $\mu$ is a Lipschitz continuous function, and
$y^\star$ is a {\color{black} locally} asymptotically stable equilibrium of the ODE
$\dot{y}=g\big(\mu(y),y\big)$.}

\section{Actor-Critic Algorithms}
\label{sec:AC-alg}
As mentioned in Section~\ref{sec:PG-alg}, the unit of observation in
our policy gradient algorithm (Algorithm~\ref{alg_traj}) is a system
trajectory. This may result in high variance for the gradient
estimates, especially when the length of the trajectories is long. To
address this issue, in this section we propose two actor-critic
algorithms that {\color{black} approximate} some quantities in the
gradient estimates {\color{black} by linear combinations of basis
  functions} and update the parameters {\color{black} (linear coefficients)} incrementally (after each
state-action transition). We present two actor-critic algorithms for
optimizing ~\eqref{eq:unconstrained-discounted-risk-measure}. These
algorithms are based on the gradient estimates of
Sections~\ref{subsec:grad-theta}-\ref{subsec:grad-lambda_nu}. While
the first algorithm (SPSA-based) is fully incremental and updates all
the parameters $\theta,\nu,\lambda$ at each time-step, the second one
updates $\theta$ at each time-step and updates $\nu$ and $\lambda$
only at the end of each trajectory, thus is regarded as a semi-trajectory-based method. Algorithm~\ref{alg:AC} contains the pseudo-code of these
algorithms. The projection operators $\Gamma_\Theta$, $\Gamma_{\mathcal{N}}$, and
$\Gamma_\Lambda$ are defined as in Section~\ref{sec:PG-alg} and are
necessary to ensure the convergence of the algorithms. 
{\color{black}
At each step of our actor critic algorithms (steps indexed by $k$ in
Algorithm~\ref{alg_traj} and in Algorithm~\ref{alg:AC}) there are two parts:
\begin{itemize}
\item {\bf Inner loop (critic update)}: For a fixed policy (given as $\theta\in\Theta$), take action $a_k \sim \mu(\cdot|x_k,s_k; \theta_k)$, observe the cost
$c(x_k, a_k)$, the constraint cost $d(x_k,a_k)$, and the next state $(x_{k+1},s_{k+1})$. Using the method of temporal differences (TD) from Chapter 6 of \citet{sutton1998introduction}, estimate the value function $V^\theta(x, s)$.
\item {\bf Outer loop (actor update)}:
Estimate the gradient of $V^\theta(x, s)$ for policy parameter $\theta$, and hence the gradient
of the Lagrangian $L(\nu,\theta,\lambda)$, using the unbiased
sampling based point estimator for gradients with respect to $\theta$
and $\lambda$ and either: (1) using the SPSA method
\eqref{nu_up_kncre_SPSA} to obtain an incremental estimator for
gradient with respect to $\nu$ or (2) only calculating the gradient estimator with respect to
$\nu$ at the end of the trajectory (see \eqref{nu_up_kncre_semi_traj} for more details). Update the policy
parameter $\theta\in\Theta$ in the descent direction, the VaR approximation $\nu\in\mathcal N$ in the descent direction, and the Lagrange
multiplier $\lambda\in\Lambda$ in the ascent direction on specific timescales that ensure convergence to locally optimal solutions.
\end{itemize}}

{\color{black} Next, we introduce the following assumptions for the step-sizes of the actor-critic method in Algorithm~\ref{alg:AC}.}
{\color{black} \begin{assumption}[Step Sizes]\label{ass:steps_ac}
The step size schedules $\{\zeta_1(k)\}$, $\{\zeta_2(k)\}$, $\{\zeta_3(k)\}$, and $\{\zeta_4(k)\}$ satisfy
\begin{align}
\label{eq:step1_kncre}
&\sum_k \zeta_1(k) = \sum_k \zeta_2(k) = \sum_k \zeta_3(k) = \sum_k \zeta_4(k)=\infty, \\
\label{eq:step2_kncre}
&\sum_k \zeta_1(k)^2,\;\;\;\sum_k \zeta_2(k)^2,\;\;\;\sum_k \zeta_3(k)^2,\;\;\;\sum_k \zeta_4(k)^2<\infty, \\
\label{eq:step3_kncre}
&\zeta_1(k) = o\big(\zeta_2(k)\big), \;\;\; \zeta_2(k) = o\big(\zeta_3(k)\big),\;\;\; \zeta_3(k) = o\big(\zeta_4(k)\big).
\end{align}
Furthermore, the SPSA step size $\{\Delta_k\}$ in the actor-critic algorithm satisfies $\Delta_k\rightarrow 0$ as $k\rightarrow\infty$ and $\sum_k (\zeta_2(k)/\Delta_k)^2<\infty$.
\end{assumption}}
{\color{black} These step-size schedules} satisfy the standard conditions for stochastic approximation algorithms, and ensure that the critic update is on the fastest time-scale $\big\{\zeta_4(k)\big\}$, the policy and VaR parameter updates are on the intermediate time-scale, with the $\nu$-update $\big\{\zeta_3(k)\big\}$ being faster than the $\theta$-update $\big\{\zeta_2(k)\big\}$, and finally the Lagrange multiplier update is on the slowest time-scale $\big\{\zeta_1(k)\big\}$. This results in four time-scale stochastic approximation algorithms. 

\begin{algorithm}
\begin{small}
\begin{algorithmic}[h!]
\STATE {\bf Input:} Parameterized policy $\mu(\cdot|\cdot;\theta)$ and value function feature vector $\phi(\cdot)$ (both over the augmented MDP $\bar{\mathcal{M}}$), confidence level $\alpha$, and cost tolerance $\beta$
\STATE {\bf Initialization:} policy $\theta=\theta_0$; VaR parameter $\nu=\nu_0$; Lagrangian parameter $\lambda=\lambda_0$; value function weight vector $v=v_0$ ; initial condition $(x_{0},s_{0})=(x^0,\nu)$
\WHILE{TRUE}
\STATE \textbf{// (1) SPSA-based Algorithm:}
\FOR{$k = 0,1,2,\ldots$}
%\STATE Simulate the augmented states and action $(x^t,s^t,a^t)$ with the current policy $\theta_t$ where $s_t=(s_{t-1}-C(x_{t-1},a_{t-1}))/\gamma$ and $s_0=\nu_t$. 
\STATE Draw action $\;a_k\sim\mu(\cdot|x_k,s_k;\theta_k)$; $\quad\quad\quad\quad\quad\quad\quad$ Observe cost $\;\bar{C}_{\lambda_k}(x_k,s_k,a_k)$;
\STATE Observe next state $(x_{k+1},s_{k+1})\sim \bar{P}(\cdot|x_k,s_k,a_k)$; $\;$ \begin{footnotesize}{\em // note that $s_{k+1}=(s_k-D\big(x_k,a_k)\big)/\gamma\;$}\end{footnotesize}
\STATE \textbf{// AC Algorithm:}
\vspace{-0.1in}
\begin{align}
\textrm{\bf TD Error:} \quad & \delta_k(v_k) = \bar{C}_{\lambda_k}(x_k,s_k,a_k) + \gamma v_k^\top\phi(x_{k+1},s_{k+1}) - v_k^\top\phi(x_k,s_k) \label{TD-calc} \\
\textrm{\bf Critic Update:} \quad & v_{k+1}=v_k+\zeta_4(k)\delta_k(v_k)\phi(x_k,s_k) \label{v_up_kncre} \\
\textrm{{\bf $\nu$ Update:}}\quad & \nu_{k+1} = \Gamma_{\mathcal{N}}\left(\nu_k \!-\! \zeta_3(k)\Big(\lambda_k \!+\! \frac{v_k^\top\big[\phi\big(x^0,\nu_k+\Delta_k\big)- \phi(x^0,\nu_k-\Delta_k)\big]}{2\Delta_k}\Big)\!\!\right) \label{nu_up_kncre_SPSA} \\
\textrm{{\bf $\theta$ Update:}}\quad &\theta_{k+1} = \Gamma_\Theta\Big(\theta_k-\frac{\zeta_2(k)}{1-\gamma}\nabla_\theta\log\mu(a_k|x_k,s_k;\theta)\cdot\delta_k(v_k)\Big) \label{theta_up_kncre} \\
\textrm{{\bf $\lambda$ Update:}}\quad &\lambda_{k+1} = \Gamma_\Lambda\Big(\lambda_k + \zeta_1(k)\big(\nu_k \!-\! \beta + \frac{1}{(1-\alpha)(1-\gamma)}\mathbf 1 \{x_k={\color{black} x_{\text{Tar}}}\}(-s_k)^+\big)\Big) \label{lambda_up_kncre}
\end{align}
\vspace{-0.2in}
%\STATE \textbf{// NAC Algorithm:}
%\vspace{-0.1in}
%\begin{align}
%\textrm{{\bf Critic Update:}}\quad &w_{k+1} = \left(I-\zeta_4(k)\nabla_\theta \log\mu(a_k|x_k,s_k;\theta)\vert_{\theta=\theta_k}\left(\nabla_\theta \log\mu(a_k|x_k,s_k;\theta)\vert_{\theta=\theta_k}\right)^\top\right)w_k\nonumber\\
%&\quad\quad\,\,+\zeta_4(k)\delta_k(v_k)\nabla_\theta \log\mu(a_k|x_k,s_k;\theta)\vert_{\theta=\theta_k} \label{w_NAC_up_kncre} \\
%\textrm{{\bf $\theta$ Update:}}\quad &\theta_{k+1} =  \Gamma_\Theta\Big(\theta_k-\frac{\zeta_2(k)}{1-\gamma}w_k\Big) \label{theta_NAC_up_kncre} \\
%\textrm{{\bf Other Updates:}}\quad & \textrm{Follow from Eqs.~\ref{TD-calc},~\ref{v_up_kncre},~\ref{nu_up_kncre_SPSA}, and~\ref{lambda_up_kncre}.}\nonumber
%\end{align}
%\vspace{-0.3in}
\STATE {\bf if } $x_k= {\color{black} x_{\text{Tar}}}\;$ (reach a target state), $\;\;${\bf then } set $\;(x_{k+1},s_{k+1})=(x^0,\nu_{k+1})$
\ENDFOR
%\STATE
\STATE \textbf{// (2) Semi Trajectory-based Algorithm:}
\STATE{Initialize $t=0$}
\FOR{$k = 0,1,2,\ldots$}
\STATE Draw action $a_k\sim\mu(\cdot|x_k,s_k;\theta_k)$, observe cost $\bar{C}_{\lambda_k}(x_k,s_k,a_k)$, and next state $(x_{k+1},s_{k+1})\sim \bar{P}(\cdot|x_k,s_k,a_k)$; $\;\;$Update $(\delta_k,v_k,\theta_k,\lambda_k)$ using Eqs.~\eqref{TD-calc},~\eqref{v_up_kncre},~\eqref{theta_up_kncre}, and~\eqref{lambda_up_kncre} %or update $(\delta_k,v_k,w_k,\theta_k,\lambda_k)$ using Eqs.~\ref{TD-calc},~\ref{v_up_kncre},~\ref{lambda_up_kncre},~\ref{w_NAC_up_kncre} and~\ref{theta_NAC_up_kncre}.
\IF{$x_k= {\color{black} x_{\text{Tar}}}$} 
\STATE Update $\nu$ as 
\vspace{-0.1in}
\begin{align}
\textrm{{\bf $\nu$ Update:}}\quad&\nu_{k+1}=\Gamma_{\mathcal{N}}\left(\nu_k-\zeta_3(k)\Big(\lambda_k-\frac{\lambda_k}{1-\alpha}\mathbf 1\big\{x_k={\color{black} x_{\text{Tar}}},s_k\leq 0\big\}\Big)\right) \label{nu_up_kncre_semi_traj} 
\end{align}
\vspace{-0.1in}
\STATE Set $\;(x_{k+1},s_{k+1})=(x^0,\nu_{k+1})$ and $t=0$
\ELSE
\STATE $t\leftarrow t+1$
\ENDIF
\ENDFOR 
\IF{$\{\lambda_k\}$ converges to $\lambda_{\max}$, {\color{black} i.e., $|\lambda_{i^*}-\lambda_{\max}|\leq \epsilon$ for some tolerance parameter $\epsilon>0$}}
\STATE{Set $\lambda_{\max}\leftarrow 2\lambda_{\max}$.}
\ELSE
\STATE {\bf return} parameters $v,w,\nu,\theta,\lambda$, and {\bf break}
\ENDIF
\ENDWHILE
\end{algorithmic}
\end{small}
\caption{Actor-Critic Algorithms for CVaR MDP}
\label{alg:AC}
\end{algorithm}

%%%%%%%%%%%%%%%%%%%%%%%%%%%%%%%%%%%%%%%%%%%%%%%%%%%%%%%%%%%%%%
%%%%%%%%%%%%%%%%%%%%%%%%%%%%%%%%%%%%%%%%%%%%%%%%%%%%%%%%%%%%%%
%%%%%%%%%%%%%%%%%%%%%%%%%%%%%%%%%%%%%%%%%%%%%%%%%%%%%%%%%%%%%%

\subsection{Gradient w.r.t.~the Policy Parameters $\theta$}
\label{subsec:grad-theta}

The gradient of the objective function w.r.t.~the policy $\theta$ in \eqref{eq:grad-theta} may be rewritten as
\begin{equation}
\label{eq:grad-theta1}
\nabla_\theta L(\nu,\theta,\lambda) = \nabla_\theta\left(\E\big[{\color{black}\mathcal G}^\theta(x^0)\big] + \frac{\lambda}{(1-\alpha)}\E\Big[\big({\color{black}\mathcal J}^\theta(x^0)- \nu\big)^+\Big]\right).
\end{equation}
Given the original MDP $\mathcal{M}=(\X,\A,C,D,P,P_0)$ and the parameter $\lambda$, we define the augmented MDP $\bar{\mathcal{M}}=(\bar{\X},\bar{\A},\bar{C}_\lambda,\bar{P},\bar{P}_0)$ as $\bar{\X}=\X\times\mathcal S$, $\bar{\A}=\A$, $\bar{P}_0(x,s)=P_0(x)\mathbf{1}\{s_0=s\}$, and
\begin{alignat*}{1}
\bar{C}_{\lambda}(x,s,a)&=\left\{\begin{array}{cc}
\lambda(-s)^+/ (1-\alpha)& \text{if $x={\color{black} x_{\text{Tar}}}$,}\\
C(x,a)& \text{otherwise,}
\end{array}\right.\\
\bar{P}(x^\prime,s^\prime|x,s,a)&=\left\{\begin{array}{ll}
\P(x^\prime|x,a)\mathbf 1\{s^\prime=\big(s-D(x,a)\big)/\gamma\}&\text{if $x\in\mathcal X'$,}\\
\mathbf 1\{x'={\color{black} x_{\text{Tar}}},s'=0\}&\text{if $x={\color{black} x_{\text{Tar}}}$,}\end{array}\right.
\end{alignat*}
where ${\color{black} x_{\text{Tar}}}$ is the target state of the
original MDP $\mathcal{M}$, {\color{black} $\mathcal S$ and $s_0$ are
  respectively the finite state space and the initial state of the $s$
  part of the state in the augmented MDP $\bar{\mathcal{M}}$}. {\color{black}Furthermore, we denote by $s_{\text{Tar}}$
the $s$ part of {\color{black}the state in $\bar{\mathcal{M}}$} when a policy $\theta$ reaches a target
state $x_{\text{Tar}}$ (which we
  assume occurs before an upper-bound $T$ number of steps), i.e., 
\[
s_{\text{Tar}}=\frac{1}{\gamma^T}\left(\nu-\sum_{k=0}^{T-1}\gamma^kD(x_k,a_k)\right),
\] 
such that the initial state is given by
  $s_0=\nu$}.
  {\color{black} We will now use $n=|\bar\X|$ to indicate
    the size of the \emph{augmented} state space $\bar{\X}$ instead of the
    size of the original state space $\X$.} {\color{black} It can be later seen that the augmented
  state $s$ in the MDP $\bar{\mathcal{M}}$ keeps track of the
  cumulative CVaR constraint cost.}
  {\color{black} Similar to the analysis in \cite{Bauerle11MD}, the major motivation of introducing the aforementioned augmented MDP $\bar{\mathcal{M}}$ is that, by utilizing the augmented state $s\in\mathcal S$ that monitors the running constraint cost and thus the feasibility region of the original CVaR constrained MDP, one is able to define a Bellman operator on $\bar{\mathcal{M}}$ (whose exact definition can be found in Theorem \ref{thm:converge_kncre_v}), whose fixed point solution is equal to the solution of the original CVaR Lagrangian problem. Therefore by combining these properties, this reformulation allows one to transform the CVaR Lagrangian problem to a standard MDP problem.} 
  
We define a class of parameterized stochastic policies $\big\{\mu(\cdot|x,s;\theta),(x,s)\in\bar{\X},\theta\in\Theta\subseteq\R^{\kappa_1}\big\}$ for this augmented MDP. {\color{black} Recall that  ${\color{black}\mathcal G}^\theta(x)$ is the discounted cumulative cost and ${\color{black}\mathcal J}^\theta(x)$ is the discounted cumulative constraint cost.} Therefore, the total (discounted) cost of {\color{black} a} trajectory can be written as
\begin{equation}
\label{eq:aug-loss}
\sum_{k=0}^{T}\gamma^k\bar{C}_\lambda(x_k,s_k,a_k)\mid x_0=x,s_0=s,\mu \;\;=\;\; {\color{black}\mathcal G}^\theta(x) + \frac{\lambda}{(1-\alpha)}\big({\color{black}\mathcal J}^\theta(x)-s\big)^+.
\end{equation}
From~\eqref{eq:aug-loss}, it is clear that the quantity in the parenthesis of \eqref{eq:grad-theta1} is the value function of the policy $\theta$ at state $(x^0,\nu)$ in the augmented MDP $\bar{\mathcal{M}}$, i.e.,~$V^\theta(x^0,\nu)$. Thus, it is easy to show that\footnote{Note that the second equality in Equation~\eqref{eq:PG-Thm} is the result of the policy gradient theorem~\citep{Sutton00PG,Peters05NA}.}
\begin{equation}
\label{eq:PG-Thm}
\nabla_\theta L(\nu,\theta,\lambda) = \nabla_\theta V^\theta(x^0,\nu) = \frac{1}{1-\gamma}\sum_{x,s,a}\pi_\gamma^\theta(x,s,a|x^0,\nu)\;\nabla\log\mu(a|x,s;\theta)\;Q^\theta(x,s,a), \footnote{{\color{black} Notice that the state and action spaces of the original MDP are finite, and there is only a finite number of outcomes in the transition of $s$ (due to the assumption of a bounded first hitting time). Therefore the augmented state $s$ belongs to a finite state space as well. }}
\end{equation}
where $\pi_\gamma^\theta$ is the discounted {\color{black} occupation measure}  (defined in Section~\ref{sec:preliminaries}) and $Q^\theta$ is the action-value function of policy $\theta$ in the augmented MDP $\bar{\mathcal{M}}$. We can show that $\frac{1}{1-\gamma}\nabla\log\mu(a_k|x_k,s_k;\theta)\cdot\delta_k$ is an unbiased estimate of $\nabla_\theta L(\nu,\theta,\lambda)$, where 
\[
\delta_k=\bar{C}_{\lambda}(x_k,s_k,a_k)+\gamma\widehat{V}(x_{k+1},s_{k+1})-\widehat{V}(x_k,s_k)
\]
is the temporal-difference (TD) error in the MDP $\bar{\mathcal{M}}$ {\color{black} from~(\ref{TD-calc})}, and $\widehat{V}$ is an unbiased estimator of $V^\theta$ (see e.g.,~\citet{bhatnagar2009natural}). In our actor-critic algorithms, the critic uses linear approximation for the value function $V^\theta(x,s)\approx v^\top\phi(x,s)=\widetilde{V}^{\theta,v}(x,s)$, where the feature vector $\phi(\cdot)$ belongs to a low-dimensional space $\reals^{\kappa_1}$ with dimension $\kappa_1$. The linear approximation $\widetilde{V}^{\theta,v}$ belongs to a low-dimensional subspace
$S_{V}=\left\{\Phi v|v\in\reals^{\kappa_1}\right\}$, {\color{black}
  where $\Phi$ is the $n\times\kappa_1$ matrix whose rows are the
  transposed feature vectors $\phi^\top(\cdot)$. To ensure that the
  set of feature vectors forms a well-posed linear approximation to
  the value function, we impose the following assumption {\color{black}on} the basis functions.
 \begin{assumption}[Independent Basis Functions]\label{ass:basis}
The basis functions $\big\{\phi^{(i)}\big\}_{i=1}^{\kappa_1}$ are
linearly independent. In particular, $\kappa_1 \le n$ and $\Phi$ is
full column rank. Moreover, for every $v\in\reals^{\kappa_1}$, $\Phi v\neq e$, where $e$ is the $n$-dimensional vector with all entries equal to one.
\end{assumption}}
The following theorem shows that the critic update $v_k$ converges almost surely to $v^*$, the minimizer of the Bellman residual. Details of the proof can be found in Appendix \ref{subsec:convergence-proof-AC}.
\begin{theorem}\label{thm:converge_kncre_v}
Define $v^*\in\arg\min_{v}\|B_\theta[\Phi v]-\Phi v\|_{d^\theta_\gamma}^2$ as the minimizer to the Bellman residual, where the Bellman operator is given by
\[
B_\theta[V](x,s)=\sum_{a}\mu(a|x,s;\theta)\left\{\bar{C}_\lambda(x,s,a)+\sum_{x^\prime,s^\prime}\gamma\bar{P}(x^\prime,s^\prime|x,s,a)V(x^\prime,s^\prime)\right\}
\]
and $\tilde V^*(x,s)=(v^*)^\top\phi(x,s)$ is the projected Bellman fixed point of $ V^\theta(x,s)$, i.e., $\tilde V^*(x,s)=\Pi B_\theta[\tilde V^*](x,s)$. Suppose the $\gamma$-{\color{black} occupation measure} $\pi_{\gamma}^\theta$ is used to generate samples of $(x_k,s_k,a_k)$ for any $k\in\{0,1,\ldots,\}$. Then under Assumptions~\ref{ass:steps_ac}--\ref{ass:basis}, the $v$-update in the actor-critic algorithm converges to $v^*$ almost surely.
\end{theorem}

\subsection{Gradient w.r.t.~the Lagrangian Parameter $\lambda$}% and VaR Parameter $\nu$}
\label{subsec:grad-lambda_nu}
We may rewrite the gradient of the objective function w.r.t.~the Lagrangian parameters $\lambda$ in \eqref{eq:grad-lambda} as
\begin{equation}
\label{eq:grad-lambda1}
\nabla_\lambda L(\nu,\theta,\lambda) = \nu - \beta + \nabla_\lambda\left(\E\big[{\color{black}\mathcal G}^\theta(x^0)\big] + \frac{\lambda}{(1-\alpha)}\E\Big[\big({\color{black}\mathcal J}^\theta(x^0)- \nu\big)^+\Big]\right)\stackrel{\text{(a)}}{=} \nu - \beta + \nabla_\lambda V^\theta(x^0,\nu).
\end{equation}
Similar to Section~\ref{subsec:grad-theta}, equality {(a)} comes from the fact that the quantity in parenthesis in \eqref{eq:grad-lambda1} is $V^\theta(x^0,\nu)$, the value function of the policy $\theta$ at state $(x^0,\nu)$ in the augmented MDP $\bar{\mathcal{M}}$. Note that the dependence of $V^\theta(x^0,\nu)$ on $\lambda$ comes from the definition of the cost function $\bar{C}_{\lambda}$ in $\bar{\mathcal{M}}$. We now derive an expression for $\nabla_\lambda V^\theta(x^0,\nu)$, which in turn will give us an expression for $\nabla_\lambda L(\nu,\theta,\lambda)$.
\begin{lemma}
\label{lem:grad_lambda}
The gradient of $V^\theta(x^0,\nu)$ w.r.t.~the Lagrangian parameter $\lambda$ may be written as
\begin{equation}
\label{eq:grad-lambda-V}
\nabla_\lambda V^\theta(x^0,\nu) = \frac{1}{1-\gamma}\sum_{x,s,a}\pi_\gamma^\theta(x,s,a|x^0,\nu)\frac{1}{(1-\alpha)}\mathbf{1}\{x={\color{black} x_{\text{Tar}}}\}(-s)^+.
\end{equation}
\end{lemma}
\begin{prooff}
See Appendix~\ref{subsec:grad-lambda-comp}.
\end{prooff}

From Lemma~\ref{lem:grad_lambda} and \eqref{eq:grad-lambda1}, it is easy to see that $\nu-\beta+\frac{1}{(1-\gamma)(1-\alpha)}\mathbf{1}\{x={\color{black} x_{\text{Tar}}}\}(-s)^+$ is an unbiased estimate of $\nabla_\lambda L(\nu,\theta,\lambda)$. An issue with this estimator is that its value is fixed to $\nu_k-\beta$ all along a {\color{black} trajectory}, and only changes at the end to $\nu_k-\beta+\frac{1}{(1-\gamma)(1-\alpha)}(-{\color{black} s_{\text{Tar}}})^+$. This may affect the incremental nature of our actor-critic algorithm. To address this issue, {\color{black} \citet{chow2014algorithms} previously proposed a different approach to estimate the gradients w.r.t.~$\theta$ and $\lambda$} which involves another value function approximation to the constraint. However this approach is less desirable in many practical applications as it increases the approximation error and impedes the speed of convergence.

Another important issue is that the above estimator is unbiased only if the samples are generated from the distribution $\pi_\gamma^\theta(\cdot|x^0,\nu)$. If we just follow the {\color{black} policy $\theta$}, then we may use $\nu_k-\beta+\frac{\gamma^k}{(1-\alpha)}\mathbf{1}\{x_k={\color{black} x_{\text{Tar}}}\}(-s_k)^+$ as an estimate for $\nabla_\lambda L(\nu,\theta,\lambda)$. Note that this is an issue for all discounted actor-critic algorithms: their (likelihood ratio based) estimate for the gradient is unbiased only if the samples are generated from $\pi_\gamma^\theta$, and not when we simply follow the policy. %Although this issue was known in the community, there is a recent paper that investigates it in details~\citep{Thomas14BN}. Moreover, this might be a main reason that we have no convergence analysis (to the best of our knowledge) for (likelihood ratio based) discounted actor-critic algorithms.\footnote{Note that the discounted actor-critic algorithm with convergence proof in~\citep{Bhatnagar10AC} is based on SPSA.} 
This might also be the reason why, to the best of our knowledge, no rigorous convergence analysis can be found in the literature for (likelihood ratio based) discounted actor-critic algorithms under the sampling distribution.\footnote{Note that the discounted actor-critic algorithm with convergence proof in~\citep{Bhatnagar10AC} is based on SPSA.}

\vspace{-0.1in}
\subsection{Sub-Gradient w.r.t.~the VaR Parameter $\nu$}
\label{subsec:grad-lambda_nu}
\vspace{-0.05in}
We may rewrite the sub-gradient of our objective function w.r.t.~the
VaR parameter $\nu$ in~\eqref{eq:grad-nu} as
\begin{equation}
\label{eq:grad-nu1}
\partial_\nu L(\nu,\theta,\lambda) \ni \lambda\bigg(1-\frac{1}{(1-\alpha)}\mathbb{P}\Big(\sum_{k=0}^\infty\gamma^kD(x_k,a_k)\geq\nu\mid x_0=x^0;\theta\Big)\bigg).
\end{equation}
From the definition of the augmented MDP $\bar{\mathcal{M}}$, the probability in \eqref{eq:grad-nu1} may be written as $\mathbb{P}({\color{black}s_{\text{Tar}}}\leq 0\mid x_0=x^0,s_0=\nu;\theta)$, where ${\color{black}s_{\text{Tar}}}$ is the $s$ part of the state in $\bar{\mathcal{M}}$ when we reach a target state, i.e.,~$x={\color{black} x_{\text{Tar}}}$ (see Section~\ref{subsec:grad-theta}). Thus, we may rewrite \eqref{eq:grad-nu1} as 
\begin{equation}
\label{eq:grad-nu2}
\partial_\nu L(\nu,\theta,\lambda) \ni \lambda\Big(1-\frac{1}{(1-\alpha)}\mathbb{P}\big({\color{black}s_{\text{Tar}}}\leq 0\mid x_0=x^0,s_0=\nu;\theta\big)\Big).
\end{equation}
From \eqref{eq:grad-nu2}, it is easy to see that $\lambda-\lambda\mathbf 1\{{\color{black} s_{\text{Tar}}}\leq 0\}/(1-\alpha)$ is an unbiased estimate of the sub-gradient of $L(\nu,\theta,\lambda)$ w.r.t.~$\nu$. An issue with this (unbiased) estimator is that it can only be applied at the end of a {\color{black}trajectory} (i.e.,~when we reach the target state ${\color{black} x_{\text{Tar}}}$), and thus, using it prevents us from having a fully incremental algorithm. In fact, this is the estimator that we use in our {\em semi-trajectory-based} actor-critic algorithm. 

One approach to estimate this sub-gradient incrementally is to use the {\em simultaneous perturbation stochastic approximation} (SPSA) method~{\color{black}(Chapter 5 of \citet{Bhatnagar13SR})}. The idea of SPSA is to estimate the sub-gradient $g(\nu)\in\partial_{\nu}  L(\nu,\theta,\lambda)$ using two values of $g$ at $\nu^-=\nu-\Delta$ and $\nu^+=\nu+\Delta$, where $\Delta>0$ is a positive perturbation (see~{\color{black}Chapter 5 of \citet{Bhatnagar13SR}} or \citet{Prashanth13AC} for the detailed description of $\Delta$).\footnote{SPSA-based gradient estimate was first proposed in~\citet{Spall92MS} and has been widely used in various settings, especially those involving a high-dimensional parameter. The SPSA estimate described above is two-sided. It can also be implemented single-sided, where we use the values of the function at $\nu$ and $\nu^+$. We refer the readers to~{\color{black}Chapter 5 of \citet{Bhatnagar13SR}} for more details on SPSA and to~\citet{Prashanth13AC} for its application to learning in mean-variance risk-sensitive MDPs.} In order to see how SPSA can help us to estimate our sub-gradient incrementally, note that 
\begin{equation}
\label{eq:grad-nu3}
\partial_\nu L(\nu,\theta,\lambda) = \lambda + \partial_\nu\left(\E\big[{\color{black}\mathcal J}^\theta(x^0)\big] + \frac{\lambda}{(1-\alpha)}\E\Big[\big({\color{black}\mathcal J}^\theta(x^0)- \nu\big)^+\Big]\right) \stackrel{\text{(a)}}{=} \lambda + \partial_\nu V^\theta(x^0,\nu).
\end{equation}
Similar to Sections~\ref{subsec:grad-theta}--\ref{subsec:grad-lambda_nu}, equality {(a)} comes from the fact that the quantity in  parenthesis in \eqref{eq:grad-nu3} is $V^\theta(x^0,\nu)$, the value function of the policy $\theta$ at state $(x^0,\nu)$ in the augmented MDP $\bar{\mathcal{M}}$. Since the critic uses a linear approximation for the value function, i.e.,~$V^\theta(x,s)\approx v^\top\phi(x,s)$, in our actor-critic algorithms (see Section~\ref{subsec:grad-theta} and Algorithm~\ref{alg:AC}), the SPSA estimate of the sub-gradient would be of the form $g(\nu)\approx\lambda+v^\top\big[\phi(x^0,\nu^+)-\phi(x^0,\nu^-)\big]/2\Delta$.

\subsection{Convergence of Actor-Critic Methods}
In this section, we  will prove that the actor-critic algorithms converge to a {\color{black} locally optimal policy for the CVaR-constrained optimization problem}. Define 
\[
{\color{black} \epsilon_\theta(v_k)=\|B_\theta[\Phi v_k]-\Phi v_k\|_{\infty}}
\]
as the residual of the value function approximation at step $k$,
induced by policy $\mu(\cdot|\cdot,\cdot;\theta)$. By the triangle
inequality and fixed point theorem $B_\theta[V^*]=V^*$, it can be
easily seen that {\color{black} $\|V^*-\Phi v_k\|_{\infty}\leq
  \epsilon_\theta(v_k)+\|B_\theta[\Phi
  v_k]-B_\theta[V^*]\|_{\infty}\leq
  \epsilon_\theta(v_k)+\gamma\|\Phi v_k-V^*\|_{\infty}$. The last
  inequality follows from the contraction property of the Bellman operator. Thus, one concludes that $\|V^*-\Phi v_k\|_{\infty}\leq \epsilon_\theta(v_k)/(1-\gamma)$.} Now, we state the main theorem for the convergence of actor-critic methods.
\begin{theorem}\label{thm:converge_kncre}
Suppose $\epsilon_{\theta_k}(v_k)\rightarrow 0$ and the
$\gamma$-{\color{black} occupation measure}  $\pi_{\gamma}^\theta$ is used to
generate samples of $(x_k,s_k,a_k)$ for any $k\in\{0,1,\ldots\}$. For
the SPSA-based algorithms, suppose the feature vector satisfies
the technical Assumption~\ref{assume_lip} {\color{black}(provided in Appendix~\ref{subsec:actor_update})} and {\color{black}suppose the SPSA step-size satisfies the condition
$\epsilon_{\theta_k}(v_k)=o(\Delta_k)$, i.e., $\epsilon_{\theta_k}(v_k)/\Delta_k\rightarrow 0$}. Then under Assumptions~\ref{ass:finite_time}--\ref{ass:feasibility} and \ref{ass:steps_ac}--\ref{ass:basis}, the sequence of {\color{black}policy updates in Algorithm \ref{alg:AC} converges almost surely to a locally optimal policy for the CVaR-constrained optimization problem}.
\end{theorem}
Details of the proof can be found in Appendix \ref{subsec:convergence-proof-AC}. 

\section{Extension to Chance-Constrained Optimization of MDPs}\label{sec:gen_chance}
In many applications, in particular in engineering (see, for example, \citet{ono2014chance}), \emph{{\color{black} chance constraints}} are imposed to ensure mission success with high probability. Accordingly, in this section we extend the analysis of CVaR-constrained MDPs to {\color{black} chance-constrained} MDPs (i.e.,~\eqref{eq:norm_reward_eqn3}). 
As for CVaR-constrained MDPs, we employ a Lagrangian relaxation procedure~{\color{black}(Chapter 3 of \citet{bertsekas1999nonlinear})} to convert a {\color{black} chance-constrained} optimization problem into the following unconstrained problem:  
\begin{equation}
\label{eq:unconstrained-discounted-risk-measure_extend}
\max_\lambda\min_{\theta,\alpha}\bigg(L(\theta,\lambda):= {\color{black}\mathcal G}^\theta(x^0)+\lambda\Big(\mathbb P\big({\color{black}\mathcal J}^\theta(x^0)\geq \alpha\big)-\beta\Big)\bigg),
\end{equation}
where $\lambda$ is the Lagrange multiplier. {\color{black}Recall
  Assumption~\ref{ass:feasibility} which assumed strict feasibility}, i.e.,~there exists a transient policy $\mu(\cdot|x;\theta)$ such that $\mathbb P\big({\color{black}\mathcal J}^\theta(x^0)\geq \alpha\big)< \beta$. This is needed to guarantee the existence of a local saddle point. \\

%\noindent
%{\bf TODO: You need to motivate this problem better, either here or at the beginning of the paper. This is not a problem/formulation known to the ML people.}
\subsection{Policy Gradient Method}
In this section we propose a policy gradient method for {\color{black} chance-constrained} MDPs (similar to Algorithm~\ref{alg_traj}). Since we do not need to estimate the $\nu$-parameter in {\color{black} chance-constrained} optimization, the corresponding policy gradient algorithm can be simplified and at each inner loop of Algorithm~\ref{alg_traj} we only perform the following updates at the end of each trajectory:
\begin{align*}
\textrm{\bf $\theta$ Update:} \quad &  \theta_{k+1} = \Gamma_\Theta\bigg[\theta_k -\frac{\zeta_2(k)}{N}\bigg(\sum_{j=1}^N\nabla_\theta\log\mathbb{P}(\xi_{j,k}){\color{black}\mathcal G}(\xi_{j,k}) + \lambda_k\nabla_\theta\log\mathbb{P}(\xi_{j,k})\mathbf{1}\big\{{\color{black}\mathcal J}(\xi_{j,k})\geq\alpha\big\}\bigg)\bigg] \\
\textrm{{\bf $\lambda$ Update:}}\quad&\lambda_{k+1} = \Gamma_\Lambda\bigg[\lambda_k + \zeta_1(k)\bigg( \!\!\!- \beta + \frac{1}{N}\sum_{j=1}^N\mathbf{1}\big\{{\color{black}\mathcal J}(\xi_{j,k})\geq\alpha\big\}\bigg)\bigg]
\end{align*}
Considering  the multi-time-scale step-size rules in
Assumption~\ref{ass:steps_pg}, the $\theta$ update is on the fast time-scale $\big\{\zeta_2(k)\big\}$ and the Lagrange multiplier $\lambda$ update is on the slow time-scale $\big\{\zeta_1(k)\big\}$. This results in a two time-scale stochastic approximation algorithm. In the following theorem, we prove that our policy gradient algorithm converges to a {\color{black}locally optimal policy for the chance-constrained problem}. 
\begin{theorem}
Under Assumptions~\ref{ass:finite_time}--\ref{ass:steps_pg}, the sequence of policy updates in Algorithm~\ref{alg_traj} converges to a {\color{black}locally optimal policy $\theta^*$ for the chance-constrained optimization problem} almost surely.
\end{theorem}
\begin{prooff}[{Sketch}]
By taking the gradient of $L(\theta,\lambda)$ w.r.t.~$\theta$, we have
\begin{equation*}
\nabla_\theta L(\theta,\lambda)=\nabla_\theta {\color{black}\mathcal G}^\theta(x^0) + \lambda \nabla_\theta\mathbb P\big({\color{black}\mathcal J}^\theta(x^0)\geq \alpha\big)=\sum_{\xi} \nabla_\theta\mathbb{P}_\theta(\xi){\color{black}\mathcal G}(\xi) + \lambda\sum_\xi \nabla_\theta\mathbb{P}_\theta(\xi)\mathbf{1}\big\{{\color{black}\mathcal J}(\xi)\geq\alpha\big\}.
\end{equation*}
On the other hand, the gradient of $L(\theta,\lambda)$ w.r.t.~$\lambda$ is given by
\[
\nabla_\lambda L(\theta, \lambda) = \mathbb P\big({\color{black}\mathcal J}^\theta(x^0)\geq \alpha\big) - \beta.
\]
One can easily verify that the $\theta$ and $\lambda$ updates are therefore unbiased estimates of $\nabla_\theta L(\theta, \lambda)$ and $\nabla_\lambda L(\theta, \lambda)$, respectively. Then the rest of the proof follows analogously from the convergence proof of Algorithm~\ref{alg_traj} in steps 2 and 3 of Theorem~\ref{thm:converge_h}.
\end{prooff}

\subsection{Actor-Critic Method}
In this section, we present an actor-critic algorithm for the {\color{black} chance-constrained} optimization. 
Given the original MDP $\mathcal{M}=(\X,\A,C,D,P,P_0)$ and parameter $\lambda$, we define the augmented MDP $\bar{\mathcal{M}}=(\bar{\X},\bar{\A},\bar{C}_\lambda,\bar{P},\bar{P}_0)$ as in the CVaR counterpart, except {\color{black}that} $\bar{P}_0(x,s)={P}_0(x)\mathbf 1\{s=\alpha\}$ and 
\begin{alignat*}{1}
\bar{C}_{\lambda}(x,s,a)&=\left\{\begin{array}{cc}
\lambda\mathbf 1\{s\leq 0\}& \text{if $x={\color{black} x_{\text{Tar}}}$,}\\
C(x,a)& \text{otherwise.}
\end{array}\right.
\end{alignat*}
Thus, the total cost of {\color{black}a} trajectory can be written as
\begin{equation}
\label{eq:aug-loss_AC}
\sum_{k=0}^{T}\bar{C}_\lambda(x_k,s_k,a_k)\mid x_0=x, s_0=\beta,\;\mu = {\color{black}\mathcal G}^\theta(x) + \lambda \mathbb P({\color{black}\mathcal J}^\theta(x)\geq \beta).
\end{equation}
Unlike the actor-critic algorithms for CVaR-constrained optimization, here the value function approximation parameter $v$, policy $\theta$, and Lagrange multiplier estimate $\lambda$ are updated episodically, i.e., after each episode ends by time $T$ when $(x_k,s_k)=(x_{\text{Tar}},s_{\text{Tar}})$\footnote{Note that ${\color{black}s_{\text{Tar}}}$ is the state of $s_t$ when $x_t$ hits the (recurrent) target state ${\color{black} x_{\text{Tar}}}$.}, as follows:
\begin{align}
\textrm{\bf {\color{black}Critic} Update:} \quad & 
v_{k+1}=v_k+\zeta_3(k)\sum_{h=0}^{T}\phi(x_h,s_h)\delta_{h}(v_k)\\
\textrm{{\bf Actor Updates:}}\quad & \theta_{k+1} = \Gamma_\Theta\Big(\theta_k-\zeta_2(k)\sum_{h=0}^{T}\nabla_\theta\log\mu(a_h|x_h,s_h;\theta)\vert_{\theta=\theta_k}\cdot\delta_h(v_k)\Big)  \\
\quad&\lambda_{k+1}=\Gamma_\Lambda\Big(\lambda_k+\zeta_1(k) \big(-\beta+\mathbf 1\{{\color{black}s_{\text{Tar}}}\leq 0\}\big)\Big)   
\end{align}
From analogous analysis as for the CVaR actor-critic method, the following theorem shows that the critic update $v_k$ converges almost surely to $v^*$. 
\begin{theorem}\label{thm:converge_kncre_v_extend}
Let $v^*\in\arg\min_{v}\|B_\theta[\Phi v]-\Phi v\|_{d^\theta}^2$ be {\color{black}a} minimizer of the Bellman residual, where the undiscounted Bellman operator at every $(x,s)\in\bar{\mathcal X}'$ is given by 
\[
B_\theta[V](x,s)=\sum_{a\in\mathcal A}\mu(a|x,s;\theta)\Big\{\bar{C}_\lambda(x,s,a)+\sum_{(x^\prime,s^\prime)\in\bar{\mathcal X}' }\bar{P}(x^\prime,s^\prime|x,s,a)V(x^\prime,s^\prime)\Big\}
\]
and $\tilde V^*(x,s)=\phi^\top(x,s)v^*$ is the projected Bellman fixed point of $ V^\theta(x,s)$, i.e., $\tilde V^*(x,s)=\Pi B_\theta[\tilde V^*](x,s)$ for $(x,s)\in\bar{\mathcal X}'$. Then under Assumptions~\ref{ass:steps_ac}--\ref{ass:basis}, the $v$-update in the actor-critic algorithm converges to $v^*$ almost surely.
\end{theorem}
\begin{prooff}[{Sketch}]
The proof of this theorem follows the same steps as {those}  in the proof of Theorem~\ref{thm:converge_kncre_v}, except replacing the $\gamma$-{\color{black} occupation measure}  $d^\theta_\gamma$ {with} the {\color{black} occupation measure}  {\color{black}$d^\theta$ (the total visiting probability). Similar analysis can also be found in the proof of Theorem 10 in \citet{tamar2013variance}. Under Assumption \ref{ass:finite_time}, the occupation measure of any transient states $x\in\mathcal X'$ (starting at an arbitrary initial transient state $x_0\in\mathcal X'$) can be written as $d^\mu(x|x^0)=\sum_{t=0}^{T_{\mu,x}} \mathbb P(x_t=x|x^0;\mu)$ when  $\gamma=1$. 
This further implies the total visiting probabilities are bounded as follows: $d^\mu(x|x^0)\leq T_{\mu,x}$ and $\pi^\mu(x,a|x^0)\leq T_{\mu,x}$ for any $x,x_0\in\mathcal X'$. Therefore, when the sequence of states $\{(x_h,s_h)\}_{h=0}^T$ is sampled by the $h$-step transition distribution $\mathbb P(x_h,s_h\mid x^0,s^0,\theta)$, $\forall h\leq T$}, the unbiased estimators of 
\[
A:=\!\!\!\!\sum_{(y,s^\prime)\in\bar{\mathcal X}',a^\prime\in\mathcal A}\!\!\!\!\pi^\theta(y,s^\prime,a^\prime|x,s)\phi(y,s^\prime)\Big(\phi^\top(y,s^\prime)- \sum_{(z,s^{\prime\prime})\in\bar{\mathcal X}'}\bar{P}(z,s^{\prime\prime}|y,s^\prime,a) \phi^\top(z,s^{\prime\prime})\Big)
\]
and 
\[
b:=\sum_{(y,s^\prime)\in\bar{\mathcal X}', a^\prime\in\mathcal A}\pi^\theta(y,s^\prime,a^\prime|x,s)\phi(y,s^\prime)\bar{C}_{\lambda}(y,s^\prime,a^\prime)
\]
are given by $\sum_{h=0}^{T}\phi(x_h,s_h)(\phi^\top(x_h,s_h)-\phi^\top(x_{h+1},s_{h+1}))$ and $\sum_{h=0}^{T}\phi(x_h,s_h)\bar{C}_{\lambda}(x_h,s_h,a_h)$, respectively. Note that in this theorem, we directly use the results from Theorem 7.1 in~\citet{BertsekasDP01} to show that every eigenvalue of matrix $A$ has positive real part, instead of using the technical result in Lemma~\ref{lem:tech_A_CVaR}.
\end{prooff}

Recall that {\color{black}$\epsilon_\theta(v_k)=\|B_\theta[\Phi
  v_k]-\Phi v_k\|_{\infty}$} is the residual of the value function
approximation at step $k$ induced by policy
$\mu(\cdot|\cdot,\cdot;\theta)$. By the triangle inequality and
fixed-point theorem of stochastic stopping problems, i.e.,
$B_\theta[V^*]=V^*$ from Theorem 3.1 in~\citet{BertsekasDP01}, it
can be easily seen that {\color{black}$\|V^*-\Phi v_k\|_{\infty}\leq
  \epsilon_\theta(v_k)+\|B_\theta[\Phi
  v_k]-B_\theta[V^*]\|_{\infty}\leq
  \epsilon_\theta(v_k)+\kappa\|\Phi v_k-V^*\|_{\infty}$} for some
$\kappa\in(0,1)$. Similar to the actor-critic algorithm
{\color{black}for} CVaR-constrained optimization, the last
inequality also follows from the contraction mapping {\color{black}property} of $B_\theta$ from Theorem 3.2 in~\citet{BertsekasDP01}. Now, we state the main theorem for the convergence of the actor-critic {\color{black}method}.
\begin{theorem}\label{thm:converge_kncre_extend}
Under Assumptions~\ref{ass:finite_time}--\ref{ass:basis}, if $\epsilon_{\theta_k}(v_k)\rightarrow 0$, then the sequence of policy updates converges almost surely to a {\color{black}locally optimal policy $\theta^*$ for the chance-constrained optimization problem}.
\end{theorem}
\begin{prooff}[{Sketch }]
{\color{black}
From Theorem~\ref{thm:converge_kncre_v_extend}, the critic
update  converges to the minimizer of the Bellman residual. Since the
critic update converges on the fastest scale, as in the proof of
Theorem~\ref{thm:converge_kncre}, one can replace $v_k$ by
$v^*(\theta_k)$ in the convergence proof of the actor
update. Furthermore, {by sampling the sequence of states $\{(x_h,s_h)\}_{h=0}^T$ with the $h$-step transition distribution $\mathbb P(x_h,s_h\mid x^0,s^0,\theta)$, $\forall h\leq T$}, the unbiased estimator of the gradient of the linear approximation to the Lagrangian function is given by
\[
\nabla_\theta \tilde L^v(\theta,\lambda):=\sum_{(x,s)\in\bar{\mathcal X}',a\in\mathcal A}\pi^\theta(x,s,a|x_0=x^0,s_0=\nu)\nabla_{\theta}\log\mu(a|x,s;\theta)\tilde A^{\theta,v}(x,s,a),
\]
where $\tilde \Q^{\theta,v}(x,s,a)-v^\top\phi(x,s)$ is given by
$\sum_{h=0}^{T}\nabla_\theta\log\mu(a_h|x_h,s_h;\theta)\vert_{\theta=\theta_k}\cdot\delta_h(v^*)$
and the unbiased estimator of $\nabla_\lambda
L(\theta,\lambda)=-\beta+\mathbb P({\color{black}s_{\text{Tar}}}\leq 0)$ is given by
$-\beta+\mathbf 1\{{\color{black}s_{\text{Tar}}}\leq 0\}$. Analogous to
equation~\eqref{diff_A_A_tilde} in the proof of Theorem
\ref{thm:theta_AC}, by convexity of quadratic functions, we have for any value function approximation $v$,
\[
\sum_{(y,s')\in\bar{\mathcal X}',a^\prime\in\mathcal{A}} \pi^\theta(y,s^\prime,a^\prime|x,s)(A_\theta(y,s',a^\prime)-\tilde A^v_{\theta}(y,s',a^\prime))
\leq 2T\frac{\epsilon_\theta(v)}{1-\kappa},
\]
which further implies that $\nabla_\theta L(\theta,\lambda)-\nabla_\theta \tilde L^v(\theta,\lambda)\rightarrow 0$ when $\epsilon_\theta(v)\rightarrow 0$ at $v=v^*(\theta_k)$. The rest of the proof follows identical arguments as in steps 3 to 5 of the proof of Theorem~\ref{thm:converge_kncre}.
}
\end{prooff}

\section{Examples}\label{sec:example}
In this section we illustrate the effectiveness of our risk-{\color{black}constrained} policy gradient  and actor-critic algorithms by testing them on an American option stopping problem and on a long-term personalized {\color{black}advertisement-recommendation (ad-recommendation)} problem.
\subsection{The Optimal Stopping Problem}
{\color{black} We consider an optimal stopping problem of purchasing certain types of goods, in which the state at each
time step $k \leq T$ consists of a purchase cost $c_k$ and time $k$,
i.e.,~$x=(c_k,k)$, where $T$ is the {\color{black} deterministic upper
  bound of the random stopping time}. The purchase cost sequence
$\{c_k\}_{k=0}^T$ is randomly generated by a Markov chain with two
modes. Specifically, due to future market uncertainties, at time $k$
the random purchase cost at the next time step $c_{k+1}$ either grows
by a factor $f_u>1$, i.e., $c_{k+1}=f_uc_k$, with probability $p$, or
drops by a factor $f_d<1$, i.e., $c_{k+1}=f_dc_k$, with probability $1-p$. Here $f_u$ and $f_d$ are constants that represent the rates of appreciation (due to anticipated shortage of supplies from vendors) and depreciation (due to reduction of demands in the market) respectively.}
The agent (buyer)
should decide either to accept the present cost
{\color{black}($u_k=1$)} or wait {\color{black}($u_k=0$)}. If he/she
accepts the cost or when the system terminates at time $k=T$, the
purchase cost is set at $\max(K,c_k)$, where $K$ is the maximum cost
  threshold. {\color{black} Otherwise, to account for a steady rate of inflation, at each time step the buyer receives an extra cost of $p_h$ that is independent to the purchase cost.} Moreover, there is a
discount factor $\gamma\in(0,1)$ to account for the increase in the
buyer's affordability. Note that if we change cost to reward and
minimization to maximization, this is exactly the American option
pricing problem, a standard testbed to evaluate risk-sensitive
algorithms (e.g.,~see \citet{tamar2012policy}). Since the state
space {\color{black}size $n$ is exponential in $T$}, finding an exact
solution via {\color{black}dynamic programming (DP) quickly becomes}
infeasible, and thus {\color{black}the problem} requires approximation and sampling techniques.

The optimal stopping problem can be reformulated as follows
\begin{equation}
\label{eq:norm_reward_eqn_example}
\min_\theta \mathbb E\left[{\color{black}\mathcal G}^\theta(x^0)\right]\quad\quad
\text{{\color{black}subject to}} \quad\quad
\text{CVaR}_\alpha\big({\color{black}\mathcal
  G}^\theta(x^0)\big)\leq\beta\quad\text{or}\quad{\color{black}\mathbb{P}\left({\color{black}\mathcal
      G}^\theta(x^0)\ge\beta\right)\leq1-\alpha}\footnote{{\color{black}
    To ensure that the notation is consistent between the CVaR and chance constraints, in the chance constraint definition the confidence level is denoted by $\alpha$  and  the tolerance threshold of $\mathcal {G}^\theta(x^0)$ is denoted by $\beta$.}},
\end{equation}
where the discounted cost and constraint cost functions are identical
{\color{black}(${\color{black}\mathcal G}^\theta(x)={\color{black}\mathcal J}^\theta(x)$)} and are
both given by ${\color{black}\mathcal G}^\theta(x)=\sum_{k=0}^T\gamma^k\left(\mathbf
  1\{u_k=1\}{\color{black}\max(K,c_k)}+\mathbf
  1\{u_k=0\}p_h\right)\mid x_0=x,\;\mu$. We set the parameters of the
MDP as follows: $x_0=[1;0]$, $p_h=0.1$, $T = 20$,
{\color{black}$K=5$}, $\gamma=0.95$, $f_u=2$, $f_d=0.5$, and $p=0.65$. The confidence level and constraint threshold are given by $\alpha=0.95$ and $\beta=3$. The number of sample trajectories $N$ is set to $500,000$ and the parameter bounds are $\lambda_{\max}=5,000$ and $\Theta=[-20,20]^{\kappa_1}$, {\color{black}where the dimension of the basis functions is $\kappa_1=1024$}. {\color{black}We implement radial basis functions (RBFs) as
  feature functions and search over the class of Boltzmann policies
  $\left\{\theta: \theta=\{\theta_{x,a}\}_{x\in\mathcal X,a\in\mathcal
      A},\,\, \mu_\theta(a|x) = \frac{\exp(\theta_{x,a}^\top
      x)}{\sum_{a\in\mathcal A}\exp(\theta_{x,a}^\top x)}\right\}$.}

%%%%%%%%%%%%%%%%%%%%%%%%%%%%%%%%%%%%%%%%%%%%%%%%%%%%%%%%%%%
%%%%%%%%%%%%%%%%%%%%%%%%%%%%%%%%%%%%%%%%%%%%%%%%%%%%%%%%%%%
%%%%%%%%%%%%%%%%%%%%%%%%%%%%%%%%%%%%%%%%%%%%%%%%%%%%%%%%%%%
We {consider} the following trajectory-based algorithms:
\begin{enumerate}
\item \textbf{PG:} This is a policy gradient algorithm that minimizes the expected discounted cost function without considering any risk criteria.
\item \textbf{PG-CVaR/{\color{black}PG-CC}:} These are the CVaR/{\color{black} chance-constrained}
  simulated trajectory-based policy gradient algorithms given
  in Section \ref{sec:PG-alg}. \\
\end{enumerate}
The experiments for each algorithm comprise the following two phases:
\begin{enumerate}
\item \textbf{Tuning phase:} We run the algorithm and update the policy until $(\nu,\theta,\lambda)$ converges.
\item \textbf{Converged run:} {\color{black}Having obtained a
    converged policy $\theta^*$ in the tuning phase,} in the converged run phase, we perform a Monte Carlo simulation of $10,000$ trajectories and report the results as averages over these trials.
\end{enumerate}

We also consider  the following incremental algorithms:
\begin{enumerate}
\item \textbf{AC:} This is an actor-critic algorithm that minimizes the expected discounted cost function without considering any risk criteria. This is similar to Algorithm~1~in~\citet{Bhatnagar10AC}.
\item \textbf{AC-CVaR/AC-VaR:} These are the CVaR/{\color{black} chance-constrained} semi-trajectory actor-critic algorithms given in Section \ref{sec:AC-alg}. 
\item \textbf{AC-CVaR-SPSA:} This is the CVaR-constrained SPSA actor-critic algorithm given in Section \ref{sec:AC-alg}. 
\end{enumerate}

Similar to the trajectory-based algorithms, we use RBF features for $[x;s]$ and consider the family of augmented state Boltzmann policies. Similarly, the experiments comprise two phases: {1)} the tuning phase, where the set of parameters $(v,\nu,\theta,\lambda)$ is obtained after the algorithm converges, and { 2)} the converged run, where the policy is simulated with $10,000$ trajectories.

We compare the performance of PG-CVaR and {PG-CC} (given in  Algorithm~\ref{alg_traj}), and AC-CVaR-SPSA, AC-CVaR, and AC-VaR (given in  Algorithm~\ref{alg:AC}), with PG and AC, their risk-neutral counterparts. Figures~\ref{fig:discounted_perf_traj_CVaR} and~\ref{fig:discounted_perf_traj_VaR} show the distribution of the discounted cumulative cost ${\color{black}\mathcal G}^\theta(x^0)$ for the policy $\theta$ learned by each of these algorithms. The results indicate that the risk-{\color{black}constrained} algorithms yield a higher expected cost, but less worst-case variability, compared to the risk-neutral methods. More precisely, the cost distributions of the risk-{\color{black}constrained} algorithms have lower right-tail (worst-case) distribution than their risk-neutral counterparts. Table~\ref{tab:discounted_perf} summarizes the performance of these algorithms. The numbers reiterate what we concluded from Figures~\ref{fig:discounted_perf_traj_CVaR} and~\ref{fig:discounted_perf_traj_VaR}.

{\color{black}
Notice that while the risk averse policy satisfies the CVaR constraint, it is not tight (i.e., the constraint is not matched). In fact this is a problem of local optimality, and other experiments
in the literature (for example see the numerical results in \citet{Prashanth13AC} and in \citet{bhatnagar2012online}) have the same problem of producing
solutions which obey the constraints but not tightly. {\color{black}
  However, since both the expectation and the CVaR risk metric are sub-additive and convex, one can always construct a policy that is a linear combination of the risk neutral optimal policy and the risk averse policy such that it matches the constraint threshold and has a lower cost compared to the risk averse policy. }
}
\begin{figure*}[th!]
\vspace{-0.05in}
\centering
\includegraphics[width=2.8in,angle=0]{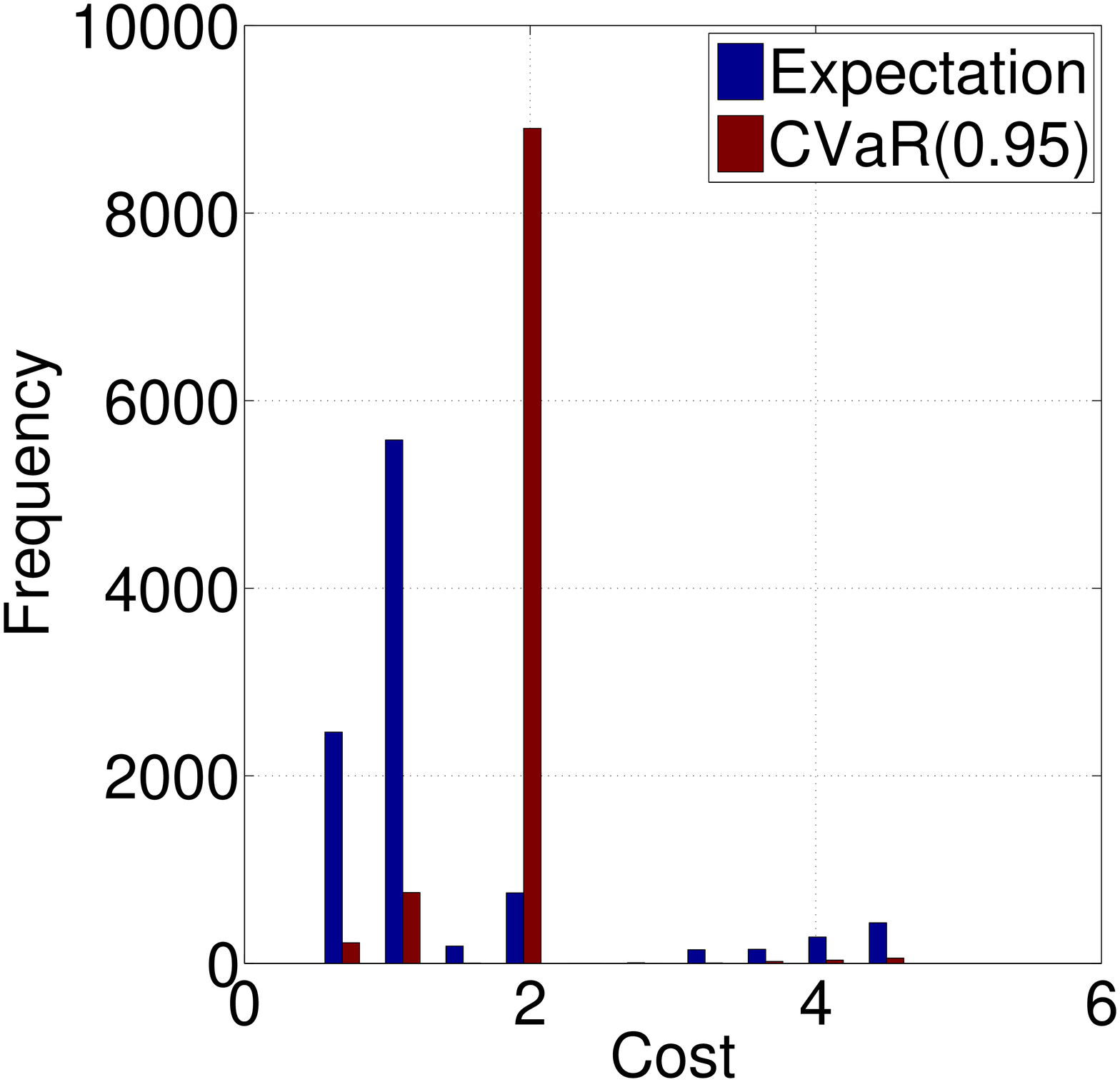}
\includegraphics[width=2.8in,angle=0]{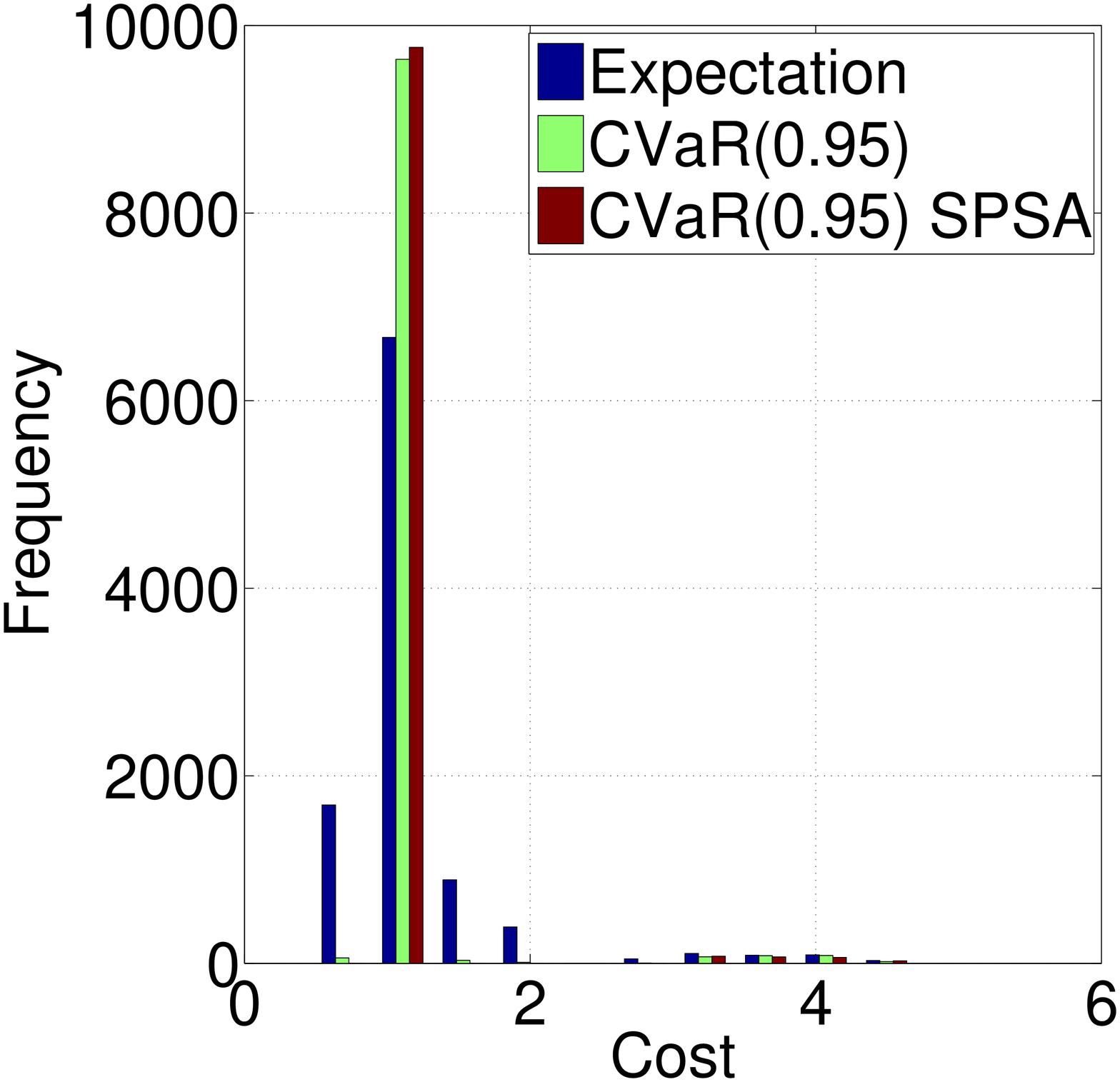}
%     \caption{Performance Comparison of a plain AC algorithm vs. RSAC algorithm using the average reward $\rho$ distribution}
\vspace{-0.15in}
\caption{Cost distributions for the policies learned by the {\color{black}CVaR-constrained} and risk-neutral policy gradient and actor-critic algorithms. The left figure corresponds to the PG methods and the right figure corresponds to the AC algorithms.
%Performance comparison  using the distribution of $D^\theta(x^0)$ for policy gradient algorithms. 
% in both average as well as discounted settings.
}\label{fig:discounted_perf_traj_CVaR}
\vspace{-0.025in}
\end{figure*}

\begin{figure*}[th!]
\vspace{-0.05in}
\centering
\includegraphics[width=2.8in,angle=0]{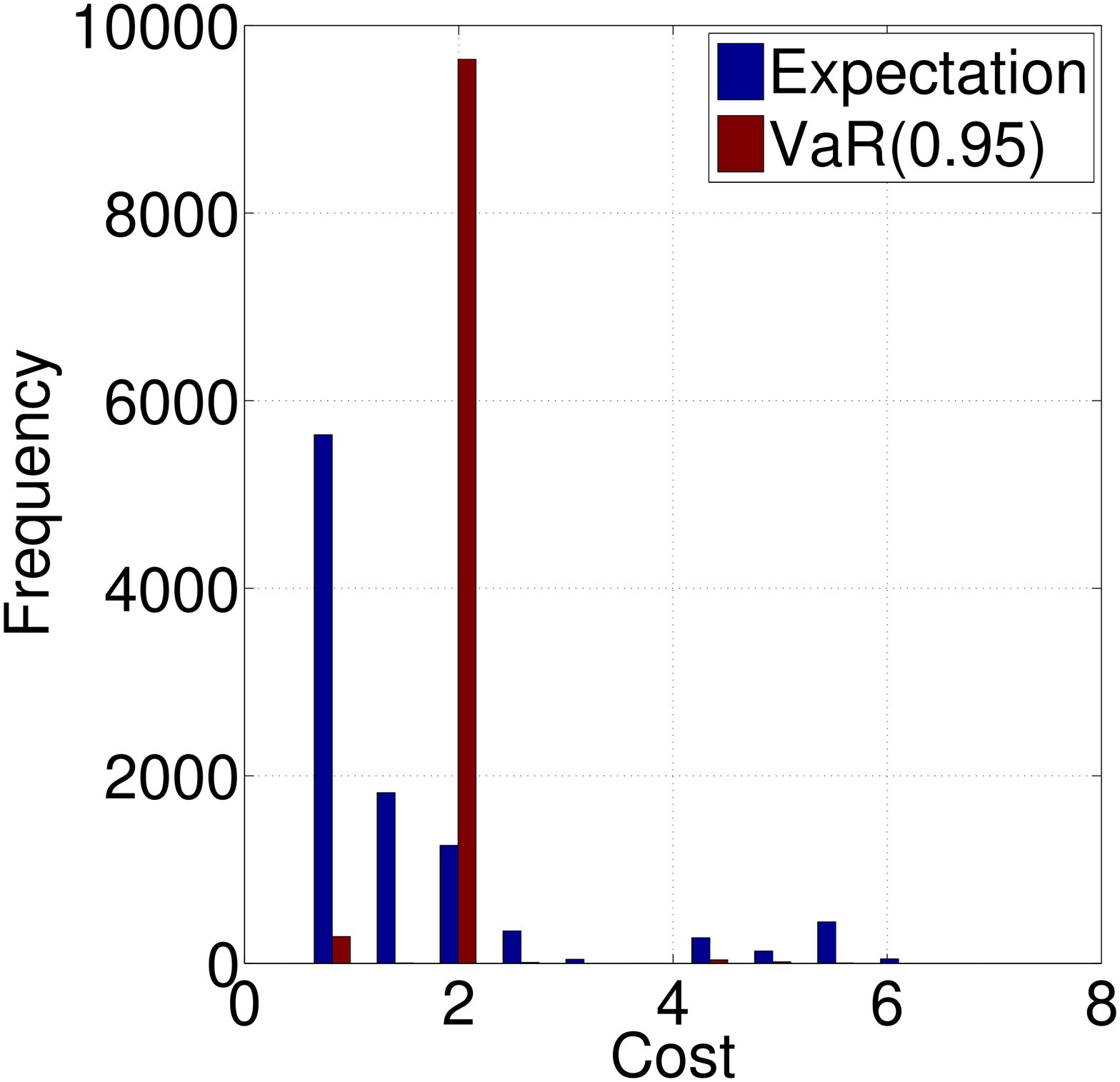}
\includegraphics[width=2.8in,angle=0]{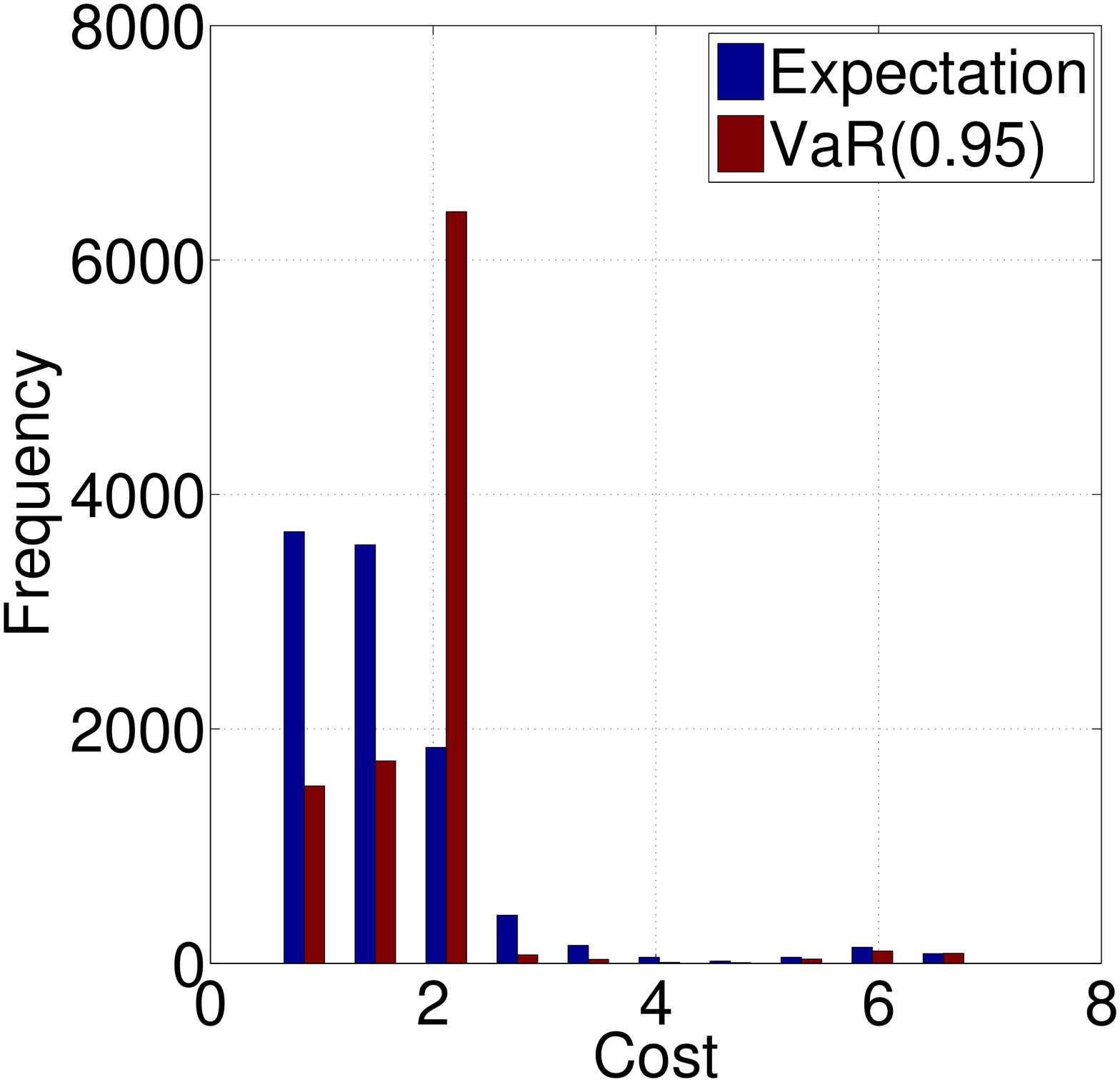}
%     \caption{Performance Comparison of a plain AC algorithm vs. RSAC algorithm using the average reward $\rho$ distribution}
\vspace{-0.15in}
\caption{Cost distributions for the policies learned by the {\color{black} chance-constrained} and risk-neutral policy gradient and actor-critic algorithms. The left figure corresponds to the PG methods and the right figure corresponds to the AC algorithms.
%Performance comparison  using the distribution of $D^\theta(x^0)$ for policy gradient algorithms. 
% in both average as well as discounted settings.
}\label{fig:discounted_perf_traj_VaR}
\vspace{-0.025in}
\end{figure*}

\begin{table*}[th!]
\centering
\begin{tabular}{ |c| c| c| c| c|c| }
  \hline
  &$\mathbb E\big({\color{black}\mathcal G}^\theta(x^0)\big)$ & $\sigma\big({\color{black}\mathcal G}^\theta(x^0)\big)$ & $\text{CVaR}\big({\color{black}\mathcal G}^\theta(x^0)\big)$& $\text{VaR}\big({\color{black}\mathcal G}^\theta(x^0)\big)$\\
  \hline
    PG & 1.177 & 1.065&4.464&4.005\\
  \hline
  PG-CVaR & 1.997& 0.060 & 2.000 & 2.000   \\
      \hline
      {\color{black}PG-CC} & 1.994 &0.121 &2.058 &2.000 \\
  \hline
  AC &1.113 & 0.607 & 3.331&3.220\\
   \hline
  AC-CVaR-SPSA & 1.326 &0.322 &2.145 & 1.283  \\
    \hline
   AC-CVaR &1.343 &0.346 &2.208 & 1.290 \\
    \hline
       AC-VaR &1.817 &0.753 &4.006 & 2.300 \\
    \hline
\end{tabular}
%\vspace{-0.025in}
\caption{\small{Performance comparison of the policies learned by the risk-{\color{black}constrained} and risk-neutral algorithms. {\color{black}In this table $\sigma\big({\color{black}\mathcal G}^\theta(x^0)\big)$ stands for the standard deviation of the total cost.}}}
%Performance comparison  using the empirical risk metrics of $D^\theta(x^0)$. 
% in both average as well as discounted settings.
\label{tab:discounted_perf}
\vspace{-0.05in}
\end{table*}

\subsection{A Personalized Ad-Recommendation System}

Many companies such as banks and retailers use user-specific targeting of advertisements to attract more customers and increase their revenue. When a user requests a webpage that contains a box for an advertisement, the system should decide which advertisement (among those in the current campaign) to show to this particular user based on a vector containing all her features, often collected by a cookie. Our goal here is to generate a strategy that for each user of the website selects an ad that when it is presented to her has the highest probability to be clicked on. These days, almost all the industrial personalized ad recommendation systems use supervised learning or contextual bandits algorithms. These methods are based on the i.i.d.~assumption of the visits (to the website) and do not discriminate between a visit and a visitor, i.e.,~each visit is considered as a new visitor that has been sampled i.i.d.~from the population of the visitors. As a result, these algorithms are myopic and do not try to optimize for the long-term performance. Despite their success, these methods seem to be insufficient as users establish longer-term relationship with the websites they visit, i.e.,~the ad recommendation systems should deal with more and more returning visitors. The increase in returning visitors violates (more) the main assumption underlying the supervised learning and bandit algorithms, i.e.,~there is no difference between a visit and a visitor, and thus, shows the need for a new class of solutions.

The reinforcement learning (RL) algorithms that have been designed to optimize the long-term performance of the system (expected sum of rewards/costs) seem to be suitable candidates for ad recommendation systems {\color{black}\citep{shani2002mdp}}. The nature of these algorithms allows them to take into account all the available knowledge about the user at the current visit, and then selects an offer to maximize the total number of times she will click over multiple visits, also known as the user's life-time value (LTV). Unlike myopic approaches, RL algorithms differentiate between a visit and a visitor, and consider all the visits of a user (in chronological order) as a {\color{black}trajectory} generated by her. In this approach, while the visitors are i.i.d.~samples from the population of the users, their visits are not. This long-term approach to the ad recommendation problem allows us to make decisions that are not usually possible {\color{black}with} myopic techniques, such as to propose an offer to a user that might be a loss to the company in the short term, but has the effect that makes the user engaged with the website/company and brings her back to spend more money in the future.

For our second case study, we use an Adobe personalized ad-recommendation {\color{black}\citep{theocharous2013lifetime}} simulator that has been trained based on real data captured with permission from the website of a Fortune 50 company that receives hundreds of visitors per day. The simulator produces a vector of $31$ real-valued features that provide a compressed representation of all of the available information about a user. The advertisements are clustered into four high-level classes that the agent must select between. After the agent selects an advertisement, the user either clicks (reward of $+1$) or does not click (reward of $0$) and the feature vector describing the user is updated. In this case, we test our algorithm by maximizing the customers' life-time value in $15$ time steps {\color{black}subject to} a bounded tail risk.

Instead of using the cost-minimization framework from the main paper, 
by defining the return random variable (under a fixed policy $\theta$) $\mathcal R^\theta(x^0)$ as the (discounted) total number of clicks along a user's trajectory, here we formulate the personalized ad-recommendation problem as a return maximization problem where the tail risk corresponds to the worst case return distribution:
\begin{equation}
\label{eq:norm_reward_eqn_example}
\max_\theta \mathbb E\left[\mathcal R^\theta(x^0)\right]\quad\quad \text{{\color{black}subject to}} \quad\quad {\color{black}\text{CVaR}_{1-\alpha}\big(-\mathcal R^\theta(x^0)\big)\leq\beta}.
\end{equation}
We set the parameters of the MDP as $T = 15$ and $\gamma=0.98$, the confidence level and constraint threshold as $\alpha=0.05$ and {\color{black}$\beta=-0.12$}, the number of sample trajectories $N$ to $1,000,000$, and the parameter bounds as $\lambda_{\max}=5,000$ and $\Theta=[-60,60]^{\kappa_1}$, {\color{black}where the dimension of the basis functions is $\kappa_1=4096$}. Similar to the optimal stopping problem, we implement both the trajectory based algorithm (PG, PG-CVaR) and the actor-critic algorithms (AC, AC-CVaR) for risk-neutral and risk sensitive optimal control. Here we used the $3^{\text{rd}}$ order Fourier basis with cross-products in~\citet{konidaris2011value} as features and search over the family of Boltzmann policies. We compared the performance of PG-CVaR and AC-CVaR, our risk-{\color{black}constrained} policy gradient (Algorithm~\ref{alg_traj}) and actor-critic (Algorithms~\ref{alg:AC}) algorithms, with their risk-neutral counterparts (PG and AC). Figure~\ref{fig:discounted_perf_traj_adobe} shows the distribution of the discounted cumulative return $\mathcal R^\theta(x^0)$ for the policy $\theta$ learned by each of these algorithms. The results indicate that the risk-{\color{black}constrained} algorithms yield a lower expected reward, but have higher left tail (worst-case) reward distributions. Table~\ref{tab:discounted_perf_adobe} summarizes the findings of this experiment. 

\begin{figure*}[th!]
\vspace{-0.05in}
\centering
\includegraphics[width=2.8in,angle=0]{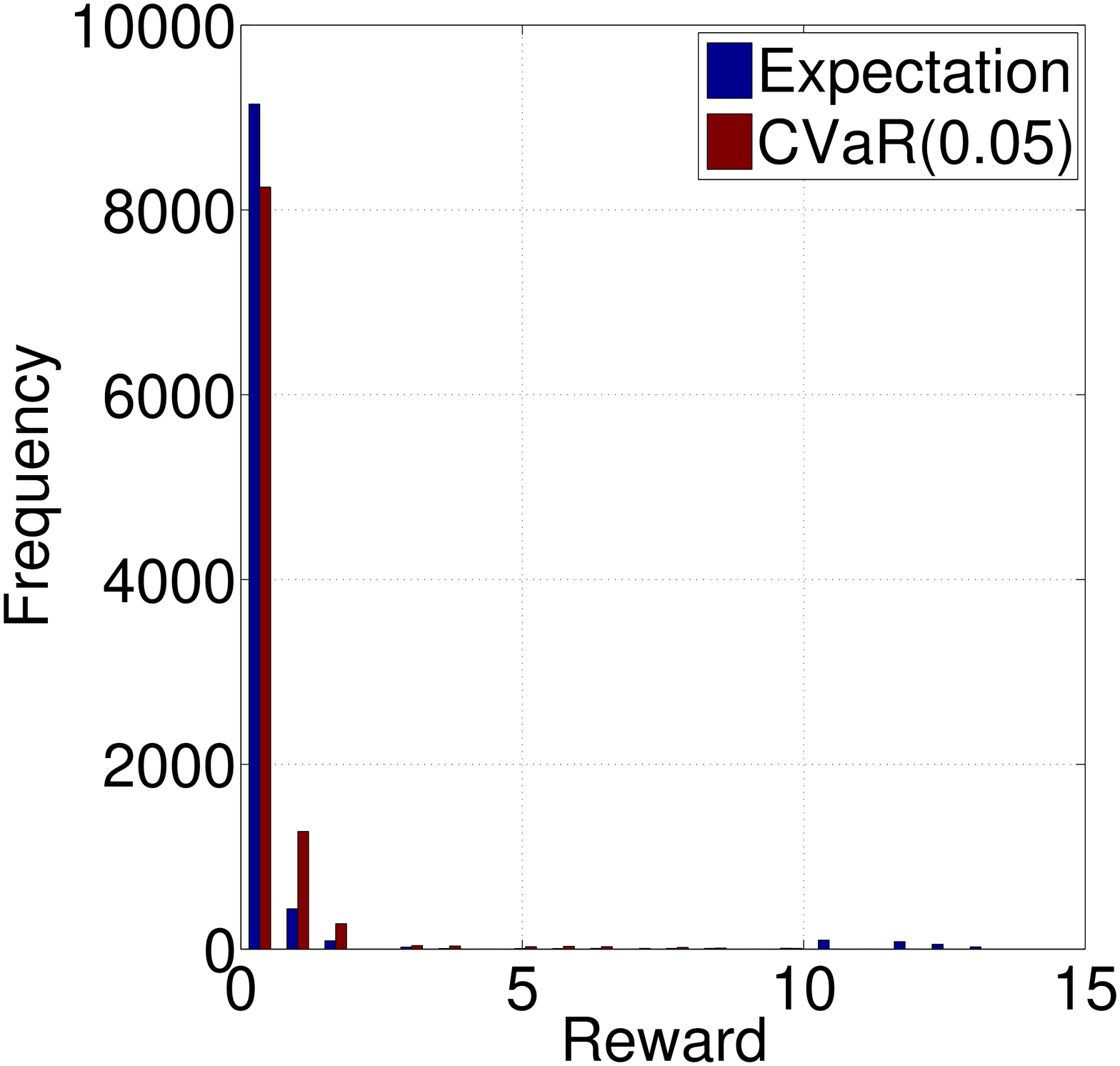}
\includegraphics[width=2.8in,angle=0]{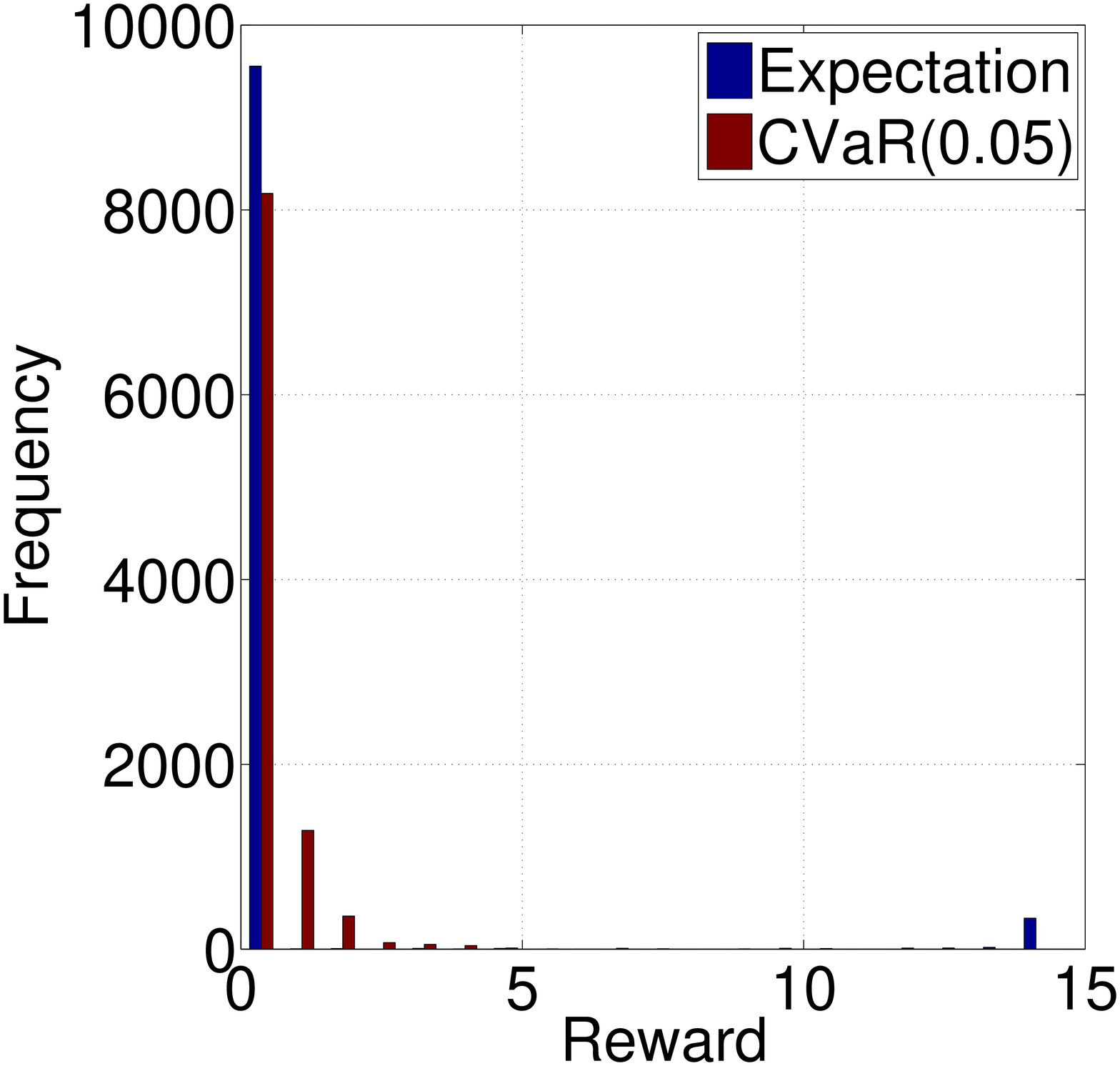}
%     \caption{Performance Comparison of a plain AC algorithm vs. RSAC algorithm using the average reward $\rho$ distribution}
\vspace{-0.15in}
\caption{Reward distributions for the policies learned by the {\color{black}CVaR-constrained} and risk-neutral policy gradient and actor-critic algorithms. The left figure corresponds to the PG methods and the right figure corresponds to the AC algorithms.
%Performance comparison  using the distribution of $D^\theta(x^0)$ for policy gradient algorithms. 
% in both average as well as discounted settings.
}\label{fig:discounted_perf_traj_adobe}
\vspace{-0.025in}
\end{figure*}

\begin{table*}[th!]
\centering
\begin{tabular}{ |c| c| c| c| c|c| }
  \hline
  &$\mathbb E\big(\mathcal R^\theta(x^0)\big)$ & $\sigma\big(\mathcal R^\theta(x^0)\big)$ & $\text{CVaR}\big(\mathcal R^\theta(x^0)\big)$& $\text{VaR}\big(\mathcal R^\theta(x^0)\big)$\\
  \hline
    PG & 0.396&1.898&0.037&1.000\\
  \hline
  PG-CVaR &0.287 &0.914&0.126& 1.795  \\
  \hline
  AC & 0.581&2.778&0&0\\
    \hline
   AC-CVaR &0.253 &0.634&0.137&1.890\\
    \hline
    \end{tabular}
%\vspace{-0.025in}
\caption{\small{Performance comparison of the policies learned by the {\color{black}CVaR-constrained} and risk-neutral algorithms. {\color{black}In this table $\sigma\big(\mathcal R^\theta(x^0)\big)$ stands for the standard deviation of the total reward.}}}
%Performance comparison  using the empirical risk metrics of $D^\theta(x^0)$. 
% in both average as well as discounted settings.
\label{tab:discounted_perf_adobe}
\vspace{-0.05in}
\end{table*}

\section{Conclusions and Future Work}
\label{sec:conclusions}

We proposed novel policy gradient and actor-critic algorithms for CVaR-constrained and {\color{black} chance-constrained} optimization in MDPs, and proved their convergence. Using an optimal stopping problem and a personalized ad-recommendation problem, we showed that our algorithms resulted in policies whose cost distributions have lower right-tail compared to their risk-neutral counterparts. This is  important for a risk-averse decision-maker, especially if the right-tail contains catastrophic costs. Future work includes: { 1)} Providing convergence proofs for our AC algorithms when the samples are generated by following the policy and not from its discounted {\color{black} occupation measure} ,{ \color{black} 2)} Using importance sampling methods \citep{Bardou09CV,Tamar15PG} to improve gradient estimates in the right-tail of the cost distribution (worst-case events that are observed with low probability), {\color{black}and 3) Applying the algorithms presented in this paper to a variety of applications ranging from operations research to robotics and finance}. \\
% Acknowledgements should go at the end, before appendices and references

%\noindent
%{\bf TODO: It would be good if we expand both the conclusions and future work a bit more.} 
\vspace{-0.3in}
\acks{We would like to thank Csaba Szepesvari for his comments that
  helped us with the derivation of the algorithms, Georgios
  Theocharous for sharing his ad-recommendation simulator with us, and
  Philip Thomas for helping us with the experiments with the
  simulator. {\color{black} We would also like to thank the reviewers
    for their very helpful comments and suggestions, which
    helped us to significantly improve the paper.} Y-L. Chow is partially supported by The Croucher
  Foundation doctoral scholarship. L. Janson was partially supported
  by NIH training grant T32GM096982. M. Pavone was partially supported
  by the Office of Naval Research, Science of Autonomy Program, under
  Contract N00014-15-1-2673.}

\begin{small}
\bibliography{CVaR-rl}
\end{small}

% Manual newpage inserted to improve layout of sample file - not
% needed in general before appendices/bibliography.

\newpage
\appendix
\section{Convergence of Policy Gradient Methods}
\label{sec:appendix_PG}

\subsection{Computing the Gradients}
\label{subsec:grad-comp}
{\bf i) $\nabla_\theta L(\nu,\theta,\lambda)$: Gradient of $L(\nu,\theta,\lambda)$ w.r.t.~$\theta$} 
By expanding the expectations in the definition of the objective function $L(\nu,\theta,\lambda)$ in \eqref{eq:unconstrained-discounted-risk-measure}, we obtain
\begin{equation*}
L(\nu,\theta,\lambda)=\sum_{\xi} \mathbb{P}_\theta(\xi){\color{black}\mathcal G}(\xi) + \lambda\nu + \frac{\lambda}{1-\alpha}\sum_\xi \mathbb{P}_\theta(\xi) \big({\color{black}\mathcal J}(\xi)-\nu\big)^+-\lambda\beta.
\end{equation*}
By taking the gradient with respect to $\theta$, we have
\begin{equation*}
\nabla_\theta L(\nu,\theta,\lambda)=\sum_{\xi} \nabla_\theta\mathbb{P}_\theta(\xi){\color{black}\mathcal G}(\xi) + \frac{\lambda}{1-\alpha}\sum_\xi \nabla_\theta\mathbb{P}_\theta(\xi) \big({\color{black}\mathcal J}(\xi)-\nu\big)^+.
\end{equation*}
This gradient can be rewritten as 
\begin{equation}\label{eq:L_theta}
\nabla_\theta L(\nu,\theta,\lambda)=\sum_{\xi:{\color{black}\mathbb{P}_\theta(\xi)\neq 0}} \mathbb{P}_\theta(\xi)\cdot\nabla_\theta\log\mathbb{P}_\theta(\xi)\left({\color{black}\mathcal G}(\xi) + \frac{\lambda}{1-\alpha}\big({\color{black}\mathcal J}(\xi)-\nu\big)\mathbf{1}\big\{{\color{black}\mathcal J}(\xi)\geq\nu\big\}\right),
\end{equation}
where {\color{black}in the case of $\mathbb{P}_\theta(\xi)\neq 0$, the term $\nabla_\theta\log\mathbb{P}_\theta(\xi)$ is given by:}
\begin{equation*}
\begin{split}
\nabla_\theta\log\mathbb{P}_\theta(\xi)=&\nabla_\theta\left\{\sum_{k=0}^{T-1} \log P(x_{k+1}|x_k,a_k)+\log\mu(a_k|x_k;\theta)+\log \mathbf 1\{x_0=x^0\}\right\}\\
=&\sum_{k=0}^{T-1}\nabla_\theta\log\mu(a_k|x_k;\theta)\\
=& \sum_{k=0}^{T-1}\frac{1}{\mu(a_k|x_k;\theta)}\nabla_\theta\mu(a_k|x_k;\theta).
\end{split}
\end{equation*}

\noindent
{\bf ii) $\partial_\nu L(\nu,\theta,\lambda)$: Sub-differential of $L(\nu,\theta,\lambda)$ w.r.t.~$\nu$}
From the definition of $L(\nu,\theta,\lambda)$, we can easily see that $L(\nu,\theta,\lambda)$ is a convex function in $\nu$ for any fixed $\theta\in \Theta$. Note that for every fixed $\nu$ and any $\nu^\prime$, we have
\begin{equation*}
\big({\color{black}\mathcal J}(\xi)-\nu'\big)^+ - \big({\color{black}\mathcal J}(\xi)-\nu\big)^+ \geq g\cdot(\nu'-\nu),
\end{equation*}
where $g$ is any element in the set of sub-derivatives: 
\begin{equation*}
g\in\partial_\nu\big({\color{black}\mathcal J}(\xi)-\nu\big)^+ :=
\begin{cases}
-1 & \text{if $\nu< {\color{black}\mathcal J}(\xi)$}, \\
-q:q\in[0,1] & \text{if $\nu={\color{black}\mathcal J}(\xi)$}, \\
0 & \text{otherwise}.
\end{cases}
\end{equation*} 
Since $L(\nu,\theta,\lambda)$ is finite-valued for any $\nu\in\reals$, by the additive rule of sub-derivatives, we have 
\begin{equation}\label{eq:L_nu}
\partial_\nu L(\nu,\theta,\lambda) = \left\{-\frac{\lambda}{1-\alpha}\sum_\xi\mathbb{P}_\theta(\xi)\mathbf{1}\big\{{\color{black}\mathcal J}(\xi)>\nu\big\} - \frac{\lambda q}{1-\alpha}\sum_\xi\mathbb{P}_\theta(\xi)\mathbf{1}\big\{{\color{black}\mathcal J}(\xi)=\nu\big\} + \lambda\mid q\in[0,1]\right\}.
\end{equation}
In particular for $q=1$, we may write the sub-gradient of $L(\nu,\theta,\lambda)$ w.r.t.~$\nu$ as
\[
\partial_\nu L(\nu,\theta,\lambda)\vert_{q=0}= \lambda - \frac{\lambda}{1-\alpha}\sum_\xi\mathbb{P}_\theta(\xi)\cdot\mathbf{1}\big\{{\color{black}\mathcal J}(\xi)\geq\nu\big\}
\]
or
\[ \lambda - \frac{\lambda}{1-\alpha}\sum_\xi\mathbb{P}_\theta(\xi)\cdot\mathbf{1}\big\{{\color{black}\mathcal J}(\xi)\geq\nu\big\} \in \partial_\nu L(\nu,\theta,\lambda).
\]

\noindent
{\bf iii) $\nabla_\lambda L(\nu,\theta,\lambda)$: Gradient of $L(\nu,\theta,\lambda)$ w.r.t.~$\lambda$}
Since $L(\nu,\theta,\lambda)$ is a linear function in $\lambda$, one can express the gradient of $L(\nu,\theta,\lambda)$ w.r.t.~$\lambda$ as follows:
\begin{equation}\label{eq:L_lambda}
\nabla_\lambda L(\nu,\theta,\lambda) = \nu - \beta +\frac{1}{1-\alpha}\sum_{\xi}\mathbb{P}_\theta(\xi)\cdot\big({\color{black}\mathcal J}(\xi)- \nu\big)\mathbf{1}\big\{{\color{black}\mathcal J}(\xi)\geq\nu\big\}.
\end{equation}

\subsection{Proof of Convergence of the Policy Gradient Algorithm}
\label{subsec:conv-proof}
In this section, we prove the convergence of the policy gradient algorithm (Algorithm~\ref{alg_traj}).
{\color{black}
Before going through the details of the convergence proof, a high level overview of the proof technique is given as follows.  
\begin{enumerate}
\item First, by convergence properties of multi-time scale discrete stochastic approximation algorithms, we show that each update $(\nu_k,\theta_k,\lambda_k)$ converges almost surely to a stationary point $(\nu^\ast,\theta^*,\lambda^*)$ of the corresponding continuous time system. In particular, by adopting the step-size rules defined in Assumption \ref{ass:steps_pg}, we show that the convergence rate of $\nu$ is fastest, followed by the convergence rate of $\theta$, while the convergence rate of $\lambda$ is the slowest among the set of parameters.
\item By using Lyapunov analysis, we show that the continuous time system is locally asymptotically stable at the stationary point $(\nu^\ast,\theta^*,\lambda^*)$. 
\item Since the Lyapunov function used in the above analysis is the Lagrangian function $L(\nu,\theta,\lambda)$, we conclude that the stationary point $(\nu^\ast,\theta^*,\lambda^*)$ is a local saddle point. Finally by the local saddle point theorem, we deduce that $\theta^*$ is a locally optimal solution for the CVaR-constrained MDP problem.
\end{enumerate}
This convergence proof procedure is standard for stochastic approximation algorithms, see~\citep{bhatnagar2009natural,bhatnagar2012online} for further references.}

Since $\nu$ converges on the faster timescale than $\theta$ and $\lambda$, the $\nu$-update can be rewritten by assuming $(\theta,\lambda)$ as invariant quantities, i.e., 
\begin{equation}\label{nu_up_h_conv}
\nu_{k+1} = \Gamma_{\mathcal{N}}\bigg[\nu_k - \zeta_3(k)\bigg(\lambda - \frac{\lambda}{(1-\alpha)N}\sum_{j=1}^N\mathbf{1}\big\{{\color{black}\mathcal J}(\xi_{j,k})\geq\nu_k\big\}\bigg)\bigg].
\end{equation}
Consider the continuous time dynamics of $\nu$ defined using differential inclusion
\begin{equation}\label{dyn_sys_nu_h}
\dot{\nu}\in \Upsilon_{\nu}\left[-g(\nu)\right], \quad\quad \forall g(\nu)\in\partial_\nu L(\nu,\theta,\lambda),
\end{equation}
where 
\[
\Upsilon_\nu[K(\nu)]:=\lim_{0<\eta\rightarrow 0} \frac{\Gamma_{\mathcal{N}}(\nu+\eta K(\nu))-\Gamma_{\mathcal{N}}(\nu)}{\eta}.
\]
Here $\Upsilon_\nu[K(\nu)]$ is the left directional derivative of the function $\Gamma_{\mathcal{N}}(\nu)$ in the direction of $K(\nu)$. By using the left directional derivative $\Upsilon_\nu\left[ -g(\nu) \right]$ in the sub-gradient descent algorithm for $\nu$, the gradient will point {\color{black}in} the descent direction along the boundary of $\nu$ whenever the $\nu$-update hits its boundary. 

Furthermore, since $\nu$ converges on {\color{black}a} faster timescale than $\theta$, and $\lambda$ is on the slowest time-scale, the $\theta$-update can be rewritten using the converged $\nu^*(\theta)$, assuming $\lambda$ as an invariant quantity, i.e.,

\begin{equation}\label{theta_up_h_conv}
\begin{split}
\theta_{k+1}=&\Gamma_\Theta
\bigg[\theta_{k}-\zeta_2(k)\bigg(\frac{1}{N}\sum_{j=1}^N\nabla_\theta\log\mathbb{P}_\theta(\xi_{j,k})\vert_{\theta=\theta_k}{\color{black}\mathcal G}(\xi_{j,k}) \nonumber \\ 
&\hspace{0.725in}+ \frac{\lambda}{(1-\alpha)N}\sum_{j=1}^N\nabla_\theta\log\mathbb{P}_\theta(\xi_{j,k})\vert_{\theta=\theta_k}\big({\color{black}\mathcal J}(\xi_{j,k})-\nu\big)\mathbf{1}\big\{{\color{black}\mathcal J}(\xi_{j,k})\geq\nu^*(\theta_k)\big\}\bigg)\bigg].
\end{split}\end{equation}
Consider the continuous time dynamics of $\theta\in \Theta$:
\begin{equation}\label{dyn_sys_kheta_h}
\dot{\theta}=\Upsilon_\theta\left[ -\nabla_\theta L(\nu,\theta,\lambda) \right]\vert_{\nu=\nu^*(\theta)},
\end{equation}
where
\[
\Upsilon_\theta[K(\theta)]:=\lim_{0<\eta\rightarrow 0} \frac{\Gamma_\Theta(\theta+\eta K(\theta))-\Gamma_\Theta(\theta)}{\eta}.
\]
Similar to the analysis of $\nu$, $\Upsilon_\theta[K(\theta)]$ is the left directional derivative of the function $\Gamma_\Theta(\theta)$ in the direction of $K(\theta)$. By using the left directional derivative $\Upsilon_\theta\left[ - \nabla_\theta L(\nu,\theta,\lambda)\right]$ in the gradient descent algorithm for $\theta$, the gradient will point {\color{black}in} the descent direction along the boundary of $\Theta$ whenever the $\theta$-update hits its boundary.

Finally, since {\color{black}the} $\lambda$-update converges in {\color{black}the} slowest time-scale, the $\lambda$-update can be rewritten using the converged $\theta^*(\lambda)$ and $\nu^*(\lambda)$, i.e.,
\begin{equation}\label{lamda_up_h_conv}
\lambda_{k+1}=\Gamma_\Lambda\left(\lambda_k+\zeta_1(k) \bigg(\nu^*(\lambda_k)+\frac{1}{1-\alpha} \frac{1}{N}\sum_{j=1}^N\big({\color{black}\mathcal J}(\xi_{j,k})- \nu^*(\lambda_k)\big)^+ -\beta\bigg)\right).
\end{equation}
Consider the continuous time system
\begin{equation}\label{dyn_sys_lambda_h}
\dot{\lambda}(t)=\Upsilon_\lambda\left[ \nabla_\lambda L(\nu,\theta,\lambda)\bigg\vert_{\theta=\theta^*(\lambda),\nu=\nu^*(\lambda)}\right], \quad\quad \lambda(t)\geq 0,
\end{equation}
where
\[
\Upsilon_\lambda[K(\lambda)]:=\lim_{0<\eta\rightarrow 0} \frac{\Gamma_\Lambda\big(\lambda+\eta K(\lambda)\big)-\Gamma_\Lambda(\lambda)}{\eta}.
\]
Again, similar to the analysis of $(\nu,\theta)$, $\Upsilon_\lambda[K(\lambda)]$ is the left directional derivative of the function $\Gamma_\Lambda(\lambda)$ in the direction of $K(\lambda)$. By using the left directional derivative $\Upsilon_\lambda\left[ \nabla_\lambda L(\nu,\theta,\lambda)\right]$ in the gradient ascent algorithm for $\lambda$, the gradient will point {\color{black}in} the ascent direction along the boundary of $[0,\lambda_{\max}]$ whenever the $\lambda$-update hits its boundary.

Define
\[
L^*(\lambda)=L(\nu^\ast(\lambda),\theta^\ast(\lambda),\lambda),
\]
for $\lambda\geq 0$ where
$(\theta^\ast(\lambda),\nu^\ast(\lambda))\in\Theta\times[-\frac{D_{\max}}{1-\gamma},\frac{D_{\max}}{1-\gamma}]$
is a local minimum of $L(\nu,\theta,\lambda)$ for fixed $\lambda\geq
0$, i.e., $L(\nu,\theta,\lambda)\geq
L(\nu^\ast(\lambda),\theta^\ast(\lambda),\lambda)$ for any
$(\theta,\nu)\in\Theta\times[-\frac{D_{\max}}{1-\gamma},\frac{D_{\max}}{1-\gamma}]\cap
\mathcal B_{(\theta^\ast(\lambda),\nu^\ast(\lambda))}(r)$ for some $r>0$.

Next, we want to show that the ODE \eqref{dyn_sys_lambda_h} is
actually a gradient ascent of the Lagrangian function using the
envelope theorem {\color{black}from} mathematical economics
\citep{milgrom2002envelope}. The envelope theorem describes sufficient
conditions for the derivative of $L^*$ with respect to $\lambda$
{\color{black}to equal} the partial derivative of the objective function $L$ with respect to $\lambda$, holding $(\theta,\nu)$ at its local optimum $(\theta,\nu)=(\theta^\ast(\lambda),\nu^\ast(\lambda))$.  We will show that  $ \nabla_\lambda L^\ast(\lambda)$ coincides with $\nabla_\lambda L(\nu,\theta,\lambda)\vert_{\theta=\theta^\ast(\lambda),\nu=\nu^\ast(\lambda)}$ as follows.

\begin{theorem}\label{thm:envelop}
The value function $L^*$ is absolutely continuous. Furthermore,
\begin{equation}\label{envel_eq}
 L^*(\lambda)=L^*(0)+\int_{0}^{\lambda}\nabla_{\lambda'} L(\nu,\theta,\lambda')\Big\vert_{\theta=\theta^*(s),\nu=\nu^*(s),\lambda'=s}ds,\,\, \lambda\geq 0.  
 \end{equation}
 \end{theorem}
 \begin{prooff}
The proof follows from analogous arguments {\color{black}to} Lemma 4.3 in \citet{borkar2005actor}.
From the definition of $L^*$, observe that for any $\lambda^{\prime
},\lambda^{\prime \prime }\geq 0$ with $\lambda^{\prime }<\lambda^{\prime \prime }$,
\[
\begin{split}
 |L^*(\lambda^{\prime \prime })-L^*(\lambda^{\prime })| \leq & \sup_{\theta\in\Theta,\nu\in[-\frac{D_{\max}}{1-\gamma},\frac{D_{\max}}{1-\gamma}]}|L(\nu,\theta,\lambda^{\prime\prime })-L(\nu,\theta,\lambda^{\prime })|\\ 
=&\sup_{\theta\in\Theta,\nu\in[-\frac{D_{\max}}{1-\gamma},\frac{D_{\max}}{1-\gamma}]}\left\vert \int_{\lambda^{\prime }}^{\lambda^{\prime \prime
}}\nabla_\lambda L(\nu,\theta,s)ds\right\vert \\
\leq& \int_{\lambda^{\prime }}^{\lambda^{\prime \prime
}}\sup_{\theta\in\Theta,\nu\in [\frac{-D_{\max}}{1-\gamma},\frac{D_{\max}}{1-\gamma}]}\left|\nabla_\lambda L(\nu,\theta,s)\right|ds\leq \frac{3 D_{\max}}{(1-\alpha)(1-\gamma)}(\lambda^{\prime\prime}-\lambda^\prime).
\end{split}
\]
This implies that $L^*$ is absolutely continuous. Therefore, $L^*$ is continuous everywhere and differentiable almost everywhere.

By the Milgrom--Segal envelope theorem {\color{black}in} mathematical economics (Theorem 1 of \citet{milgrom2002envelope}), one concludes that the derivative of $L^*(\lambda)$ coincides with the derivative of $L(\nu,\theta,\lambda)$ at the point of differentiability $\lambda$ and $\theta=\theta^*(\lambda)$, $\nu=\nu^*(\lambda)$. Also since $L^*$ is absolutely continuous, the limit of $ (L^*(\lambda)-L^*(\lambda^\prime))/(\lambda-\lambda^\prime)$ at $\lambda\uparrow \lambda^\prime$ (or $\lambda\downarrow \lambda^\prime$) coincides with the lower/upper directional derivatives if $\lambda^\prime$ is  a point of non-differentiability. Thus, there is only a countable number of non-differentiable points in $L^*$ and the set of non-differentiable points of $L^*$ has measure zero. Therefore, expression \eqref{envel_eq} holds and one concludes that  $\nabla_\lambda L^\ast(\lambda)$ coincides with $\nabla_\lambda L(\nu,\theta,\lambda)\vert_{\theta=\theta^*(\lambda),\nu=\nu^*(\lambda)}$. 
\end{prooff}

Before getting into the main result, we have the following technical
proposition whose proof directly follows from the definition of
$\log\mathbb{P}_\theta(\xi)$ and
{\color{black}Assumption~\ref{ass:differentiability} that
  $\nabla_\theta\mu(a_k|x_k;\theta)$ is Lipschitz in $\theta$.}
\begin{proposition}\label{L_lips}
$\nabla_\theta L(\nu,\theta,\lambda)$ is Lipschitz in $\theta$.
\end{proposition}
\begin{prooff}
Recall that 
\[
\nabla_\theta L(\nu,\theta,\lambda)=\sum_{\xi} \mathbb{P}_\theta(\xi)\cdot\nabla_\theta\log\mathbb{P}_\theta(\xi)\left({\color{black}\mathcal G}(\xi) + \frac{\lambda}{1-\alpha}\big({\color{black}\mathcal J}(\xi)-\nu\big)\mathbf{1}\big\{{\color{black}\mathcal J}(\xi)\geq\nu\big\}\right)
\]
and $\nabla_\theta\log\mathbb{P}_\theta(\xi)=\sum_{k=0}^{T-1}\nabla_\theta\mu(a_k|x_k;\theta)/\mu(a_k|x_k;\theta)$ whenever $\mu(a_k|x_k;\theta)\in ( 0,1]$. Now Assumption (A1) implies that $\nabla_\theta\mu(a_k|x_k;\theta)$ is a Lipschitz function in $\theta$ for any $a\in\A$ and $k\in\{0,\ldots,T-1\}$ and $\mu(a_k|x_k;\theta)$ is differentiable in $\theta$. Therefore, by recalling that 
\[
\mathbb{P}_\theta(\xi)=\prod_{k=0}^{T-1} P(x_{k+1}|x_k,a_k)\mu(a_k|x_k;\theta)\mathbf 1\{x_0=x^0\}
\]
and by combining these arguments and noting that the sum of products of Lipschitz functions is Lipschitz, one concludes that $\nabla_\theta L(\nu,\theta,\lambda)$ is Lipschitz in $\theta$.
\end{prooff}
\begin{remark}\label{remark_lip} {\color{black}The fact that}
$\nabla_\theta L(\nu,\theta,\lambda)$ is Lipschitz in $\theta$ implies that 
\[
\|\nabla_\theta L(\nu,\theta,\lambda)\|^2\leq 2(\|\nabla_\theta L(\nu,\theta_0,\lambda)\|+\|\theta_0\|)^2+2\|\theta\|^2
\]
which further implies that
\[
\|\nabla_\theta L(\nu,\theta,\lambda)\|^2\leq K_1(1+\|\theta\|^2).
\]
for $K_1=2\max(1,(\|\nabla_\theta L(\nu,\theta_0,\lambda)\|+\|\theta_0\|)^2)>0$. Similarly, {\color{black}the fact that} $\nabla_\theta\log\mathbb{P}_{\theta}(\xi)$ is Lipschitz implies that
\[
\|\nabla_\theta\log\mathbb{P}_{\theta}(\xi)\|^2\leq K_2(\xi)(1+\|\theta\|^2)
\]
for a positive random variable $K_2(\xi)$. Furthermore, since $T<\infty $ w.p. $1$, $\mu(a_k|x_k;\theta)\in ( 0,1]$ and $\nabla_\theta\mu(a_k|x_k;\theta)$ is Lipschitz for any $k<T$, $K_2(\xi)<\infty$ w.p. $1$.
\end{remark}
\begin{remark}\label{remark:second_d_bounded}
For any given $\theta\in \Theta$, $\lambda\geq 0$, and $g(\nu)\in\partial_\nu L( \nu,\theta,\lambda)$, we have 
\begin{equation}
\label{eq:remark2}
|g(\nu)|\leq 3\lambda(1+|\nu|)/(1-\alpha). 
\end{equation}
To see this, recall that {the set of $g(\nu)\in\partial_\nu L(\nu,\theta,\lambda)$ can be parameterized by $q\in[0,1]$} as
\[
g(\nu;q)=-\frac{\lambda}{(1-\alpha)}\sum_{\xi}\mathbb P_\theta(\xi)\mathbf 1\left\{{\color{black}\mathcal J}(\xi)>\nu\right\}-\frac{\lambda q}{1-\alpha}\sum_{\xi}\mathbb P_\theta(\xi)\mathbf 1\left\{{\color{black}\mathcal J}(\xi) = \nu\right\}+\lambda.
\]
It is obvious that
$\left|\mathbf 1\left\{{\color{black}\mathcal J}(\xi) = \nu\right\}\right|,\left|\mathbf 1\left\{{\color{black}\mathcal J}(\xi)>\nu\right\}\right|\leq 1+|\nu|$. Thus, 
$\left|\sum_{\xi}\mathbb P_\theta(\xi)\mathbf 1\left\{{\color{black}\mathcal J}(\xi)>\nu\right\}\right|\leq \sup_{\xi}\left|\mathbf 1\left\{{\color{black}\mathcal J}(\xi)> \nu\right\}\right| \leq  1+|\nu|$, and
$\left|\sum_{\xi}\mathbb P_\theta(\xi)\mathbf 1\left\{{\color{black}\mathcal J}(\xi) = \nu\right\}\right| \leq  1+|\nu|$. Recalling {\color{black}that} $0<(1-q),\,(1-\alpha)<1$, these arguments imply the claim of \eqref{eq:remark2}.
\end{remark}

We are now in a position to prove the convergence analysis of Theorem~\ref{thm:converge_h}. \\
\begin{prooff}[{Proof of Theorem \ref{thm:converge_h}}]
We split the proof into the following four steps:

%%%%%%%%%%%%%%%%%%%%%%%%%%%%%%%%%%%%%%%%%%%%%%%%%%%%%%%%%%%%%%

\noindent\paragraph{Step~1 (Convergence of $\nu$-update)} 
Since $\nu$ converges {\color{black}on} a faster time scale than $\theta$ and $\lambda$, {\color{black}according to Lemma 1 in Chapter 6 of \citet{borkar2008stochastic}, one can analyze the convergence properties of $\nu$ in the following update rule for arbitrary quantities of $\theta$ and $\lambda$ (i.e., here we have $\theta=\theta_k$ and $\lambda=\lambda_k$): }
\begin{equation}\label{update_s_h_2}
\nu_{k+1}=\Gamma_{\mathcal{N}}\left(\nu_{k}+\zeta_3(k)\left(\frac{\lambda}{(1-\alpha)N}\sum_{j=1}^N\mathbf{1}\big\{{\color{black}\mathcal J}(\xi_{j,k})\geq \nu_k\big\}-\lambda+\delta \nu_{k+1}\right)\right),
\end{equation}
and {\color{black}the Martingale difference term with respect to $\nu$ is given by}
\begin{equation}\label{eq:MG_diff_nu_h}
\delta \nu_{k+1}=\frac{\lambda}{1-\alpha}\left(-\frac{1}{N}\sum_{j=1}^N\mathbf{1}\big\{{\color{black}\mathcal J}(\xi_{j,k})\geq \nu_k\big\}+\sum_{\xi}\mathbb P_\theta(\xi) \mathbf 1\{{\color{black}\mathcal J}(\xi)\geq \nu_{k}\}\right).
\end{equation}
First, one can show that $\delta\nu_{k+1}$ is square integrable, i.e., 
\[
\mathbb E[ \|\delta\nu_{k+1}\|^2\mid F_{\nu,k}]\leq 4\left(\frac{\lambda_{\max}}{1-\alpha}\right)^2
\]
where $\mathcal F_{\nu,k}= \sigma\big(\nu_m,\,\delta \nu_m,\,m\leq k\big)$ is the filtration of $\nu_k$ generated by different independent trajectories. 

Second, since the history trajectories are generated based on the sampling probability mass function $\mathbb{P}_{\theta}(\xi)$, expression \eqref{eq:L_nu} implies that $\mathbb E\left[\delta\nu_{k+1}\mid \mathcal F_{\nu,k}\right]=0$.  Therefore, the $\nu$-update is a stochastic approximation of the ODE~\eqref{dyn_sys_nu_h} with a Martingale difference error term, i.e.,
\[
\frac{\lambda}{1-\alpha}\sum_{\xi}\mathbb P_\theta(\xi) \mathbf 1\{{\color{black}\mathcal J}(\xi)\geq \nu_{k}\}-\lambda\in-\partial_\nu L(\nu,\theta,\lambda)\vert_{\nu=\nu_{k}}.
\]
Then one can invoke Corollary~4 in Chapter~5 of~\citet{borkar2008stochastic} (stochastic approximation theory for non-differentiable systems) to show that the sequence $\{\nu_k\},\;\nu_k\in [-\frac{D_{\max}}{1-\gamma},\frac{D_{\max}}{1-\gamma}]$ converges almost surely to a fixed point $\nu^*\in [-\frac{D_{\max}}{1-\gamma},\frac{D_{\max}}{1-\gamma}]$ of {\color{black}the} differential inclusion~\eqref{dyn_sys_nu_h}, where 
\[
\nu^\ast\in N_{c}:=\left\{\nu\in \left[-\frac{D_{\max}}{1-\gamma},\frac{D_{\max}}{1-\gamma}\right]:\Upsilon_\nu[-g(\nu)]=0,\, g(\nu)\in \partial_\nu L(\nu,\theta,\lambda)\right\}.
\]
To justify the assumptions of this {\color{black}corollary}, 1) from
Remark~\ref{remark:second_d_bounded}, the Lipschitz property is
satisfied, i.e., $\sup_{g(\nu)\in\partial_\nu
  L(\nu,\theta,\lambda)}|g(\nu)|\leq 3\lambda(1+|\nu|)/(1-\alpha)$, 2)
$[-\frac{D_{\max}}{1-\gamma},\frac{D_{\max}}{1-\gamma}]$ and
$\partial_\nu L(\nu,\theta,\lambda)$ are convex compact sets by
definition, which implies $\{(\nu,g(\nu))\mid g(\nu)\in\partial_\nu
L(\nu,\theta,\lambda)\}$ is a closed set, and further implies
$\partial_\nu L(\nu,\theta,\lambda)$ is an upper semi-continuous set
valued mapping, 3) the step-size rule follows from Assumption~\ref{ass:steps_pg}, 4) the Martingale difference assumption follows from~\eqref{eq:MG_diff_nu_h}, and 5) $\nu_k\in[-\frac{D_{\max}}{1-\gamma},\frac{D_{\max}}{1-\gamma}]$, $\forall i$ implies that $\sup_{k}\|\nu_k\|<\infty$ almost surely. 

Consider the ODE {\color{black}for} $\nu\in \reals$ in~\eqref{dyn_sys_nu_h}, we define the set-valued derivative of $L$ as follows:
\[
D_t L(\nu,\theta,\lambda)=\big\{g(\nu)\Upsilon_\nu\big[-g(\nu)\big] \mid \forall g(\nu)\in \partial_\nu L(\nu,\theta,\lambda)\big\}.
\]
One {\color{black}can} conclude that 
\[
\max_{g(\nu)}D_t L(\nu,\theta,\lambda)= \max\big\{g(\nu)\Upsilon_\nu\big[-g(\nu)\big] \mid g(\nu)\in \partial_\nu L(\nu,\theta,\lambda)\big\}.
\]
We now show that $\max_{g(\nu)}D_t L(\nu,\theta,\lambda)\leq 0$ and
this quantity is non-zero if $\Upsilon_\nu\big[-g(\nu)\big]\neq 0$ for
every $g(\nu)\in \partial_\nu L(\nu,\theta,\lambda)$ by considering
{\color{black}three cases. To distinguish the latter two cases, we
  need to define,
\[ {\color{black}\mathcal J}(\nu):=\left\{g(\nu)\in \partial L_\nu(\nu,\theta,\lambda)\Big|
  \,\forall \eta_0>0, \,\exists
  \eta\in(0,\eta_0] \text{ such that }\theta-\eta
  g(\nu)\not\in\left[-\frac{D_{\max}}{1-\gamma},\frac{D_{\max}}{1-\gamma}\right] \right\}. \]}

\noindent \emph{Case 1: $\nu\in (-\frac{D_{\max}}{1-\gamma},\frac{D_{\max}}{1-\gamma})$.}\\
For every $g(\nu)\in \partial_\nu L(\nu,\theta,\lambda)$, there exists a sufficiently small $\eta_{0}>0$ such that $\nu-\eta_0g(\nu)\in [-\frac{D_{\max}}{1-\gamma},\frac{D_{\max}}{1-\gamma}]$ and 
\[
\Gamma_{\mathcal{N}}\big(\theta-\eta_0g(\nu)\big)-\theta=-\eta_0g(\nu).
\]
Therefore, the definition of $\Upsilon_\theta[-g(\nu)]$ implies 
\begin{equation}\label{scen_1_h}
\max_{g(\nu)}D_t L(\nu,\theta,\lambda)=\max\big\{-g^2(\nu) \mid g(\nu)\in \partial_\nu L(\nu,\theta,\lambda)\big\}\leq 0.
\end{equation}
The maximum is attained because $\partial_\nu L(\nu,\theta,\lambda)$ is a convex compact set and $g(\nu)\Upsilon_\nu\big[-g(\nu)\big]$ is a continuous function.
At the same time, we have $\max_{g(\nu)}D_t L(\nu,\theta,\lambda)<0$ whenever $0\not\in \partial_\nu L(\nu,\theta,\lambda)$.

\noindent \emph{Case 2: {\color{black}$\nu\in
    \{-\frac{D_{\max}}{1-\gamma},\frac{D_{\max}}{1-\gamma}\}$ and
    ${\color{black}\mathcal J}(\nu)$ is empty}.}\\
The condition $\nu-\eta g(\nu)\in [-\frac{D_{\max}}{1-\gamma},\frac{D_{\max}}{1-\gamma}]$ implies that
\[
\Upsilon_\nu\big[-g(\nu)\big]=-g(\nu).
\]
Then we obtain
\begin{equation}\label{scen_2_h}
\max_{g(\nu)}D_t L(\nu,\theta,\lambda)=\max\big\{-g^2(\nu) \mid g(\nu)\in \partial_\nu L(\nu,\theta,\lambda)\big\}\leq 0.
\end{equation}
Furthermore, we have $\max_{g(\nu)}D_t L(\nu,\theta,\lambda)<0$ whenever $0\not\in \partial_\nu L(\nu,\theta,\lambda)$.
 
\noindent \emph{Case 3: {\color{black}$\nu\in
    \{-\frac{D_{\max}}{1-\gamma},\frac{D_{\max}}{1-\gamma}\}$ and
    ${\color{black}\color{black}\mathcal J}(\nu)$ is nonempty}}.\\
First, consider any $g(\nu)\in {\color{black}\mathcal J}(\nu)$. For any $\eta>0$, define $\nu_\eta:=\nu-\eta g(\nu)$. The above condition implies that when $0<\eta\rightarrow 0$, 
$\Gamma_{\mathcal{N}}\big[\nu_\eta\big]$ is the projection of $\nu_\eta$ to the tangent space of $[-\frac{D_{\max}}{1-\gamma},\frac{D_{\max}}{1-\gamma}]$. For any element $\hat\nu\in[-\frac{D_{\max}}{1-\gamma},\frac{D_{\max}}{1-\gamma}]$, since the set
$\{\nu\in[-\frac{D_{\max}}{1-\gamma},\frac{D_{\max}}{1-\gamma}]:\|\nu-\nu_\eta\|_2\leq \|\hat\nu-\nu_\eta\|_2\}$ is compact, the projection of $\nu_\eta$ on $[-\frac{D_{\max}}{1-\gamma},\frac{D_{\max}}{1-\gamma}]$ exists. Furthermore, since $f(\nu):=\frac{1}{2}(\nu-\nu_\eta)^2$ is a strongly convex function and $\nabla f(\nu)=\nu-\nu_\eta$, by {\color{black}the} first order optimality condition, one obtains
\[
\nabla f(\nu_\eta^\ast)(\nu- \nu_\eta^\ast)=(\nu_\eta^\ast-\nu_\eta)(\nu- \nu_\eta^\ast) \geq 0, \quad \forall \nu \in \left[-\frac{D_{\max}}{1-\gamma},\frac{D_{\max}}{1-\gamma}\right]
\]
where $\nu_\eta^\ast$ is {\color{black}the} unique projection of $\nu_\eta$ (the projection is unique because $f(\nu)$ is strongly convex and $[-\frac{D_{\max}}{1-\gamma},\frac{D_{\max}}{1-\gamma}]$ is a convex compact set). Since the projection (minimizer) is unique, the above equality holds if and only if $\nu=\nu_\eta^\ast$.

Therefore, for any $\nu\in[-\frac{D_{\max}}{1-\gamma},\frac{D_{\max}}{1-\gamma}]$ and $\eta>0$,
\[
\begin{split}
&g(\nu)\Upsilon_\nu\big[-g(\nu)\big]=g(\nu)\left(\lim_{0<\eta\rightarrow 0}\frac{\nu_\eta^\ast-\nu}{\eta}\right)\\
=&\left(\lim_{0<\eta\rightarrow 0}\frac{\nu-\nu_\eta}{\eta}\right)\left(\lim_{0<\eta\rightarrow 0}\frac{\nu_\eta^\ast-\nu}{\eta}\right)=\lim_{0<\eta\rightarrow 0}\frac{-\|\nu_\eta^\ast-\nu\|^2}{\eta^2}+
\lim_{0<\eta\rightarrow 0}\big(\nu_\eta^\ast-\nu_\eta\big)\left(\frac{\nu_\eta^\ast-\nu}{\eta^2}\right)\leq 0.
\end{split}
\]
Second, for any $g(\nu)\in \partial_\nu L(\nu,\theta,\lambda)\cap {\color{black}\mathcal J}(\nu)^c$, one obtains $\nu-\eta g(\nu)\in [-\frac{D_{\max}}{1-\gamma},\frac{D_{\max}}{1-\gamma}]$, for any $\eta\in(0,\eta_0]$ and some $\eta_0>0$. In this case, the arguments follow from case 2 and the following expression holds: $\Upsilon_\nu\big[-g(\nu)\big]=-g(\nu)$.

Combining these arguments, one concludes that 
\begin{equation}\label{scen_3_h}
\small
\begin{split}
&\max_{g(\nu)}D_t L(\nu,\theta,\lambda)\\
\leq &\max\left\{\max\big\{g(\nu)\;\Upsilon_\nu\big[-g(\nu)\big] \mid g(\nu)\in {\color{black}\mathcal J}(\nu)\big\},\max\big\{-g^2(\nu) \mid g(\nu)\in \partial_\nu L(\nu,\theta,\lambda)\cap {\color{black}\mathcal J}(\nu)^c\big\}\right\}\leq 0.
\end{split}
\end{equation}
This quantity is non-zero whenever $0\not\in\{g(\nu)\;\Upsilon_\nu\big[-g(\nu)\big]\mid \forall g(\nu)\in \partial_\nu L(\nu,\theta,\lambda)\}$ (this is because, for any $g(\nu)\in \partial_\nu L(\nu,\theta,\lambda)\cap {\color{black}\mathcal J}(\nu)^c$, one obtains $g(\nu)\;\Upsilon_\nu\big[-g(\nu)\big]=-g(\nu)^2$). 
Thus, by similar arguments one may conclude that $\max_{g(\nu)}D_t L(\nu,\theta,\lambda)\leq 0$ and it is non-zero if $\Upsilon_\nu\big[-g(\nu)\big]\neq 0$ for every $g(\nu)\in \partial_\nu L(\nu,\theta,\lambda)$. 
 
Now for any given $\theta$ and $\lambda$, define the following Lyapunov function
\[
\mathcal L_{\theta,\lambda}(\nu)=L(\nu,\theta,\lambda)-L(\nu^*,\theta,\lambda)
\]
where $\nu^*$ is a minimum point (for any given $(\theta,\lambda)$, $L$ is a convex function in $\nu$). 
Then $\mathcal L_{\theta,\lambda}(\nu)$ is a positive definite function, i.e., $\mathcal L_{\theta,\lambda}(\nu)\geq 0$.
On the other hand, by the definition of a minimum point, one easily obtains $0\in\{g(\nu^*)\;\Upsilon_\nu\big[-g(\nu^*)\big]\vert_{\nu=\nu^*}\mid \forall g(\nu^*)\in \partial_\nu L(\nu,\theta,\lambda)\vert_{\nu=\nu^*}\}$ which means that $\nu^\ast$ is also a stationary point, i.e., $\nu^\ast\in N_c$.

Note that $\max_{g(\nu)}D_t \mathcal L_{\theta,\lambda}(\nu)=\max_{g(\nu)}D_t L(\nu,\theta,\lambda)\leq 0$ and this quantity is non-zero if $\Upsilon_\nu\big[-g(\nu)\big]\neq 0$ for every $g(\nu)\in \partial_\nu L(\nu,\theta,\lambda)$. {\color{black}Therefore, by the Lyapunov theory for asymptotically stable differential inclusions (see Theorem 3.10  and Corollary 3.11 in \citet{benaim2006stochastic}, where the Lyapunov function $\mathcal L_{\theta,\lambda}(\nu)$ satisfies Hypothesis 3.1 and the property in \eqref{scen_3_h} is equivalent to Hypothesis 3.9 in the reference), the above arguments imply that with any initial condition $\nu(0)$, the state trajectory $\nu(t)$ of \eqref{dyn_sys_nu_h} converges to $\nu^\ast$, i.e., $L(\nu^\ast,\theta,\lambda)\leq L(\nu(t),\theta,\lambda)\leq L(\nu(0),\theta,\lambda)$ for any $t\geq 0$. }

{\color{black}As stated earlier,} the sequence $\{\nu_k\},\;\nu_k\in
[-\frac{D_{\max}}{1-\gamma},\frac{D_{\max}}{1-\gamma}]$ {\color{black}constitutes a
stochastic approximation to the differential
inclusion~\eqref{dyn_sys_nu_h}, and thus converges almost surely its
solution} \citep{borkar2008stochastic}, which further converges almost
surely to $\nu^*\in N_c$. Also, it can be easily seen that $N_{c}$ is
a closed subset of the compact set
$[-\frac{D_{\max}}{1-\gamma},\frac{D_{\max}}{1-\gamma}]$,
{\color{black}and therefore a compact set itself}.

%%%%%%%%%%%%%%%%%%%%%%%%%%%%%%%%%%%%%%%%%%%%%%%%%%%%%%%%%%%%%%

\noindent\paragraph{Step~2 (Convergence of $\theta$-update)} 
Since $\theta$ converges {\color{black}on} a faster time scale than $\lambda$ and $\nu$ converges faster than $\theta$, {\color{black} according to Lemma 1 in Chapter 6 of \citet{borkar2008stochastic} one can prove convergence of the $\theta$ update for any arbitrary $\lambda$ (i.e., $\lambda=\lambda_k$). Furthermore, in the $\theta$-update, we have that $\|\nu_k-\nu^*(\theta_k)\|\rightarrow 0$ almost surely. By the continuity condition of $\nabla_\theta L(\nu,\theta,\lambda)$, this also implies $\|\nabla_\theta L(\nu,\theta,\lambda)\vert_{\theta=\theta_k,\nu=\nu_k}-\nabla_\theta L(\nu,\theta,\lambda)\vert_{\theta=\theta_k,\nu=\nu^*(\theta_k)}\|\rightarrow 0$}. Therefore, the $\theta$-update can be rewritten as a stochastic approximation, i.e., 
\begin{equation}\label{update_theta_h_2}
\theta_{k+1}=\Gamma_\Theta\left(\theta_{k}+\zeta_2(k)\bigg(-\nabla_\theta L(\nu,\theta,\lambda)\vert_{\theta=\theta_k,\nu=\nu^*(\theta_k)}+\delta\theta_{k+1}\bigg)\right),
\end{equation}
where 
\begin{equation}\label{eq:MG_diff_theta_h}
\begin{split}
\delta\theta_{k+1}=&\nabla_\theta L(\nu,\theta,\lambda)\vert_{\theta=\theta_k,\nu=\nu^*(\theta_k)}\!-\!\frac{1}{N}\sum_{j=1}^N\nabla_\theta\log\mathbb{P}_{\theta}(\xi_{j,k})\mid_{\theta=\theta_k} {\color{black}\mathcal G}(\xi_{j,k})\\
&- \frac{\lambda}{(1-\alpha)N}\sum_{j=1}^N\nabla_\theta\log\mathbb{P}_\theta(\xi_{j,k})\vert_{\theta=\theta_k}\big({\color{black}\mathcal J}(\xi_{j,k})-\nu^*(\theta_k)\big)\mathbf{1}\big\{{\color{black}\mathcal J}(\xi_{j,k})\geq\nu^*(\theta_k)\big\}\\
&{\color{black}+ \frac{\lambda}{(1-\alpha)N}\sum_{j=1}^N\nabla_\theta\log\mathbb{P}_\theta(\xi_{j,k})\vert_{\theta=\theta_k}\big(\nu^*(\theta_k)-\nu_k\big)\mathbf{1}\big\{{\color{black}\mathcal J}(\xi_{j,k})\geq\nu^*(\theta_k)\big\}}\\
&{\color{black}+ \frac{\lambda}{(1-\alpha)N}\sum_{j=1}^N\nabla_\theta\log\mathbb{P}_\theta(\xi_{j,k})\vert_{\theta=\theta_k}\big({\color{black}\mathcal J}(\xi_{j,k})-\nu_k\big)\left(\mathbf{1}\big\{{\color{black}\mathcal J}(\xi_{j,k})\geq\nu_k\big\}-\mathbf{1}\big\{{\color{black}\mathcal J}(\xi_{j,k})\geq\nu^*(\theta_k)\big\}\right)}.
\end{split}
\end{equation}

{\color{black}First, we consider the last two components in \eqref{eq:MG_diff_theta_h}. Recall that $\|\nu_k-\nu^*(\theta_k)\|\rightarrow 0$ almost surely. Furthermore by noticing that $\nabla_\theta\log\mathbb{P}_\theta(\xi_{j,k})$ is Lipschitz in $\theta$, $\theta$ lies in a compact set $\Theta$, both $\mathcal J(\xi_{j,k})$ and $\nu_k$ are bounded, and $\nu, \nu^*(\theta_k)$ lie in a compact set $\mathcal N$, one immediately concludes that as $i\rightarrow\infty$,
\begin{equation}\label{eq:tail_terms}
\begin{split}
&\big(\nu^*(\theta_k)-\nu_k\big)\mathbf{1}\big\{{\color{black}\mathcal J}(\xi_{j,k})\geq\nu^*(\theta_k)\big\}\rightarrow 0,\quad \text{almost surely}\\
&\big({\color{black}\mathcal J}(\xi_{j,k})-\nu_k\big)\left(\mathbf{1}\big\{{\color{black}\mathcal J}(\xi_{j,k})\geq\nu_k\big\}-\mathbf{1}\big\{{\color{black}\mathcal J}(\xi_{j,k})\geq\nu^*(\theta_k)\big\}\right)\rightarrow 0,\quad\text{almost surely}
\end{split}
\end{equation}
}
Second, one can show that $\delta\theta_{k+1}$ is square integrable, i.e., $\mathbb E[ \|\delta\theta_{k+1}\|^2\mid F_{\theta,k}]\leq K_k(1+\|\theta_k\|^2)$ for some $K_k>0$, where $\mathcal F_{\theta,k}= \sigma\big(\theta_m,\,\delta \theta_m,\,m\leq k\big)$ is the filtration of $\theta_k$ generated by different independent trajectories. To see this, notice that
\[
\begin{split}
 &\|\delta\theta_{k+1}\|^2\\
 \leq&  2 \left(\nabla_\theta L(\nu,\theta,\lambda)\vert_{\theta=\theta_k,\nu=\nu^*(\theta_k)}\right)^2+\frac{2}{N^2}\left(\frac{C_{\max}}{1-\gamma}+\frac{2\lambda D_{\max}}{(1-\alpha)(1-\gamma)}\right)^2\left(\sum_{j=1}^N\nabla_\theta\log\mathbb{P}_{\theta}(\xi_{j,k})\mid_{\theta=\theta_k}\right)^2\\
 \leq & 2 K_{1,k}(1+\|\theta_k\|^2)+\frac{2^{N}}{N^2}\left(\frac{C_{\max}}{1-\gamma}+\frac{2\lambda_{\max} D_{\max}}{(1-\alpha)(1-\gamma)}\right)^2 \left(\sum_{j=1}^N\left\|\nabla_\theta\log\mathbb{P}_{\theta}(\xi_{j,k})\mid_{\theta=\theta_k}\right\|^2\right)\\
 \leq & 2 K_{1,k}(1+\|\theta_k\|^2)+\frac{2^{N}}{N^2}\left(\frac{C_{\max}}{1-\gamma}+\frac{2\lambda_{\max} D_{\max}}{(1-\alpha)(1-\gamma)}\right)^2 \left(\sum_{j=1}^NK_2(\xi_{j,k})(1+\|\theta_k\|^2)\right)\\
 \leq & 2\!\left(K_{1,k}\!+\!\frac{2^{N-1}}{N}\left(\frac{C_{\max}}{1-\gamma}+\frac{2\lambda_{\max} D_{\max}}{(1-\alpha)(1-\gamma)}\right)^2\max_{1\leq j\leq N}K_2(\xi_{j,k})\right)\!(1\!+\!\|\theta_k\|^2).
 \end{split}
\]
The Lipschitz upper bounds are due to {\color{black}the} results in Remark \ref{remark_lip}. Since $K_2(\xi_{j,k})<\infty$ w.p. $1$, there exists $K_{2,k}<\infty$ such that $\max_{1\leq j\leq N}K_2(\xi_{j,k})\leq K_{2,k}$. By combining these results, one concludes that
$\mathbb E[ \|\delta\theta_{k+1}\|^2\mid F_{\theta,k}]\leq K_k(1\!+\!\|\theta_k\|^2)$ where 
\[
K_k=2\left(K_{1,k}\!+\!\frac{2^{N-1}K_{2,k}}{N}\left(\frac{C_{\max}}{1-\gamma}+\frac{2\lambda_{\max} D_{\max}}{(1-\alpha)(1-\gamma)}\right)^2\right)<\infty.
\]

Third, since the history trajectories are generated based on the
sampling probability mass function $\mathbb{P}_{\theta_k}(\xi)$,
expression \eqref{eq:L_theta} implies that $\mathbb
E\left[\delta\theta_{k+1}\mid \mathcal F_{\theta,k}\right]=0$.
Therefore, the $\theta$-update is a stochastic approximation of the
ODE~\eqref{dyn_sys_kheta_h} with a Martingale difference error
term. In addition, from the convergence analysis of
{\color{black}the} $\nu$-update, $\nu^\ast(\theta)$ is an
asymptotically stable equilibrium point {\color{black}for the
  sequence} $\{\nu_k\}$. From \eqref{eq:L_nu}, $\partial_{\nu}
L(\nu,\theta,\lambda)$ is a Lipschitz set-valued mapping in $\theta$
(since $\mathbb P_\theta(\xi)$ is Lipschitz in $\theta$),
{\color{black}and thus} it can be easily seen that $\nu^\ast(\theta)$ is a Lipschitz continuous mapping of $\theta$.

Now consider the continuous time {\color{black}dynamics} for $\theta\in \Theta$, {\color{black}given} in~\eqref{dyn_sys_kheta_h}. We may write
\begin{equation}\label{eq:lyap_kneq_1}
\frac{d L(\nu,\theta,\lambda)}{dt}\bigg\vert_{\nu=\nu^*(\theta)}=\big(\nabla_\theta L(\nu,\theta,\lambda)\vert_{\nu=\nu^*(\theta)}\big)^\top\;\Upsilon_\theta\big[-\nabla_\theta L(\nu,\theta,\lambda)\vert_{\nu=\nu^*(\theta)}\big].
\end{equation}
By considering the following cases, we now show that  ${d L(\nu,\theta,\lambda)}/{dt}\vert_{\nu=\nu^*(\theta)}\leq 0$ and this quantity is non-zero whenever $\left\|\Upsilon_\theta\left[-\nabla_\theta L(\nu,\theta,\lambda)\vert_{\nu=\nu^*(\theta)}\right]\right\|\neq 0$. \\

\noindent 
\emph{Case 1: When $\theta\in\Theta^\circ = \Theta\setminus\partial\Theta$.}\\
 Since $\Theta^\circ$ is the interior of the set $\Theta$ and $\Theta$ is a convex compact set, there exists a sufficiently small $\eta_{0}>0$ such that $\theta-\eta_0\nabla_\theta L(\nu,\theta,\lambda)\vert_{\nu=\nu^*(\theta)}\in \Theta$ and 
\[
\Gamma_\Theta\big(\theta-\eta_0\nabla_\theta L(\nu,\theta,\lambda)\vert_{\nu=\nu^*(\theta)}\big)-\theta=-\eta_0\nabla_\theta L(\nu,\theta,\lambda)\vert_{\nu=\nu^*(\theta)}.
\]
Therefore, the definition of $\Upsilon_\theta\big[-\nabla_\theta L(\nu,\theta,\lambda)\vert_{\nu=\nu^*(\theta)}\big]$ implies 
\begin{equation}\label{scen_1_h}
\frac{d L(\nu,\theta,\lambda)}{dt}\bigg\vert_{\nu=\nu^*(\theta)}=-\left\|\nabla_\theta L(\nu,\theta,\lambda)\vert_{\nu=\nu^*(\theta)}\right\|^2\leq 0.
\end{equation}
At the same time, we have ${d L(\nu,\theta,\lambda)}/{dt}\vert_{\nu=\nu^*(\theta)}<0$ whenever $\|\nabla_\theta L(\nu,\theta,\lambda)\vert_{\nu=\nu^*(\theta)}\|\neq 0$. \\

\noindent 
\emph{Case 2: When $\theta\in \partial\Theta$ and $\theta-\eta\nabla_\theta L(\nu,\theta,\lambda)\vert_{\nu=\nu^*(\theta)}\in \Theta$ for any $\eta\in(0,\eta_0]$ and some $\eta_0>0$.}\\
The condition $\theta-\eta\nabla_\theta L(\nu,\theta,\lambda)\vert_{\nu=\nu^*(\theta)}\in \Theta$ implies that
\[
\Upsilon_\theta\big[-\nabla_\theta L(\nu,\theta,\lambda)\vert_{\nu=\nu^*(\theta)}\big]=-\nabla_\theta L(\nu,\theta,\lambda)\vert_{\nu=\nu^*(\theta)}.
\]
Then we obtain
\begin{equation}\label{scen_1_h}
\begin{split}
\frac{d L(\nu,\theta,\lambda)}{dt}\bigg\vert_{\nu=\nu^*(\theta)}=-\left\|\nabla_\theta L(\nu,\theta,\lambda)\vert_{\nu=\nu^*(\theta)}\right\|^2\leq 0.
\end{split}
\end{equation}
Furthermore, ${d L(\nu,\theta,\lambda)}/{dt}\vert_{\nu=\nu^*(\theta)}<0$ when $\|\nabla_\theta L(\nu,\theta,\lambda)\vert_{\nu=\nu^*(\theta)}\|\neq 0$. \\

\noindent 
\emph{Case 3: When $\theta\in \partial\Theta$ and $\theta-\eta\nabla_\theta L(\nu,\theta,\lambda)\vert_{\nu=\nu^*(\theta)}\not\in \Theta$ for some $\eta\in(0,\eta_0]$ and any $\eta_0>0$.}\\
For any $\eta>0$, define $\theta_\eta:=\theta-\eta\nabla_\theta L(\nu,\theta,\lambda)\vert_{\nu=\nu^*(\theta)}$. The above condition implies that when $0<\eta\rightarrow 0$, 
$\Gamma_\Theta\big[\theta_\eta\big]$ is the projection of $\theta_\eta$ to the tangent space of $\Theta$. For any element $\hat\theta\in\Theta$, since the set
$\{\theta\in\Theta:\|\theta-\theta_\eta\|_2\leq
\|\hat\theta-\theta_\eta\|_2\}$ is compact, the projection of
$\theta_\eta$ on $\Theta$ exists. Furthermore, since
$f(\theta):=\frac{1}{2}\|\theta-\theta_\eta\|_2^2$ is a strongly
convex function and $\nabla f(\theta)=\theta-\theta_\eta$, by
{\color{black}the} first order optimality condition, one obtains
\[
\nabla f(\theta_\eta^\ast)^\top(\theta- \theta_\eta^\ast)=(\theta_\eta^\ast-\theta_\eta)^\top(\theta- \theta_\eta^\ast) \geq 0, \quad \forall \theta \in \Theta,
\]
where $\theta_\eta^\ast$ is {\color{black}the} unique projection of $\theta_\eta$ (the projection is unique because $f(\theta)$ is strongly convex and $\Theta$ is a convex compact set). Since the projection (minimizer) is unique, the above equality holds if and only if $\theta=\theta_\eta^\ast$.

Therefore, for any $\theta\in\Theta$ and $\eta>0$,
\[
\begin{split}
&\big(\nabla_\theta L(\nu,\theta,\lambda)\vert_{\nu=\nu^*(\theta)}\big)^\top\;\Upsilon_\theta\big[-\nabla_\theta L(\nu,\theta,\lambda)\vert_{\nu=\nu^*(\theta)}\big]=\big(\nabla_\theta L(\nu,\theta,\lambda)\vert_{\nu=\nu^*(\theta)}\big)^\top\left(\lim_{0<\eta\rightarrow 0}\frac{\theta_\eta^\ast-\theta}{\eta}\right)\\
=&\left(\lim_{0<\eta\rightarrow 0}\frac{\theta-\theta_\eta}{\eta}\right)^\top\left(\lim_{0<\eta\rightarrow 0}\frac{\theta_\eta^\ast-\theta}{\eta}\right)=\lim_{0<\eta\rightarrow 0}\frac{-\|\theta_\eta^\ast-\theta\|^2}{\eta^2}+
\lim_{0<\eta\rightarrow 0}\big(\theta_\eta^\ast-\theta_\eta\big)^\top\left(\frac{\theta_\eta^\ast-\theta}{\eta^2}\right)\leq 0.
\end{split}
\]
By combining these arguments, one concludes that ${d L(\nu,\theta,\lambda)}/{dt}\vert_{\nu=\nu^*(\theta)}\leq 0$ and this quantity is non-zero whenever $\left\|\Upsilon_\theta\left[-\nabla_\theta L(\nu,\theta,\lambda)\vert_{\nu=\nu^*(\theta)}\right]\right\|\neq 0$.

Now, for any given $\lambda$, define the  Lyapunov function
\[
\mathcal L_\lambda(\theta)=L(\nu^*(\theta),\theta,\lambda)-L(\nu^*(\theta^*),\theta^*,\lambda),
\]
where $\theta^\ast$ is a local minimum point. Then there exists a ball centered at $\theta^\ast$ with radius $r$ such that for any $\theta\in \mathcal B_{\theta^\ast}(r)$, $\mathcal L_\lambda(\theta)$ is a locally positive definite function, i.e., $\mathcal L_\lambda(\theta)\geq 0$. On the other hand, by the definition of a local minimum point, one obtains $\Upsilon_\theta[-\nabla_\theta L(\theta^*,\nu,\lambda)\vert_{\nu=\nu^*(\theta^*)}]\vert_{\theta=\theta^*}=0$ which means that $\theta^\ast$ is a stationary point, i.e., $\theta^\ast\in\Theta_c$.

Note that $d\mathcal L_\lambda(\theta(t))/dt=d
L(\theta(t),\nu^*(\theta(t)),\lambda)/dt\leq 0$ and the
time-derivative is non-zero whenever
$\left\|\Upsilon_\theta\left[-\nabla_\theta
    L(\nu,\theta,\lambda)\vert_{\nu=\nu^*(\theta)}\right]\right\|\neq
0$. Therefore, by {\color{black}the} Lyapunov theory for
asymptotically stable systems {\color{black} from Chapter 4 of \citet{khalil2002nonlinear}}, the above
arguments imply that with any initial condition $\theta(0)\in \mathcal
B_{\theta^\ast}(r)$, the state trajectory $\theta(t)$ of \eqref{dyn_sys_kheta_h} converges to $\theta^\ast$, i.e., $L(\theta^\ast,\nu^\ast(\theta^\ast),\lambda)\leq L(\theta(t),\nu^\ast(\theta(t)),\lambda)\leq L(\theta(0),\nu^\ast(\theta(0)),\lambda)$ for any $t\geq 0$.

Based on the above properties and noting that {1)} from Proposition \ref{L_lips}, $\nabla_\theta L(\nu,\theta,\lambda)$ is a Lipschitz function in $\theta$, {2)} the step-size rule follows from Assumption~\ref{ass:steps_pg}, {3)} expression \eqref{eq:MG_diff_theta_h} implies that $\delta\theta_{k+1}$ is a square integrable Martingale difference, and {4)} $\theta_k\in\Theta$, $\forall i$ implies that $\sup_{k}\|\theta_k\|<\infty$ almost surely, one can invoke Theorem~2 in {\color{black}Chapter~6 of~\citet{borkar2008stochastic}} (multi-time scale stochastic approximation theory) to show that the sequence $\{\theta_k\},\;\theta_k\in \Theta$ converges almost surely to the solution of {\color{black}the} ODE~\eqref{dyn_sys_kheta_h}, which further converges almost surely to $\theta^*\in \Theta$. 

%%%%%%%%%%%%%%%%%%%%%%%%%%%%%%%%%%%%%%%%%%%%%%%%%%%%%%%%%%%%%%%

\noindent\paragraph{Step 3 (Local Minimum)}
Now, we want to show that {\color{black}the sequence}
$\{\theta_k,\nu_k\}$ converges to a local minimum of
$L(\nu,\theta,\lambda)$ for {\color{black} any} fixed $\lambda$.
Recall {\color{black}that}  $\{\theta_k,\nu_k\}$ converges to $(\theta^\ast,\nu^\ast):=(\theta^\ast,\nu^\ast(\theta^\ast))$. 
Previous arguments on {\color{black}the} $(\nu,\theta)$-convergence imply that with any initial condition $(\theta(0),\nu(0))$, the state trajectories $\theta(t)$ and $\nu(t)$ of \eqref{dyn_sys_nu_h} and \eqref{dyn_sys_kheta_h} converge to the set of stationary points $(\theta^\ast,\nu^\ast)$ in the positive invariant set $\Theta_c\times N_c$ and $L(\theta^\ast,\nu^\ast,\lambda)\leq L(\theta(t),\nu^\ast(\theta(t)),\lambda) \leq L(\theta(0),\nu^\ast(\theta(0)),\lambda)\leq L(\theta(0),\nu(t),\lambda)\leq L(\theta(0),\nu(0),\lambda)$ for any $t\geq 0$. 

By contradiction, suppose $(\theta^\ast,\nu^\ast)$ is not a local minimum. Then there exists $(\bar\theta,\bar\nu)\in\Theta\times[-\frac{D_{\max}}{1-\gamma},\frac{D_{\max}}{1-\gamma}]\cap \mathcal B_{(\theta^\ast,\nu^\ast)}(r)$ such that 
\[
L(\bar\theta,\bar\nu,\lambda)=\min_{(\theta,\nu)\in\Theta\times
  [-\frac{D_{\max}}{1-\gamma},\frac{D_{\max}}{1-\gamma}]\cap \mathcal B_{(\theta^\ast,\nu^\ast)}(r)}L(\nu,\theta,\lambda).
\]
The minimum is attained by {\color{black}the} Weierstrass extreme value theorem. By putting $\theta(0)=\bar\theta$, the above arguments imply that
\[
L(\bar\theta,\bar\nu,\lambda)=\min_{(\theta,\nu)\in\Theta\times
  [-\frac{D_{\max}}{1-\gamma},\frac{D_{\max}}{1-\gamma}]\cap \mathcal B_{(\theta^\ast,\nu^\ast)}(r)}L(\nu,\theta,\lambda)<L(\theta^\ast,\nu^\ast,\lambda)\leq L(\bar\theta,\bar\nu,\lambda)
\]
which is a contradiction. Therefore, the stationary point $(\theta^\ast,\nu^\ast)$ is a local minimum of $L(\nu,\theta,\lambda)$ as well.

%%%%%%%%%%%%%%%%%%%%%%%%%%%%%%%%%%%%%%%%%%%%%%%%%%%%%%%%%%%%%%

\noindent\paragraph{Step 4 (Convergence of $\lambda$-update)} Since
{\color{black}the} $\lambda$-update converges in the slowest time scale, {\color{black} according to previous analysis, we have that $\|\theta_k-\theta^*(\nu^*(\lambda_k),\lambda_k)\|\rightarrow 0$, $\|\nu_k-\nu^*(\lambda_k)\|\rightarrow 0$ almost surely. By continuity of $\nabla_\lambda L(\nu,\theta,\lambda)$, we also have the following:
\[
\left\|\nabla_\lambda L(\nu,\theta,\lambda)\bigg\vert_{\theta=\theta^*(\lambda_k),\nu=\nu^*(\lambda_k),\lambda=\lambda_k}-\nabla_\lambda L(\nu,\theta,\lambda)\bigg\vert_{\theta=\theta_k,\nu=\nu_k,\lambda=\lambda_k}\right\|\rightarrow 0, \quad\text{almost surely}.
\]
}
Therefore, the $\lambda$-update rule can be re-written as follows:
\begin{equation}\label{lamda_up_h_conv}
\lambda_{k+1}=\Gamma_\Lambda\left(\lambda_k+\zeta_1(k)\bigg(\nabla_\lambda L(\nu,\theta,\lambda)\bigg\vert_{\theta=\theta^*(\lambda_k),\nu=\nu^*(\lambda_k),\lambda=\lambda_k}+\delta\lambda_{k+1}\bigg)\right)
\end{equation}
where 
\begin{equation}\label{eq:MG_diff_theta_h}
\begin{split}
\delta\lambda_{k+1}=&-\nabla_\lambda L(\nu,\theta,\lambda)\bigg\vert_{\theta=\theta^*(\lambda),\nu=\nu^*(\lambda),\lambda=\lambda_k}\!\!\!+\bigg(\nu^*(\lambda_k)+\frac{1}{1-\alpha} \frac{1}{N}\sum_{j=1}^N\big({\color{black}\mathcal J}(\xi_{j,k})- \nu^*(\lambda_k)\big)^+ -\beta\bigg)\\
&{\color{black}+(\nu_k-\nu^*(\lambda_k))+\frac{1}{1-\alpha} \frac{1}{N}\sum_{j=1}^N\left(\big({\color{black}\mathcal J}(\xi_{j,k})- \nu_k\big)^+-\big({\color{black}\mathcal J}(\xi_{j,k})- \nu^*(\lambda_k)\big)^+\right).}
\end{split}
\end{equation}

{\color{black}From the fact that $\|\theta_k-\theta^*(\nu^*(\lambda_k),\lambda_k)\|\rightarrow 0$ almost surely as $i\rightarrow\infty$, one can conclude that the last component of the above expression vanishes, i.e., both $\|\nu_k-\nu^*(\lambda_k)\|\rightarrow 0$ and $\|\big({\color{black}\mathcal J}(\xi_{j,k})- \nu_k\big)^+-\big({\color{black}\mathcal J}(\xi_{j,k})- \nu^*(\lambda_k)\big)^+\|\rightarrow 0$ almost surely.}
Moreover, from \eqref{eq:L_lambda}, {\color{black}we see} that $\nabla_\lambda
L(\nu,\theta,\lambda)$ is a constant function of $\lambda$. Similar to
{\color{black}the} $\theta$-update,  one can easily show that $\delta\lambda_{k+1}$ is square integrable, i.e.,
\[
\mathbb E[\|\delta\lambda_{k+1}\|^2\mid \mathcal F_{\lambda,k}]\leq 2\left(\beta+\frac{3D_{\max}}{(1-\gamma)(1-\alpha)}\right)^2,
\]
where $\mathcal F_{\lambda,k}=\sigma\big(\lambda_m,\,\delta \lambda_m,\,m\leq k\big)$ is the filtration of $\lambda$ generated by different independent trajectories.
Furthermore, expression \eqref{eq:L_lambda} implies that $\mathbb
E\left[\delta\lambda_{k+1}\mid \mathcal F_{\lambda,k}\right]=0$.
Therefore, the $\lambda$-update is a stochastic approximation of the
ODE~\eqref{dyn_sys_lambda_h} with a Martingale difference error
term. In addition, from the convergence analysis of {\color{black}the} $(\theta,\nu)$-update, $(\theta^\ast(\lambda),\nu^\ast(\lambda))$ is an asymptotically stable equilibrium point {\color{black}for the sequence} $\{\theta_k,\nu_k\}$. From \eqref{eq:L_theta}, $\nabla_{\theta} L(\nu,\theta,\lambda)$ is a linear mapping in $\lambda$, {and} $(\theta^\ast(\lambda),\nu^\ast(\lambda))$ is a Lipschitz continuous mapping of $\lambda$.

Consider the ODE {\color{black}for} $\lambda\in [0,\lambda_{\max}]$ in~\eqref{dyn_sys_lambda_h}. Analogous to the arguments {\color{black}for} the $\theta$-update, we {\color{black}can} write
\[
\frac{d (-L(\nu,\theta,\lambda))}{dt}\bigg\vert_{\theta=\theta^*(\lambda),\nu=\nu^\ast(\lambda)} =-\nabla_\lambda L(\nu,\theta,\lambda)\bigg\vert_{\theta=\theta^*(\lambda),\nu=\nu^\ast(\lambda)}\!\!\Upsilon_\lambda\left[\nabla_\lambda L(\nu,\theta,\lambda)\bigg\vert_{\theta=\theta^*(\lambda),\nu=\nu^\ast(\lambda)}\right],
\]
and show that $-{d
  L(\nu,\theta,\lambda)}/{dt}\vert_{\theta=\theta^*(\lambda),\nu=\nu^\ast(\lambda)}\leq
0${.} This quantity is non-zero whenever 
\[
\left\|\Upsilon_\lambda\left[{ d L(\nu,\theta,\lambda)}/{d\lambda}\vert_{\theta=\theta^*(\lambda),\nu=\nu^\ast(\lambda)}\right]\right\|\neq 0.
\] 

Consider the Lyapunov function 
\[
\mathcal L(\lambda)=-L(\theta^*(\lambda),\nu^*(\lambda),\lambda)+L(\theta^*(\lambda^*),\nu^*(\lambda^*),\lambda^*)
\]
where $\lambda^\ast$ is a local maximum point. Then there exists a ball centered at $\lambda^\ast$ with radius $r$ such that for any $\lambda\in \mathcal B_{\lambda^\ast}(r)$, $\mathcal L(\lambda)$ is a locally positive definite function, i.e., $\mathcal L(\lambda)\geq 0$. On the other hand, by the definition of a local maximum point, one obtains 
\[
\Upsilon_\lambda\left[{ d L(\nu,\theta,\lambda)}/{d\lambda}\vert_{\theta=\theta^*(\lambda),\nu=\nu^\ast(\lambda),\lambda=\lambda^*}\right]\vert_{\lambda=\lambda^*}=0
\]
which means that $\lambda^\ast$ is also a stationary point, i.e., $\lambda^\ast\in\Lambda_c$.
Since 
\[
\frac{d\mathcal L(\lambda(t))}{dt}=-\frac{d L(\theta^*(\lambda(t)),\nu^*(\lambda(t)),\lambda(t))}{dt}\leq 0
\] 
and the  time-derivative is non-zero whenever $\left\|\Upsilon_\lambda[\nabla_\lambda L(\nu,\theta,\lambda)\mid_{\nu=\nu^\ast(\lambda),\theta=\theta^*(\lambda)}]\right\|\neq 0$, {\color{black}the} Lyapunov theory for asymptotically stable systems implies that $\lambda(t)$ converges to $\lambda^*$.  

{\color{black}Given the above results} and noting that the step size
rule {\color{black}is selected according to} Assumption~\ref{ass:steps_pg}, one can apply the multi-time scale stochastic approximation theory (Theorem 2 in Chapter 6 of \citet{borkar2008stochastic}) to show that the sequence $\{\lambda_k\}$ converges almost surely to the solution of {\color{black}the} ODE~\eqref{dyn_sys_lambda_h}, which further converges almost surely to $\lambda^*\in [0,\lambda_{\max}]$. Since $[0,\lambda_{\max}]$ is a compact set, following the same lines of arguments and recalling the envelope theorem (Theorem \ref{thm:envelop}) for local {\color{black}optima}, one further concludes that $\lambda^\ast$ is a local maximum of $L(\theta^\ast(\lambda),\nu^\ast(\lambda),\lambda)=L^\ast(\lambda)$.

%%%%%%%%%%%%%%%%%%%%%%%%%%%%%%%%%%%%%%%%%%%%%%%%%%%%%%%%%%%%%%

\noindent\paragraph{Step 5 {\color{black}(Local Saddle Point)}} 
By letting $\theta^*=\theta^*\big(\nu^*(\lambda^*),\lambda^*\big)$ and
$\nu^*=\nu^*(\lambda^*)$, {\color{black}we will show that  $(\theta^*,\nu^*,\lambda^*)$ is a local saddle point of the Lagrangian function $L(\nu,\theta,\lambda)$ if $\lambda^\ast\in[0,\lambda_{\max})$, and thus by the local saddle
  point theorem, $\theta^*$ is a locally optimal solution for the CVaR-constrained optimization.}

Suppose the sequence $\{\lambda_k\}$ generated from \eqref{lamda_up_h_conv} converges to a stationary point $\lambda^\ast\in[0,\lambda_{\max})$. Since step~3 implies that $(\theta^*,\nu^*)$ is a local minimum of $L(\nu,\theta,\lambda^\ast)$ over {\color{black}the} feasible set $(\theta,\nu)\in\Theta\times[-\frac{D_{\max}}{1-\gamma},\frac{D_{\max}}{1-\gamma}]$, there exists a $r>0$ such that
\[
L(\theta^*,\nu^*,\lambda^*)\leq L(\nu,\theta,\lambda^*),\quad\forall
(\theta,\nu)\in
\Theta\times\left[-\frac{D_{\max}}{1-\gamma},\frac{D_{\max}}{1-\gamma}\right]\cap
\mathcal B_{(\theta^\ast,\nu^\ast)}(r).
\]

In order to complete the proof, we must show
\begin{equation}\label{prop_1}
\nu^*+\frac{1}{1-\alpha}\expec\left[\big({\color{black}\mathcal J}^{\theta^*}(x^0)- \nu^*\big)^+\right] \leq  \beta,
\end{equation}
and
\begin{equation}\label{prop_2}
\lambda^* \left( \nu^*+\frac{1}{1-\alpha}\expec\left[\big({\color{black}\mathcal J}^{\theta^*}(x^0)- \nu^*\big)^+\right] - \beta\right)=0.
\end{equation}
These two equations imply 
\[
\begin{split}
L(\theta^*,\nu^*,\lambda^*)=&V^{\theta^*}(x^0) \!+\!\lambda^*\left(\nu^* + \frac{1}{1-\alpha}\expec\left[\big({\color{black}\mathcal J}^{\theta^*}(x^0)- \nu^*\big)^+\right] - \beta\right) \\
=&V^{\theta^*}(x^0) \\
\geq &V^{\theta^*}(x^0) \!+\!\lambda  \left(\nu^*+\frac{1}{1-\alpha}\expec\left[\big({\color{black}\mathcal J}^{\theta^*}(x^0)- \nu^*\big)^+\right] - \beta\right)=L(\theta^*,\nu^*,\lambda), 
\end{split}
\]
which further implies that $(\theta^*,\nu^*,\lambda^*)$ is a saddle point of $L(\nu,\theta,\lambda)$. We now show that \eqref{prop_1} and~\eqref{prop_2} hold. 

Recall that 
\[
\Upsilon_\lambda\left[\nabla_\lambda L(\nu,\theta,\lambda)\vert_{\theta=\theta^*(\lambda),\nu=\nu^*(\lambda),\lambda=\lambda^*}\right]\vert_{\lambda=\lambda^*}=0.
\]
We show \eqref{prop_1} by contradiction. Suppose 
\[
\nu^*+\frac{1}{1-\alpha}\expec\left[\big({\color{black}\mathcal J}^{\theta^*}(x^0)- \nu^*\big)^+\right] > \beta.
\]
This implies that for $\lambda^*\in[0,\lambda_{\max})$, we have
\begin{equation*}
\Gamma_\Lambda\left(\lambda^*-\eta\bigg(\beta-\Big(\nu^*+\frac{1}{1-\alpha}\expec\big[\big({\color{black}\mathcal J}^{\theta^*}(x^0)-\nu^*\big)^+\big]\Big)\bigg)\right)=\lambda^*-\eta\bigg(\beta-\Big(\nu^*+\frac{1}{1-\alpha}\expec\big[\big({\color{black}\mathcal J}^{\theta^*}(x^0)- \nu^*\big)^+\big]\Big)\bigg)
\end{equation*}
for any $\eta\in (0,\eta_{\max}]$, for some sufficiently small $\eta_{\max}>0$. Therefore, 
\[
\Upsilon_\lambda\left[\nabla_\lambda L(\nu,\theta,\lambda)\bigg\vert_{\theta=\theta^*(\lambda),\nu=\nu^*(\lambda),\lambda=\lambda^*}\right]\Bigg\vert_{\lambda=\lambda^*}
= \nu^*+\frac{1}{1-\alpha}\expec\left[\big({\color{black}\mathcal J}^{\theta^*}(x^0)- \nu^*\big)^+\right] -\beta>0.
\]
This {\color{black}is in contradiction with the fact that} $\Upsilon_\lambda\left[\nabla_\lambda L(\nu,\theta,\lambda)\vert_{\theta=\theta^*(\lambda),\nu=\nu^*(\lambda),\lambda=\lambda^*}\right]\vert_{\lambda=\lambda^*}=0$. Therefore,~\eqref{prop_1} holds.

To show that \eqref{prop_2} holds, we only need to show that $\lambda^*=0$ if 
\[
\nu^*+\frac{1}{1-\alpha}\expec\left[\big({\color{black}\mathcal J}^{\theta^*}(x^0)- \nu^*\big)^+\right] < \beta.
\]
Suppose $\lambda^*\in(0,\lambda_{\max})$, then there exists a sufficiently small $\eta_0>0$ such that
\[
\begin{split}
&\frac{1}{\eta_0}\left(\Gamma_\Lambda\bigg(\lambda^*-\eta_0\Big(\beta-\big(\nu^*+\frac{1}{1-\alpha}\expec\big[\big({\color{black}\mathcal J}^{\theta^*}(x^0)- \nu^*\big)^+\big]\big)\Big)\bigg)-\Gamma_\Lambda(\lambda^\ast)\right)\\
=& \,\nu^*+\frac{1}{1-\alpha}\expec\left[\big({\color{black}\mathcal J}^{\theta^*}(x^0)- \nu^*\big)^+\right] -\beta<0.
\end{split}
\]
This again contradicts the assumption $\Upsilon_\lambda\left[\nabla_\lambda L(\nu,\theta,\lambda)\vert_{\theta=\theta^*(\lambda),\nu=\nu^*(\lambda),\lambda=\lambda^*}\right]\vert_{\lambda=\lambda^*}=0$. Therefore \eqref{prop_2} holds. 

When $\lambda^\ast=\lambda_{\max}$ and $\nu^*+\frac{1}{1-\alpha}\expec\left[\big({\color{black}\mathcal J}^{\theta^*}(x^0)- \nu^*\big)^+\right] > \beta$,
\[
\Gamma_\Lambda\left(\lambda^*-\eta\bigg(\beta-\Big(\nu^*+\frac{1}{1-\alpha}\expec\big[\big({\color{black}\mathcal J}^{\theta^*}(x^0)-\nu^*\big)^+\big]\Big)\bigg)\right)=\lambda_{\max}
\]
for any $\eta>0$ and 
\[
\Upsilon_\lambda\left[\nabla_\lambda L(\nu,\theta,\lambda)\vert_{\theta=\theta^*(\lambda),\nu=\nu^*(\lambda),\lambda=\lambda^*}\right]\mid_{\lambda=\lambda^*}=0.
\]
In this case one cannot guarantee feasibility using the above
analysis, and $(\theta^\ast,\nu^\ast,\lambda^\ast)$ is not a local
saddle point. Such {\color{black}a} $\lambda^\ast$ is referred {\color{black}to} as a spurious fixed
point (see e.g., {\color{black}Chapter 8 of \citet{kushner1997stochastic}}). Notice that $\lambda^*$ is
bounded (otherwise we can conclude that the problem is infeasible), so
that by incrementally increasing $\lambda_{\max}$ in Algorithm \ref{alg_traj}, we can always prevent ourselves from obtaining a spurious fixed point solution.

{\color{black} Combining the above arguments, we finally conclude that
  $(\theta^*,\nu^*,\lambda^*)$ is a local saddle point of $L(\nu,\theta,\lambda)$. Then by the saddle point theorem, $\theta^*$ is
{\color{black}a locally optimal policy for the CVaR-constrained optimization problem}.}
\end{prooff}

\section{Convergence of Actor-Critic Algorithms}
\label{sec:appendix_AC}

{\color{black}Recall from Assumption~\ref{ass:steps_pg} that} the SPSA step size $\{\Delta_k\}$ satisfies $\Delta_k\rightarrow 0$ as $k\rightarrow\infty$ and $\sum_k (\zeta_2(k)/\Delta_k)^2<\infty$.
%%%%%%%%%%%%%%%%%%%%%%%%%%%%%%%%%%%%%%%%%%%%%%%%%%%%%%%%%%%%%%
%%%%%%%%%%%%%%%%%%%%%%%%%%%%%%%%%%%%%%%%%%%%%%%%%%%%%%%%%%%%%%
%%%%%%%%%%%%%%%%%%%%%%%%%%%%%%%%%%%%%%%%%%%%%%%%%%%%%%%%%%%%%%

\subsection{Gradient with Respect to $\lambda$ (Proof of Lemma \ref{lem:grad_lambda})}
\label{subsec:grad-lambda-comp}

 By taking the gradient of $V^\theta(x^0,\nu)$ w.r.t.~$\lambda$
 ({\color{black}recall} that both $V$ and $Q$ {\color{black}depend on}
 $\lambda$ through the cost function $\bar{C}$ of the augmented MDP $\bar{\mathcal{M}}$), we obtain
\begin{align}
\label{eq:grad-lambda0}
\nabla_\lambda V^\theta(x^0,\nu)&=\sum_{a\in\bar{\A}}\mu(a|x^0,\nu;\theta)\nabla_{\lambda}Q^\theta(x^0,\nu,a) \nonumber \\
&=\sum_{a\in\bar{\A}}\mu(a|x^0,\nu;\theta)\nabla_{\lambda}\Big[\bar{C}(x^0,\nu,a)+\sum_{(x',s')\in\bar{\X}}\gamma\bar{P}(x',s'|x^0,\nu,a)V^\theta(x',s')\Big] \nonumber \\
&=\underbrace{\sum_a\mu(a|x^0,\nu;\theta)\nabla_{\lambda}\bar{C}(x^0,\nu,a)}_{h(x^0,\nu)} + \gamma\sum_{a,x',s'}\mu(a|x^0,\nu;\theta)\bar{P}(x',s'|x^0,\nu,a)\nabla_{\lambda}V^\theta(x',s') \nonumber \\
&=h(x^0,\nu) + \gamma\sum_{a,x',s'}\mu(a|x^0,\nu;\theta)\bar{P}(x',s'|x^0,\nu,a)\nabla_{\lambda}V^\theta(x',s') \\
&=h(x^0,\nu) + \gamma\sum_{a,x',s'}\mu(a|x^0,\nu;\theta)\bar{P}(x',s'|x^0,\nu,a)\Big[h(x',s') \nonumber \\ 
&\hspace{0.6in}+\gamma\sum_{a',x'',s''}\mu(a'|x',s';\theta)\bar{P}(x'',s''|x',s',a')\nabla_\lambda V^\theta(x'',s'')\Big]. \nonumber
\end{align}
By unrolling the last equation using the definition of $\nabla_\lambda V^\theta(x,s)$ from~\eqref{eq:grad-lambda0}, we obtain
\begin{align*}
\label{eq:theta_gradient}
\nabla_\lambda V^\theta(x^0,\nu)&=\sum_{k=0}^\infty\gamma^k\sum_{x,s}\mathbb P(x_k=x,s_k=s\mid x_0=x^0,s_0=\nu;\theta)h(x,s) \\ 
&= \frac{1}{1-\gamma}\sum_{x,s}d_\gamma^\theta(x,s|x^0,\nu)h(x,s) = \frac{1}{1-\gamma}\sum_{x,s,a}d_\gamma^\theta(x,s|x^0,\nu)\mu(a|x,s)\nabla_\lambda\bar{C}(x,s,a) \\
&= \frac{1}{1-\gamma}\sum_{x,s,a}\pi_\gamma^\theta(x,s,a|x^0,\nu)\nabla_\lambda\bar{C}(x,s,a) \\
&= \frac{1}{1-\gamma}\sum_{x,s,a}\pi_\gamma^\theta(x,s,a|x^0,\nu)\frac{1}{1-\alpha}\mathbf 1 \{x={\color{black} x_{\text{Tar}}}\}(-s)^+. 
\end{align*}
This completes the proof.

%%%%%%%%%%%%%%%%%%%%%%%%%%%%%%%%%%%%%%%%%%%%%%%%%%%%%%%%%%%%%%%
%%%%%%%%%%%%%%%%%%%%%%%%%%%%%%%%%%%%%%%%%%%%%%%%%%%%%%%%%%%%%%%
%%%%%%%%%%%%%%%%%%%%%%%%%%%%%%%%%%%%%%%%%%%%%%%%%%%%%%%%%%%%%%%

\subsection{Proof of Convergence of the Actor-Critic Algorithms}
\label{subsec:convergence-proof-AC}
{\color{black}
Before going through the details of the convergence proof, a high level overview of the proof technique is given as follows.  
\begin{enumerate}
\item By utilizing temporal difference techniques, we show the critic
  update converges (in the fastest time scale) almost surely to a fixed point solution $v^*$ of the projected form of Bellman's equation, which is defined on the augmented MDP $\bar{\mathcal{M}}$. 
\item Similar to the analysis of the policy gradient algorithm, by
  convergence properties of multi-time scale discrete stochastic
  approximation algorithms, we show that each update
  $(\nu_k,\theta_k,\lambda_k)$ converges almost surely to a stationary point $(\nu^\ast,\theta^*,\lambda^*)$ of the corresponding continuous time system. In particular, by adopting the step-size rules defined in Assumption \ref{ass:steps_ac}, we show that the convergence rate of $v$ is fastest, followed by the convergence rate of $\nu$ and the convergence rate of $\theta$, while the convergence rate of $\lambda$ is the slowest among the set of parameters. Different from the policy gradient algorithm, the parameters of the actor-critic algorithm are updated incrementally. To adjust for this difference in the convergence analysis, modifications to the gradient estimate of $\nu$ are introduced, either via the SPSA method or the semi-trajectory method, to ensure the gradient estimates are unbiased. Following from the arguments of Lyapunov analysis, we prove that the continuous time system is locally asymptotically stable at the stationary point $(\nu^\ast,\theta^*,\lambda^*)$. 
\item Following the same line of arguments from the proof of the policy gradient algorithm, we conclude that the stationary point $(v^*, \nu^\ast,\theta^*,\lambda^*)$ is a local saddle point. Finally, by the the local saddle point theorem, we deduce that $\theta^*$ is a locally optimal solution for the CVaR-constrained MDP problem.
\end{enumerate}
This convergence proof procedure is rather standard for stochastic approximation algorithms, see~\citep{bhatnagar2009natural,bhatnagar2012online} for further references.}
\subsubsection{Proof of Theorem \ref{thm:converge_kncre_v}: Critic Update ($v$-update)} \label{subsec:v_update}
By the step {\color{black}size} conditions, one notices that $\{v_k\}$ converges {\color{black}on} a faster time scale than $\{\nu_{k}\}$, $\{\theta_k\}$,  and $\{\lambda_k\}$. {\color{black}According to Lemma 1 in Chapter 6 of \citet{borkar2008stochastic}, one can {\color{black}take} $(\nu,\theta,\lambda)$ in the $v$-update as arbitrarily fixed quantities (in this case we have $(\nu,\theta,\lambda)=(\nu_k,\theta_k,\lambda_k)$)}. Thus the critic update can be re-written as follows:
\begin{equation}\label{eq:TD_0}
v_{k+1}=v_k+\zeta_4(k)\phi(x_k,s_k)\delta_k(v_k),
\end{equation}
where the scalar 
\[
\delta_k\left(v_k\right)=-v_k^\top\phi(x_k,s_k)+ \gamma v_k^\top\phi\left(x_{k+1},s_{k+1}\right)+ \bar{C}_{\lambda}(x_k,s_k,a_k)
\]
is the temporal difference (TD) {\color{black}from~(\ref{TD-calc})}.  Define
\begin{equation}\label{eq:A}
\begin{split}
A:=&\sum_{y,a^\prime,s^\prime}\pi_{\gamma}^\theta(y,s^\prime,a^\prime|x,s)\phi(y,s^\prime)\left(\phi^\top(y,s^\prime)- \gamma\sum_{z,s^{\prime\prime}}\bar{P}(z,s^{\prime\prime}|y,s^\prime,a) \phi^\top\left(z,s^{\prime\prime}\right)\right),
\end{split}
\end{equation}
and 
\begin{equation}\label{eq:b}
b:=\sum_{y,a^\prime,s^\prime}\pi_{\gamma}^\theta(y,s^\prime,a^\prime|x,s)\phi(y,s^\prime)\bar{C}_{\lambda}(y,s^\prime,a^\prime).
\end{equation}
It is easy to see that the critic update $v_k$ in \eqref{eq:TD_0} can be re-written as the following stochastic approximation scheme:
\begin{equation}\label{eq:TD_stoch_approx}
v_{k+1}=v_k+\zeta_4(k)(b-Av_k+\delta A_{k+1}),
\end{equation}
where the noise term $\delta A_{k+1}$ is a square integrable
Martingale difference, i.e., $\mathbb E[\delta A_{k+1}\mid \mathcal
F_k]=0$ if the $\gamma$-{\color{black} occupation measure}  $\pi_{\gamma}^\theta$
{\color{black}is} used to generate samples of $(x_k,s_k,a_k) ${\color{black} with} $\mathcal F_k$ {\color{black}being} the filtration generated by different independent trajectories. By writing 
\[
\delta A_{k+1}=-(b-Av_k)+\phi(x_k,s_k)\delta_k(v_k)
\]
and noting $\mathbb E_{\pi_\gamma^\theta}[\phi(x_k,s_k)\delta_k(v_k)\mid \mathcal F_k]=-Av_k+b$,
one can easily {verify} that the stochastic approximation scheme in {\color{black}\eqref{eq:TD_stoch_approx}} is equivalent to the {\color{black}critic} iterates in \eqref{eq:TD_0} and $\delta A_{k+1}$ is a Martingale difference, i.e., $\mathbb E_{\pi_\gamma^\theta}[\delta A_{k+1}\mid \mathcal F_k]=0$.
 Let 
\[
h\left(v\right):=-Av+b.
\]
Before getting into the convergence analysis, we {\color{black}present a} technical lemma whose proof can be found in Lemma 6.10 of \citet{BertsekasT96}.
\begin{lemma}\label{lem:tech_A_CVaR}
Every eigenvalue of {\color{black}the} matrix $A$ has positive real part.
\end{lemma}
We now turn to the analysis of the {\color{black}critic} iteration. Note that the following properties hold for the {\color{black}critic} update scheme in (\ref{eq:TD_0}): 1) $h\left(v\right)$ is Lipschitz, 2) the step size satisfies the properties in Assumption~\ref{ass:steps_ac}, 3) the noise term $\delta A_{k+1}$ is a square integrable Martingale difference, 4) the function $h_c\left(v\right):=h\left(cv\right)/c$, $c\geq 1$ converges uniformly to a continuous function $h_\infty\left(v\right)$ for any $v$ in a compact set, i.e., $h_c\left(v\right)\rightarrow h_\infty\left(v\right)$ as $c\rightarrow\infty$, and 5) the ordinary differential equation (ODE) $\dot v= h_\infty\left(v\right)$
has the origin as its unique {\color{black}locally} asymptotically stable equilibrium. The fourth property can be easily verified from the fact that the magnitude of $b$ is finite and $h_\infty\left(v\right)=-Av$. The fifth property follows directly from the facts that $h_\infty\left(v\right)=-Av$ and all eigenvalues of $A$ have positive real parts. 

By Theorem 3.1 in \citet{borkar2008stochastic}, these five properties imply:
\[
\text{The {\color{black}critic} iterates $\{v_k\}$ are bounded almost surely, i.e.,}\,\,\sup_k\left\|v_k\right\|<\infty\,\,\,\text{almost surely}.
\]
The convergence of the {\color{black}critic} iterates in \eqref{eq:TD_0} can be related to the asymptotic behavior of the ODE
\begin{equation}\label{exp:bdd}
\dot v= h\left(v\right)=b-Av.
\end{equation}
{\color{black}Specifically,} Theorem 2 in Chapter 2 of
\citet{borkar2008stochastic} {\color{black}and the above conditions imply} $v_k\rightarrow v^*$ with probability $1$, where the limit $v^*$ depends on $(\nu,\theta,\lambda)$ and is the unique solution satisfying $h\left(v^*\right)=0$, i.e., $Av^*=b$. Therefore, the {\color{black}critic} iterates converge to the unique fixed point $v^*$ almost surely, as $k\rightarrow\infty$.

\subsubsection{Proof of Theorem \ref{thm:converge_kncre}}\label{subsec:actor_update}
%\noindent\paragraph{Step 1 (Convergence of $(v,w)$-update)}
%The proof of the critic parameter convergence follows directly from Theorem~\ref{thm:converge_kncre_v} and Theorem~\ref{thm:converge_kncre_w}.
\noindent\paragraph{Step 1 (Convergence of $v$-update)}
The proof of {\color{black}convergence for} the critic parameter  follows directly from Theorem~\ref{thm:converge_kncre_v}.

\noindent\paragraph{Step 2 (Convergence of SPSA Based $\nu$-update)}
In this section, we {\color{black}analyze} the $\nu$-update for the incremental actor-critic method. This update is based on the SPSA perturbation method. The idea of this method is to estimate the sub-gradient $g(\nu)\in\partial_{\nu}  L(\nu,\theta,\lambda)$ using two simulated value functions corresponding to $\nu^{-}=\nu-\Delta$ and $\nu^{+}=\nu+\Delta$. Here $\Delta\geq 0$ is a positive random perturbation that vanishes asymptotically. The SPSA-based estimate for a sub-gradient $g(\nu)\in\partial_{\nu}  L(\nu,\theta,\lambda)$ is given by
\[
g(\nu)\approx \lambda+\frac{1}{2\Delta}\left( \phi^\top\left(x^0,\nu+\Delta\right)- \phi^\top\left(x^0,\nu-\Delta\right)\right)v.
\]

First, we {\color{black}consider} the following assumption on the feature functions in order to prove {\color{black}that} the SPSA approximation is asymptotically unbiased.
\begin{assumption}\label{assume_lip}
For any $v\in\reals^{\kappa_1}$, the feature {\color{black}functions satisfy} the following conditions
\[
|\phi^\top\left(x^0,\nu+\Delta\right)v-\phi^\top\left(x^0,\nu-\Delta\right)v|\leq K_1(v)(1+\Delta).
\] 
Furthermore, the Lipschitz constants are uniformly bounded, i.e., $\sup_{v\in\reals^{\kappa_1}}K^2_1(v)<\infty$. 
\end{assumption}
This assumption is mild {\color{black}as} the expected utility objective function implies that $L(\nu,\theta,\lambda)$ is Lipschitz in $\nu$, and $\phi^\top_V\left(x^0,\nu\right)v$ is just a linear function approximation of $V^\theta(x^0,\nu)$.

{\color{black}
Next, we turn to the convergence analysis of {\color{black}the} sub-gradient estimation and $\nu$-update. Since $\nu$ converges faster than $\theta$ and $\lambda$. Consider the $\nu$-update in~\eqref{nu_up_kncre_SPSA}:
\begin{equation}
\nu_{k+1}=\Gamma_{\mathcal{N}}\left(\nu_{k}-\zeta_3(k)\left(
\lambda+\frac{1}{2\Delta_k} \left( \phi^\top\left(x^0,\nu_{k}+\Delta_k\right)- \phi^\top\left(x^0,\nu_{k}-\Delta_k\right)\right)v_k\right)\right),
\end{equation}
{\color{black}where {\color{black} according to Lemma 1 in Chapter 6 of \citet{borkar2008stochastic}}, $(\theta_k,\lambda_k$) in this expression are viewed as constant quantities.}
Since $v$ converges faster than $\nu$, Lemma 1 in Chapter 6 of \citet{borkar2008stochastic} also implies $\|v_k-v^*(\nu_k)\|\rightarrow 0$ almost surely, where $v^*(\nu)$ is the converged critic parameter. Together with the above assumption that the feature function is bounded, 
one can rewrite the $\nu$-update in~\eqref{nu_up_kncre_SPSA} as follows:
\begin{equation}\label{eq:s_grad_est}
\nu_{k+1}=\Gamma_{\mathcal{N}}\left(\nu_{k}-\zeta_3(k)\left(
\lambda+\frac{1}{2\Delta_k} \left( \phi^\top\left(x^0,\nu_{k}+\Delta_k\right)- \phi^\top\left(x^0,\nu_{k}-\Delta_k\right)\right)v^*(\nu_k) + \epsilon_k\right)\right),
\end{equation}
where 
\[
\epsilon_k = \frac{1}{2\Delta_k} \left( \phi^\top\left(x^0,\nu_{k}+\Delta_k\right)- \phi^\top\left(x^0,\nu_{k}-\Delta_k\right)\right)(v_k-v^*(\nu_k))\rightarrow 0,\quad\text{almost surely}.
\]
}
{\color{black}Equipped with this intermediate result}, we establish the bias and convergence of {\color{black}the} stochastic sub-gradient estimate. Let
\[
\overline g(\nu_{k})\in\arg\max\left\{g:g\in\partial_{\nu}  L(\nu,\theta,\lambda)\vert_{\nu=\nu_{k}}\right\},
\]
and
 \[
 \begin{split}
  \Lambda_{1,k+1}=&\left(
 \frac{\left(\phi^\top\left(x^0,\nu_{k}+\Delta_k\right)- \phi^\top\left(x^0,\nu_{k}-\Delta_k\right)\right)v^*(\nu_{k})}{2\Delta_k}-E_M(k)\right),\\
 \Lambda_{2,k}=&\lambda_k+E^L_M(k)-\overline g(\nu_{k}),\\
 \Lambda_{3,k}=&E_M(k)-E^L_M(k),\\
 \end{split}
 \]
 where
\[
\begin{split}
E_M(k):=&\mathbb E\left[\frac{1}{2\Delta_k}\left(\phi^\top\left(x^0,\nu_{k}+\Delta_k\right)- \phi^\top\left(x^0,\nu_{k}-\Delta_k\right)\right)v^*(\nu_{k})\mid \Delta_k\right],\\
E^L_M(k):=&\mathbb E\left[\frac{1}{2\Delta_k}\left(V^{\theta}\left(x^0,\nu_{k}+\Delta_k\right)- V^{\theta}\left(x^0,\nu_{k}-\Delta_k\right)\right)\mid \Delta_k\right].
\end{split}
\]
Note that \eqref{eq:s_grad_est} is equivalent to
 \begin{equation}\label{eq:s_noisy_grad}
 \nu_{k+1}=\Gamma_{\mathcal{N}}\left(\nu_{k}-\zeta_3(k)\left(
 \overline g(\nu_{k})+ \Lambda_{1,k+1}+\Lambda_{2,k}+\Lambda_{3,k}\right)\right).
 \end{equation}
First, it is clear that $ \Lambda_{1,k+1}$ is a Martingale difference as $\mathbb E[ \Lambda_{1,k+1}\mid \mathcal F_k]=0$, which implies that
\[
M_{k+1}=\sum_{j=0}^k \zeta_{3}(j) \Lambda_{1,j+1}
\]
is a Martingale w.r.t.~the filtration $\mathcal F_k$. By {\color{black}the} Martingale convergence theorem, we can show that if $\sup_{k\geq 0} \mathbb E[M^2_k]<\infty$, when $k\rightarrow\infty$, $M_k$ converges almost surely and $\zeta_3(k) \Lambda_{1,k+1}\rightarrow 0$ almost surely. To show that $\sup_{k\geq 0} \mathbb E[M^2_k]<\infty$, for any $t\geq 0$ one observes that
\[
\begin{split}
\mathbb E[M^2_{k+1}]&= \sum_{j=0}^k\left(\zeta_{3}(j)\right)^2\mathbb E[\mathbb E[ \Lambda_{1,j+1}^2\mid \Delta_j]]\\
&\leq  2 \sum_{j=0}^k\mathbb E\bigg[\left(\frac{\zeta_{3}(j)}{2\Delta_j}\right)^2\left\{\mathbb E\left[\left(\Big(
 {\phi^\top\left(x^0,\nu_j+\Delta_j\right)- \phi^\top\left(x^0,\nu_j-\Delta_j\right)\Big)v^*(\nu_j)}\right)^2\mid  \Delta_j\right]\right.\\
&\;\quad\quad\quad\quad\quad\quad\quad\quad\quad\left.+\mathbb E\left[\big(\phi^\top\left(x^0,\nu_j+\Delta_j\right)- \phi^\top\left(x^0,\nu_j-\Delta_j\right)\big)v^*(\nu_j)\mid \Delta_j\right]^2\right\}\bigg].
 \end{split}
\]
Now based on Assumption \ref{assume_lip}, the above expression implies
\[\begin{split}
\mathbb E[M^2_{k+1}]\leq & 2 \sum_{j=0}^k\mathbb E\left[\left(\frac{\zeta_{3}(j)}{2\Delta_j}\right)^22K_1^2(1+\Delta_j)^2\right].
\end{split}\]
Combining the above results with the step {\color{black}size} conditions, there exists $K=4K_1^2>0$ such that
\[
\sup_{k\geq 0}\mathbb E[M^2_{k+1}] \leq  K \sum_{j=0}^\infty\mathbb E\left[ \left(\frac{\zeta_{3}(j)}{2\Delta_j}\right)^2 \right]+\left(\zeta_{3}(j)\right)^2<\infty.
\]

Second, by the Min Common/Max Crossing theorem in {\color{black}Chapter 5 of \citet{bertsekas2009min}}, one can show that $\partial_{\nu} L(\nu,\theta,\lambda)\vert_{\nu=\nu_{k}}$ is a non-empty, convex, and compact set. Therefore, by duality of directional directives and sub-differentials, i.e.,
\[
\max \left\{g:g\in\partial_{\nu}  L(\nu,\theta,\lambda)\vert_{\nu=\nu_{k}}\right\} = \lim_{\xi\downarrow 0}\frac{ L(\nu_k+\xi,\theta,\lambda)-L(\nu_k-\xi,\theta,\lambda)}{2\xi},
\]
one concludes that for $\lambda_k=\lambda$ ({\color{black}we can
  treat $\lambda_k$ as a constant because it} converges {\color{black}on} a slower
time scale {\color{black}than $\nu_k$}),
\[
\lambda+E^L_M(k)= \overline g(\nu_{k})+O(\Delta_k),
\] 
almost surely. This further implies that
\[
\Lambda_{2,k}= O(\Delta_k),\,\,\quad \text{i.e.,} \quad \Lambda_{2,k}\rightarrow 0 \;\;\text{ as }\;\; k\rightarrow \infty, 
\]
almost surely. 

Third, since $d^\theta_\gamma(x^0,\nu|x^0,\nu)=1$, from
{\color{black}the} definition of $\epsilon_{\theta}(v^*(\nu_{k}))$, 
\[
|\Lambda_{3,k}|\leq 2 \epsilon_{\theta}(v^*(\nu_{k}))/\Delta_k.
\]
{\color{black}As} $t$ goes to infinity, $\epsilon_{\theta}(v^*(\nu_{k}))/\Delta_k\rightarrow 0$ by assumption and $\Lambda_{3,k}\rightarrow 0$. 

Finally, {\color{black}since}  $\zeta_3(k)
\Lambda_{1,k+1}\rightarrow 0$, $\Lambda_{2,k}\rightarrow 0$, and
$\Lambda_{3,k}\rightarrow 0$ almost surely, the $\nu$-update in
\eqref{eq:s_noisy_grad} is a noisy sub-gradient descent update with
vanishing disturbance bias. {\color{black}Thus, the $\nu$-update
  in~\eqref{nu_up_kncre_SPSA}} can be viewed as an Euler
discretization of {\color{black}an element of} the following differential inclusion,
\begin{equation}\label{dyn_sys_s_kncre_SPSA}
\dot{\nu}\in \Upsilon_{\nu}\left[-g(\nu)\right], \quad\quad \forall g(\nu)\in\partial_\nu L(\nu,\theta,\lambda),
\end{equation}  
{\color{black}and} the $\nu$-convergence analysis {\color{black}is}
analogous {\color{black}to} Step~1 of the proof of Theorem~\ref{thm:converge_h}.

\noindent\paragraph{Step $2^\prime$ (Convergence of Semi-trajectory $\nu$-update)}
Since $\nu$ converges on a faster timescale than $\theta$ and $\lambda$, {\color{black} according to Lemma 1 in Chapter 6 of \citet{borkar2008stochastic}, the convergence property of $\nu$ in~\eqref{nu_up_kncre_semi_traj} can be shown for any arbitrarily fixed pair of $(\theta,\lambda)$ (in this case we have $(\theta,\lambda)=(\theta_k,\lambda_k)$)}, i.e.,
\begin{equation}\label{update_s_multi}
\nu_{k+1}=\Gamma_{\mathcal{N}}\left(\nu_k-\zeta_3(k)\left(\lambda-\frac{\lambda}{1-\alpha}\left(\mathbb P\left({\color{black}s_{\text{Tar}}}\leq 0\mid x_0=x^0,s_0=\nu_k,\theta\right)+\delta \nu_{M,k+1}\right)\right)\right),
\end{equation}
{\color{black}where}
\begin{equation}\label{eq:MG_diff_s_multi}
\delta \nu_{M,k+1}=-\mathbb P\left({\color{black}s_{\text{Tar}}}\leq 0\mid x_0=x^0,s_0=\nu_k,\mu\right)+\mathbf 1\left\{x_k={\color{black} x_{\text{Tar}}},s_k\leq 0\right\}
\end{equation}
is a square integrable stochastic term, {\color{black}specifically},
\[
\mathbb E[(\delta \nu_{M,k+1})^2\mid \mathcal F_{\nu,k}]\leq 2,
\] 
where $\mathcal F_{\nu,k}= \sigma(\nu_m,\,\delta\nu_m,\,m\leq k)$ is
the filtration {\color{black}generated by}
$\nu$. {\color{black}Since $\mathbb E\left[\delta \nu_{M,k+1}\mid
    \mathcal F_{\nu,k}\right]=0$, $\delta\nu_{M,k+1}$ is a Martingale
  difference and} the $\nu$-update in \eqref{update_s_multi} is a stochastic approximation of an element {\color{black}of} the differential inclusion
\[
\frac{\lambda}{1-\alpha}\mathbb P\left({\color{black}s_{\text{Tar}}}\leq 0\mid x_0=x^0,s_0=\nu_k,\theta\right)-\lambda\in-\partial_\nu  L(\nu,\theta,\lambda)\vert_{\nu=\nu_k}.
\]
Thus, the $\nu$-update in~\eqref{nu_up_kncre_semi_traj} can be viewed as an
Euler discretization of the differential inclusion
in~\eqref{dyn_sys_s_kncre_SPSA}, and the $\nu$-convergence analysis
{\color{black}is} analogous {\color{black}to} Step~1~{\color{black}of the proof of} Theorem~\ref{thm:converge_h}.

\noindent\paragraph{Step 3 (Convergence of $\theta$-update)}
We first analyze the actor update ($\theta$-update).
Since $\theta$ converges {\color{black}on} a faster time scale than $\lambda$, {\color{black} according to Lemma 1 in Chapter 6 of \citet{borkar2008stochastic}}, one can {\color{black}take} $\lambda$ in the $\theta$-update as a fixed quantity {\color{black}(i.e., here we have that $\lambda=\lambda_k$)}. Furthermore, since $v$ and $\nu$ converge {\color{black}on} a faster scale than $\theta$, {\color{black} one also have
$\|v_k-v^*(\theta_k)\|\rightarrow 0$, $\|\nu_k-\nu^{*}(\theta_k)\|\rightarrow 0$ almost surely, and since convergence almost surely of the $\nu$ sequence implies convergence in distribution, we have $\|\pi_{\gamma}^{\theta_k}(x^\prime,s^\prime,a^\prime|x_0=x^0,s_0=\nu_k)-\pi_{\gamma}^{\theta_k}(x^\prime,s^\prime,a^\prime|x_0=x^0,s_0=\nu^*(\theta_k))\|\rightarrow 0$.
In the following analysis, we assume that the initial state $x^0\in\X$ is given. Consider the $\theta$-update in~\eqref{theta_up_kncre} 
\begin{equation}\label{theta_up}
\theta_{k+1}=\Gamma_\Theta\left(\theta_k-\zeta_2(k) \left(\nabla_\theta \log\mu(a_k|x_k,s_k;\theta)\vert_{\theta=\theta_k}\frac{\delta_k(v_k)}{1-\gamma}\right)\right).
\end{equation}
Utilizing the above convergence properties, \eqref{theta_up} can be rewritten as follows:
\[
\theta_{k+1}=\Gamma_\Theta\left(\theta_k-\zeta_2(k) \left(\nabla_\theta \log\mu(a_k|x_k,s_k;\theta)\vert_{\theta=\theta_k}\left(\frac{\delta_k(v^*(\theta_k))}{1-\gamma}+\epsilon_k\right)\right)\right),
\]
where we showed in the convergence analysis of the $\nu$ sequence that 
\[
\epsilon_k=\frac{\delta_k(v_k)}{1-\gamma}-\frac{\delta_k(v^*(\theta_k))}{1-\gamma}\rightarrow 0,\quad \text{almost surely}.
\]
}
Consider the case in which the value function for a fixed policy $\theta$ (i.e., $\theta=\theta_k$) is approximated by a learned function approximation, $\phi^\top(x,s)v^*(\theta_k)$. If the approximation is sufficiently good, we might hope to
use it in place of $V^\theta(x,s)$ and still point roughly in the direction of the
true gradient. Recall the temporal difference error (random variable)
for {\color{black}a} given {\color{black}pair} $(x_k,s_k)\in\X\times\reals$:
\[
\delta_k\left(v\right)=-v^\top\phi(x_k,s_k)+ \gamma v^\top\phi\left(x_{k+1},s_{k+1}\right)+ \bar{C}_{\lambda}(x_k,s_k,a_k).
\] 
Define the $v$-dependent approximated advantage function 
\[
\tilde A^{\theta,v}(x,s,a):=\tilde \Q^{\theta,v}(x,s,a)-v^\top\phi(x,s),
\]
where
\[
\tilde \Q^{\theta,v}(x,s,a)=\gamma\sum_{x^\prime,s^\prime} \bar{P}(x^\prime,s^\prime|x,s,a) v^\top\phi(x^\prime,s^\prime)+\bar{C}_{\lambda}(x,s,a).
\]
The following lemma, whose proof follows from {\color{black}the proof of} Lemma 3 in \citet{bhatnagar2009natural}, shows that $\delta_k(v)$ is an unbiased estimator of $\tilde A^{\theta,v}$.
\begin{lemma}\label{lem:unbiased_L}
For any given policy $\mu$ and $v\in\reals^{\kappa_1}$, we have 
\[
\tilde A^{\theta,v}(x,s,a)=\mathbb E[\delta_k(v)\mid x_k=x,s_k=s,a_k=a].
\]
\end{lemma}
%\begin{prooff}
%Note that for any $v\in\reals^{\kappa_1}$,
%\[
%\begin{split}
%\mathbb E[\delta_k(v)\mid x_k&=x,s_k=s,a_k=a,\mu] = \\
%&\bar{C}_{\lambda}(x,s,a)-v^\top\phi(x,s)+\gamma\mathbb E\left[v^\top\phi(x_{k+1},s_{k+1})\mid x_k=x,s_k=s,a_k=a\right],
%\end{split}
%\]
%where
%\[
%\mathbb E\left[v^\top\phi(x_{k+1},s_{k+1})\mid x_k=x,s_k=s,a_k=a\right]=\sum_{x^\prime,s^\prime} \bar{P}(x^\prime,s^\prime|x,s,a) v^\top\phi(x^\prime,s^\prime).
%\]
%By recalling the definition of $\tilde \Q^{\theta,v}(x,s,a)$, the proof is completed.
%\end{prooff}
Define 
\[
\begin{split}
\nabla_\theta \tilde L_{v}(\nu,\theta,\lambda):=\frac{1}{1-\gamma}\sum_{x,a,s}&\pi^\theta_\gamma(x,s,a|x_0=x^0,s_0=\nu)\nabla_{\theta}\log\mu(a|x,s;\theta)\tilde A^{\theta,v}(x,s,a)
\end{split}
\]
as the linear function approximation of $\nabla_\theta \tilde L(\nu,\theta,\lambda)$.
Similar to Proposition \ref{L_lips}, we {\color{black}present} the following technical
lemma on the Lipschitz {\color{black}property} of $\nabla_\theta \tilde L_v(\nu,\theta,\lambda)$.

\begin{proposition}\label{L_lips_approx}
$\nabla_\theta \tilde L_v(\nu,\theta,\lambda)$ is a Lipschitz function in $\theta$.
\end{proposition}
\begin{prooff}
Consider the feature vector $v$. Recall that the feature vector
satisfies the linear equation $Av=b$, where $A$ and $b$
{\color{black}are given by~(\ref{eq:A}) and (\ref{eq:b}), respectively}. {\color{black}From} Lemma~1~{\color{black}in}~\citet{bhatnagar2012online}, by exploiting the inverse of $A$ using Cramer's rule, one may show that $v$ is continuously differentiable in $\theta$. Now consider the $\gamma$-{\color{black} occupation measure}  $\pi^\theta_\gamma$. {\color{black}By applying} Theorem~2~{\color{black}in}~\citet{altman2004perturbation} (or Theorem~3.1~{\color{black}in}~\citet{shardlow2000perturbation}), it can be seen that the {\color{black} occupation measure}  $\pi^\theta_\gamma$ of the process $(x_k,s_k)$ is continuously differentiable in $\theta$. Recall from Assumption~\ref{ass:differentiability} in Section~\ref{sec:Risk-Opt} that $\nabla_\theta\mu(a_k|x_k,s_k;\theta)$ is a Lipschitz function in $\theta$ for any $a\in\A$ and $k\in\{0,\ldots,T-1\}$, and $\mu(a_k|x_k,s_k;\theta)$ is differentiable in $\theta$. By combining these arguments and noting that the sum of products of Lipschitz functions is Lipschitz, one concludes that $\nabla_\theta \tilde L_v(\nu,\theta,\lambda)$ is Lipschitz in $\theta$.
\end{prooff}

We turn to the convergence proof of $\theta$. 
\begin{theorem}\label{thm:theta_AC}
The sequence of $\theta$-updates in~\eqref{theta_up_kncre} converges
almost surely to an equilibrium point $\widehat{\theta}^*$ that
satisfies $\Upsilon_\theta\left[ -\nabla_\theta  \tilde
  L_{v^*(\theta)}(\nu^*(\theta),\theta,\lambda)\right]=0$, for
{\color{black}a} given $\lambda\in[0,\lambda_{\max}]$. Furthermore, if the function approximation error $\epsilon_\theta(v_k)$ vanishes as the feature vector $v_k$ converges to $v^*$, then {\color{black}the} sequence of $\theta$-updates converges to $\theta^*$ almost surely, where $\theta^*$ is a local minimum point of $L(\nu^*(\theta),\theta,\lambda)$ for {\color{black}a} {\color{black}given} $\lambda\in[0,\lambda_{\max}]$.
\end{theorem}
\begin{prooff}
We will mainly focus on {\color{black}proving} the convergence of $\theta_k\rightarrow \theta^*$ (second part of the theorem). Since we just showed in Proposition \ref{L_lips_approx} that $\nabla_\theta\tilde L_{v^*(\theta)}(\nu^*(\theta),\theta,\lambda)$ is Lipschitz in $\theta$, the convergence proof of $\theta_k\rightarrow\widehat{\theta}^*$ (first part of the theorem) follows from identical arguments. 

Note that the $\theta$-update {\color{black}in}~\eqref{theta_up} can be rewritten as:
\[
\theta_{k+1}=\Gamma_\Theta\left(\theta_k+\zeta_2(k)\left(-\nabla_\theta L(\nu,\theta,\lambda)\vert_{\nu=\nu^*(\theta),\theta=\theta_k}+\delta\theta_{k+1}+\delta\theta_\epsilon\right)\right),
\]
where
\begin{equation*}
\begin{split}
\delta\theta_{k+1}&=\sum_{x^\prime,a^\prime,s^\prime} \pi_{\gamma}^{\theta_k}(x^\prime,s^\prime,a^\prime|x_0=x^0,s_0=\nu^*(\theta_k))\nabla_\theta\log\mu(a^\prime|x^\prime,s^\prime;\theta)\vert_{\theta=\theta_k}\frac{\tilde A^{\theta_k,v^*(\theta_k)}(x^\prime,s^\prime,a^\prime)}{1-\gamma}\\
&-\nabla_\theta\log\mu(a_k|x_k,s_k;\theta)\vert_{\theta=\theta_k}\frac{\delta_k(v^*(\theta_k))}{1-\gamma}.
\end{split}
\end{equation*}
{\color{black}and
\begin{equation*}
\begin{split}
\delta\theta_\epsilon=&\sum_{x^\prime,a^\prime,s^\prime}
\pi_{\gamma}^{\theta_k}(x^\prime,s^\prime,a^\prime|x_0=x^0,s_0=\nu^*(\theta_k))\cdot\\
&\quad\frac{\nabla_{\theta}\log\mu(a^\prime|x^\prime,s^\prime;\theta)\vert_{\theta=\theta_k}}{1-\gamma}(A^{\theta_k}(x^\prime,s^\prime,a^\prime)-\tilde
A^{\theta_k,v^*(\theta_k)}(x^\prime,s^\prime,a^\prime))
\end{split}
\end{equation*}
}
First, one {\color{black}can} show that $\delta\theta_{k+1}$ is square integrable, {\color{black}specifically,
\[
\begin{split}
& \mathbb E[\|\delta\theta_{k+1}\|^2\mid \mathcal F_{\theta,k}]\\
& \leq \frac{2}{1-\gamma}\|\nabla_\theta\log\mu(u|x,s;\theta)\vert_{\theta=\theta_k}\,\mathbf 1\left\{\mu(u|x,s;\theta_k)>0\right\}\|^2_\infty\left(\|\tilde A^{\theta_k,v^*(\theta_k)}(x,s,a)\|_\infty^2+|\delta_{k}(v^*(\theta_k))|^2\right)\\
& \leq \frac{2}{1-\gamma}\cdot\frac{\|\nabla_\theta\mu(u|x,s;\theta)\vert_{\theta=\theta_k}\|^2_\infty}{\min\{\mu(u|x,s;\theta_k)\mid
  \mu(u|x,s;\theta_k)>0\}^2}\left(\|\tilde A^{\theta_k,v^*(\theta_k)}(x,s,a)\|_\infty^2+|\delta_{k}(v^*(\theta_k))|^2\right)\\
& \leq 64\frac{K^2\|\theta_k\|^2}{1-\gamma} \left(\max_{x,s,a}|\bar{C}_{\lambda}(x,s,a)|^2+2\max_{x,s}\|\phi(x,s)\|^2 \sup_k\|v_k\|^2\right)\\
& \leq 64\frac{K^2\|\theta_k\|^2}{1-\gamma} \left(\left|\max\left\{C_{\max},\frac{2 \lambda D_{\max}}{\gamma^T(1-\alpha)(1-\gamma)}\right\}\right|^2+2\max_{x,s}\|\phi(x,s)\|^2 \sup_k\|v_k\|^2\right),
 \end{split}
\]
for some Lipschitz constant $K$, where the indicator function in the second line can be
  explained by the fact that $\pi_{\gamma}^{\theta_k}(x,s,u)=0$ whenever
  $\mu(u\mid x,s;\theta_k)=0$ and because the expectation is
  taken with respect to $\pi_{\gamma}^{\theta_k}$. The third
  inequality uses
  Assumption~\ref{ass:differentiability} and the fact that $\mu$ takes
  on finitely-many values (and thus its nonzero values are bounded
  away from zero). Finally,} $\sup_k\|v_k\|<\infty$ {\color{black}follows from} the Lyapunov analysis in the critic update.

Second, note that
\begin{equation}\label{diff_A_A_tilde}
\begin{split}
\delta\theta_\epsilon
\leq &\frac{(1+\gamma)\|\psi_{\theta_k}\|_\infty}{(1-\gamma)^2}\epsilon_{\theta_k}(v^*(\theta_k)),
\end{split}
\end{equation}
where
$\psi_\theta(x,s,a)=\nabla_{\theta}\log\mu(a|x,s;\theta)$ is the ``compatible feature." The last inequality is due to the fact that {\color{black}since}  $\pi^\theta_\gamma$ {\color{black}is} a probability measure, convexity of quadratic functions implies
\[
\begin{split}
&\sum_{x^\prime,a^\prime,s^\prime} \pi_{\gamma}^\theta(x^\prime,s^\prime,a^\prime|x_0=x^0,s_0=\nu^*(\theta))(A^\theta(x^\prime,s^\prime,a^\prime)-\tilde A^{\theta,v}(x^\prime,s^\prime,a^\prime))\\
\leq &\sum_{x^\prime,a^\prime,s^\prime} \pi_{\gamma}^\theta(x^\prime,s^\prime,a^\prime|x_0=x^0,s_0=\nu^*(\theta))( Q^\theta(x^\prime,s^\prime,a^\prime)-\tilde \Q^{\theta,v}(x^\prime,s^\prime,a^\prime))\\
&+\sum_{x^\prime,s^\prime} d_{\gamma}^\theta(x^\prime,s^\prime|x_0=x^0,s_0=\nu^*(\theta))(V^\theta(x^\prime,s^\prime)-\widetilde{V}^{\theta,v}(x^\prime,s^\prime))\\
= &\gamma\sum_{x^\prime,a^\prime,s^\prime} \pi_{\gamma}^\theta(x^\prime,s^\prime,a^\prime|x_0=x^0,s_0=\nu^*(\theta))\sum_{x^{\prime\prime},s^{\prime\prime}}\bar{P}(x^{\prime\prime},s^{\prime\prime}|x^\prime,s^\prime,a^\prime) (V^\theta(x^{\prime\prime},s^{\prime\prime})-\phi^\top(x^{\prime\prime},s^{\prime\prime})v)\\
&+\sqrt{\sum_{x^\prime,s^\prime} d_{\gamma}^\theta(x^\prime,s^\prime|x_0=x^0,s_0=\nu^*(\theta))(V^\theta(x^\prime,s^\prime)-\widetilde{V}^{\theta,v}(x^\prime,s^\prime))^2}\\
\leq &\gamma\sqrt{\sum_{x^\prime,a^\prime,s^\prime} \pi_{\gamma}^\theta(x^\prime,s^\prime,a^\prime|x_0=x^0,s_0=\nu^*(\theta))\sum_{x^{\prime\prime},s^{\prime\prime}} \bar{P}(x^{\prime\prime},s^{\prime\prime}|x^\prime,s^\prime,a^\prime) (V^\theta(x^{\prime\prime},s^{\prime\prime})-\phi^\top(x^{\prime\prime},s^{\prime\prime})v)^2}\\
&+\frac{\epsilon_\theta(v)}{1-\gamma}\\
\leq &\sqrt{\sum_{x^{\prime\prime},s^{\prime\prime}}\left(d_{\gamma}^\theta(x^{\prime\prime},s^{\prime\prime}|x^0,\nu^*(\theta))-(1-\gamma)1\{x^0=x^{\prime\prime},\nu=s^{\prime\prime}\} \right)(V^\theta(x^{\prime\prime},s^{\prime\prime})-\phi^\top(x^{\prime\prime},s^{\prime\prime})v)^2}+\frac{\epsilon_\theta(v)}{1-\gamma}\\
\leq &\left(\frac{1+\gamma}{1-\gamma}\right)\epsilon_\theta(v).
\end{split}
\]

Then by Lemma~\ref{lem:unbiased_L}, if the
$\gamma$-{\color{black} occupation measure}  $\pi_\gamma^\theta$ is used to generate samples $(x_k,s_k,a_k)$, one obtains $\mathbb E\left[\delta\theta_{k+1}\mid \mathcal F_{\theta,k}\right]=0$, where $\mathcal F_{\theta,k}= \sigma(\theta_m,\,\delta \theta_m,\,m\leq k)$ is the filtration generated by different independent trajectories. On the other hand, $|\delta\theta_\epsilon|\rightarrow 0$ as $\epsilon_{\theta_k}(v^*(\theta_k))\rightarrow 0$. Therefore, the $\theta$-update in~\eqref{theta_up} is a stochastic approximation of {\color{black}the} continuous system $\theta(t)$, described by the ODE 
\[
\dot{\theta}=\Upsilon_\theta\left[ -\nabla_\theta L(\nu,\theta,\lambda)\vert_{\nu=\nu^*(\theta)} \right],
\] 
with an error term that is a sum of a vanishing bias and a Martingale difference. Thus, the convergence analysis of $\theta$ follows analogously from Step 2 in the proof of Theorem~\ref{thm:converge_h}, i.e.,~the sequence of $\theta$-updates in~\eqref{theta_up_kncre} converges to $\theta^*$ almost surely, where $\theta^*$ is the equilibrium point of the continuous system $\theta$ satisfying 
\begin{equation}\label{eq_cond_1}
\Upsilon_\theta\left[ -\nabla_\theta  L(\nu,\theta,\lambda) \vert_{\nu=\nu^*(\theta)}\right]=0.
\end{equation}
\end{prooff}

\noindent\paragraph{Step 4 (Local Minimum)}
The proof {\color{black}that} $(\theta^\ast,\nu^\ast)$
{\color{black}is a local minimum} follows directly from the arguments in Step 3 in the proof of Theorem~\ref{thm:converge_h}.

\noindent\paragraph{Step 5 ($\lambda$-update and Convergence to Local Saddle Point)}
Note that {\color{black}the} $\lambda$-update converges
{\color{black}on the} slowest time scale, {\color{black} according to previous analysis, we have that $\|\theta_k-\theta^*(\lambda_k)\|\rightarrow 0$, $\|\nu_k-\nu^*(\lambda_k)\|\rightarrow 0$ almost surely. By continuity of $\nabla_\lambda L(\nu,\theta,\lambda)$, we also have the following:
\[
\left\|\nabla_\lambda L(\nu,\theta,\lambda)\bigg\vert_{\theta=\theta^*(\lambda_k),\nu=\nu^*(\lambda_k),\lambda=\lambda_k}-\nabla_\lambda L(\nu,\theta,\lambda)\bigg\vert_{\theta=\theta_k,\nu=\nu_k,\lambda=\lambda_k}\right\|\rightarrow 0.
\]
}
Thus,~\eqref{nu_up_kncre_SPSA} may be rewritten as
\begin{equation}\label{lamda_up_conv}
\lambda_{k+1}=\Gamma_\Lambda\left(\lambda_k+\zeta_1(k)\left(\nabla_\lambda L(\nu,\theta,\lambda)\bigg\vert_{\theta=\theta^*(\lambda),\nu=\nu^*(\lambda),\lambda=\lambda_k}+\delta\lambda_{k+1}\right)\right),
\end{equation}
where 
\begin{equation}\label{eq:MG_diff_theta}
\begin{split}
&\delta\lambda_{k+1}=-\nabla_\lambda L(\nu,\theta,\lambda)\bigg\vert_{\theta=\theta^*(\lambda),\nu=\nu^*(\lambda),\lambda=\lambda_k}+\\
&\qquad\left({\color{black}\underbrace{(\nu_k-\nu^*(\lambda_k))}_{\|\nu_k-\nu^*(\lambda_k)\|\rightarrow 0}}+\nu^*(\lambda_k)+\frac{(-s_k)^+}{(1-\alpha)(1-\gamma)}\mathbf 1\{x_k={\color{black} x_{\text{Tar}}}\} -\beta\right).
\end{split}
\end{equation}
From~\eqref{eq:L_lambda}, $\nabla_\lambda L(\nu,\theta,\lambda)$
{\color{black}does not depend on} $\lambda$. Similar to {\color{black}the} $\theta$-update, one can easily show that $\delta\lambda_{k+1}$ is square integrable, {\color{black}specifically},
\[
\mathbb E[\|\delta\lambda_{k+1}\|^2\mid \mathcal F_{\lambda,k}]\leq 8\left(\beta^2+\left(\frac{D_{\max}}{1-\gamma}\right)^2+\left(\frac{2D_{\max}}{(1-\gamma)^2(1-\alpha)}\right)^2\right),
\]
where $\mathcal
F_{\lambda,k}=\sigma\big(\lambda_m,\,\delta\lambda_m,\,m\leq k\big)$
is the filtration of $\lambda$ generated by different independent
trajectories. Similar to the $\theta$-update, using the
$\gamma$-{\color{black} occupation measure}  $\pi_{\gamma}^\theta$, one obtains $\mathbb E\left[\delta\lambda_{k+1}\mid \mathcal F_{\lambda,k}\right]=0$. As above,
the $\lambda$-update is a stochastic approximation {\color{black}for} the continuous system $\lambda(t)$ described by the ODE
\[
\dot{\lambda}=\Upsilon_\lambda\left[ \nabla_\lambda L(\nu,\theta,\lambda)\bigg\vert_{\theta=\theta^*(\lambda),\nu=\nu^*(\lambda)}\right],
\]
with an error term that is a Martingale difference. Then the $\lambda$-convergence and the analysis of  {\color{black}local optima} follow from analogous arguments in Steps~4~and~5~in the proof of Theorem~\ref{thm:converge_h}.

\end{document}